\documentclass[11pt]{article}

\usepackage[preprint]{acl}

\usepackage{times}
\usepackage{latexsym}

\usepackage[T1]{fontenc}

\usepackage[utf8]{inputenc}

\usepackage{microtype}

\usepackage{inconsolata}

\usepackage{graphicx}
\usepackage{subcaption}
\usepackage{booktabs}
\usepackage{multirow}
\usepackage[most]{tcolorbox}
\usepackage{makecell}
\usepackage{tabularx}

\usepackage{fdsymbol}
\newcommand\dataset{TutorCaseBench}

\title{\dataset{}: A Case-Grounded Benchmark for Evaluating LLM Tutors in Supporting Students with Challenging Behaviors}

\author{
  \textbf{Yeon Su Park}$^{\varheartsuit}$\hspace{0.5em}
  \textbf{Chanyoung Park}$^{\heartsuit}$\hspace{0.5em}
  \textbf{Tae Soo Kim}$^{\varheartsuit}$\hspace{0.5em}
  \textbf{Seungju Kim}$^{\spadesuit}$\\
  \textbf{Hyunwoo Kim}$^{\varheartsuit}$\hspace{0.5em}
  \textbf{Daho Jung}$^{\varheartsuit}$\hspace{0.5em}
  \textbf{Jinsung Yoon}$^{\clubsuit}$\hspace{0.5em}
  \textbf{Juho Kim}$^{\varheartsuit,\diamondsuit}$\\[0.5em]
  {\small
  $^{\varheartsuit}$KAIST\hspace{0.5em}
  $^{\heartsuit}$Seoul National University\hspace{0.5em}
  $^{\clubsuit}$Google Cloud AI Research}\\[-0.3em]
  {\small
  $^{\spadesuit}$Korea Institute for Curriculum and Evaluation\hspace{0.5em}
  $^{\diamondsuit}$SkillBench}\\[0.1em]
  {\small\texttt{\{yeonsupark,\hspace{0.3em}taesoo.kim,\hspace{0.3em}khw0726,\hspace{0.3em}dhj1104,\hspace{0.3em}juhokim\}@kaist.ac.kr}}\\[-0.4em]
  {\small
  \texttt{pcy1203@snu.ac.kr}\hspace{0.5em}
  \texttt{seun8ju@kice.re.kr}\hspace{0.5em}
  \texttt{jsyoon0823@gmail.com}}
}

\begin{document}
\maketitle

\begin{abstract}
    LLMs are increasingly used as AI tutors, yet existing benchmarks overlook the \emph{challenging behaviors} common among students with learning and behavioral difficulties. We introduce \dataset{}, a benchmark that grounds AI-tutor evaluation in qualitative case studies from pedagogy, educational psychology, and special education. Our pipeline converts documented student categories into case-specific scenarios, tutor--student dialogues, and rubrics grounded in source evidence. The benchmark contains 923 dialogues across five challenge types, evaluated through open-ended response generation (1{,}560 items) and multiple-choice selection (2{,}537 items). Across 11 state-of-the-art LLMs, even the strongest model satisfies a rubric criterion in only 55.8\% of open-ended cases, while models are better at selecting appropriate strategies than generating situated responses. These results reveal a gap between intervention recognition and context-specific adaptation, especially for open-weight models. All data are publicly available at \url{https://github.com/kixlab/TutorCaseBench}.
\end{abstract}

\section{Introduction}

Large language models (LLMs) are increasingly deployed as conversational AI tutors for scalable, individualized instruction~\citep{jurenka2025towards}. Yet, effective tutoring requires more than providing correct answers: a tutor who misreads emotion, fails to re-engage, or escalates refusal can demotivate the student, making affective and behavioral factors as consequential as cognitive ones~\citep{baker2010frustrated,dmello2013autotutor,karumbaiah2023spatiotemporal}. This is especially important for students with learning and behavioral difficulties, whose struggles often appear as refusal, withdrawal, emotional distress, or social conflict~\citep{sugai2006pbis}. We use \textit{challenging behaviors} to refer to such disruptive or disengaged responses, which make tutoring difficult to sustain and require support tailored to the learner and context~\citep{pepe2013comparison,patnaik2023challenging}.

Existing AI-tutor benchmarks~\citep{macina-etal-2023-mathdial,macina-etal-2025-mathtutorbench,sonkar-etal-2023-class} focus on cognitive aspects of tutoring, such as math-error detection and scaffolding. Their inputs often depict idealized student behavior rather than disengagement, resistance, or other challenging behaviors~\citep{macina-etal-2023-mathdial,markel2023gpteach,liu-etal-2024-personality}, while their rubrics rely on common pedagogical criteria~\citep{shridhar-etal-2022-automatic,liu2024socraticlm}. As a result, they offer limited evidence on whether AI tutors can provide case-sensitive support for students with challenging behaviors.

\begin{figure}[t!]
    \centering
    \includegraphics[width=\linewidth]{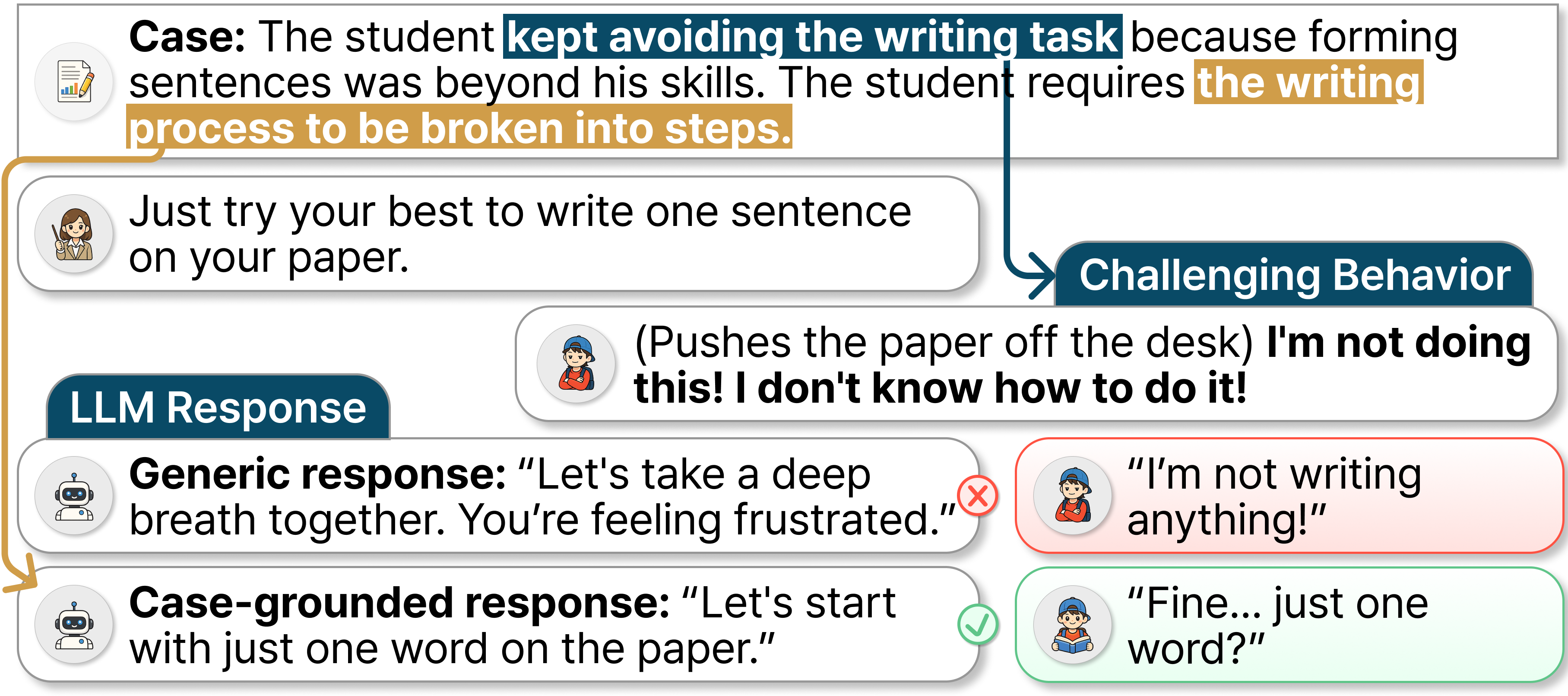}
    \caption{We introduce \dataset{}, a benchmark grounded in qualitative case studies for evaluating LLM tutors' ability to respond to challenging student behaviors. Example from \citet{park2017function}.}
    \label{fig:overview}
\end{figure}

Our work grounds this evaluation in a resource the NLP community has so far overlooked: qualitative case literature from pedagogy, educational psychology, and special education. These studies document individual students, their behaviors in learning contexts, and interventions that practitioners observed to be effective or ineffective. Unlike hypothetical personas, such cases provide evidence about student backgrounds, precipitating events, and practitioner responses, allowing us to synthesize realistic pedagogical scenarios involving challenging behaviors and documented interventions.

We introduce an automated LLM pipeline that converts qualitative case-study papers into structured tutoring evaluation cases. For each documented student category, the pipeline extracts verbatim learner and instructional evidence and constructs scenarios with learning contexts, challenging behaviors, and intervention strategies. Each scenario is then converted into a grounded tutor--student dialogue ending with the student exhibiting challenging behavior, paired with a rubric derived from the documented interventions. The resulting items are evaluated in \emph{immediate} and \emph{long-term} settings through an LLM-as-a-judge checklist rubric~\citep{zheng2023judging} and a multiple-choice task in the style of~\citet{hendrycks2021measuring}.

Applying this pipeline to qualitative case studies, we construct \dataset{}, a benchmark of 923 English dialogues across five challenge types: emotional distress, task resistance, social-interpersonal difficulty, attention/engagement failure, and academic struggle. It contains 1{,}560 open-ended and 2{,}537 multiple-choice evaluation items.

We evaluate 11 state-of-the-art LLMs on \dataset{} and find that they remain limited in providing case-specific support. Even the strongest model satisfies a documented intervention in only 55.8\% of open-ended cases, while models perform better at selecting provided strategies than at generating tailored responses. Our benchmark provides systematic evidence that LLM tutors struggle to support students with challenging behaviors even in controlled settings, motivating further investigation in more interactive and situated tutoring contexts.

\medskip
\noindent\textbf{Contributions.}
\begin{enumerate}
\item A case-grounded pipeline that converts pedagogical case studies into tutoring scenarios, dialogues, and evaluation criteria.
\item \dataset{}, a benchmark of 923 case-grounded dialogues covering five types of challenging student behavior, with both open-ended and multiple-choice evaluation settings.
\item An empirical analysis of 11 LLM tutors, revealing difficulty in providing case-specific support, especially in long-term interventions.

\end{enumerate}
\section{Related Work}

\subsection{Literature-Grounded Student Simulations}

AI tutor evaluation requires realistic student--tutor interactions, but collecting them at scale is difficult~\citep{macina-etal-2023-mathdial}. LLM-based student simulation offers a useful alternative, typically modeling learner knowledge and reasoning, such as misconceptions and knowledge components~\citep{macina-etal-2023-mathdial,lu2024generative,qi2026simulating}, or learner traits, such as personas and motivational profiles~\citep{markel2023gpteach,liu-etal-2024-personality,jin2025teachtune,marquez2026simulating}.

However, even fine-tuned models show limited alignment with real student behavior~\citep{scarlatos2026simulated}. Simulated students also rarely express frustration or disengagement, remain unnaturally attentive and linguistically sophisticated, and accept tutor feedback without resistance~\citep{martynova-etal-2025-llms}. Thus, existing frameworks mostly simulate idealized student behavior, overlooking disengaged, resistant, or off-task behaviors.

Source documents offer another way to ground dialogue synthesis. Prior work shows that document grounding improves coherence and factual consistency~\cite{feng-etal-2020-doc2dial,feng-etal-2021-multidoc2dial}, and supports persona simulation when documents describe behavior and internal states~\citep{wang2025coser}. In education, textbook-grounded dialogue generation improves content accuracy~\citep{wang-etal-2024-book2dial}. Hence, we ground synthesis in pedagogy, educational psychology, and special education literature, enabling dialogues with realistic challenging behaviors beyond cooperative learner settings.

\subsection{Evaluation of AI Tutors}

AI tutor evaluation has shifted from general-purpose NLG metrics to pedagogy-grounded frameworks. Earlier approaches relied on BLEU~\cite{papineni-etal-2002-bleu} and BERTScore~\cite{zhang2019bertscore}, comparing tutor responses against reference answers, but such metrics poorly capture pedagogical quality because multiple valid responses may exist~\cite{tack2022ai,jurenka2025towards}.

To address this, prior work has proposed pedagogy-grounded criteria. \citet{tack2022ai} found that AI responses fell short of human teachers in teacher-likeness, student understanding, and helpfulness. Later frameworks introduced criteria such as coherence, correctness, equitable tutoring~\citep{macina-etal-2023-mathdial}, as well as usefulness, care, and human-likeness~\citep{wang-etal-2024-bridging}. However, these criteria remain fragmented and lack consensus~\citep{jurenka2025towards}. Recent benchmarks have begun to consolidate them, including MRBench's eight-dimensional taxonomy~\citep{maurya-etal-2025-unifying} and MathTutorBench's focus on mathematical expertise, student understanding, and scaffolding quality~\citep{macina-etal-2025-mathtutorbench}. 

Despite this progress, existing benchmarks largely use fixed criteria, although support for students exhibiting challenging behaviors is often context-dependent. Thus, we take a case-specific approach, deriving evaluation criteria from interventions documented in pedagogical case studies.
\section{Data Construction} \label{sec:data_construction}

We construct \dataset{} from case-study papers documenting diverse learning and behavioral challenges, using an automated LLM pipeline to convert them into structured tutor--student dialogues in English for evaluation. The benchmark includes two tasks: open-ended response generation and multiple-choice question (MCQ) selection.

We use \texttt{gemini-3-flash-preview}~\cite{google2025gemini3flash} for extractive stages (Stage 1--2) and \texttt{gemini-3.1-pro-preview}~\cite{google2026gemini31pro} for generative stages (Stage 3--5), with default thinking \texttt{high} and \texttt{temperature$=$0}. For consistency, we used the Gemini models across the pipeline and rerouted only a number of persistent failures to OpenAI models (see Appendix~\ref{app:model_fallback}).

\subsection{Paper Collection} \label{sec:collection}

\paragraph{Stage 0: Paper Collection.} We collect candidate case-study papers from the ERIC database\footnote{\url{https://eric.ed.gov/}} using the ERIC API for papers that include at least one student \textit{and} one methodology descriptor. The descriptors are listed in Appendix~\ref{app:eric-descriptors}. We restrict publications to 2010--2026 to keep the corpus contemporary while ensuring sufficient volume. Since descriptor matching does not ensure sufficient detail on student behavior and teacher actions, we further filter papers by their abstracts (see Appendix~\ref{app:abstract_filtering}).

\subsection{Open-Ended Response Task} \label{sec:openended}

\dataset{}'s open-ended response task evaluates whether a model produces a next tutor response aligned with documented interventions for a tutor--student dialogue grounded in case studies.

\subsubsection{Tutor--Student Dialogue Construction} \label{sec:pipeline}

\begin{figure*}[hbt!]
    \centering
    \includegraphics[width=\linewidth]{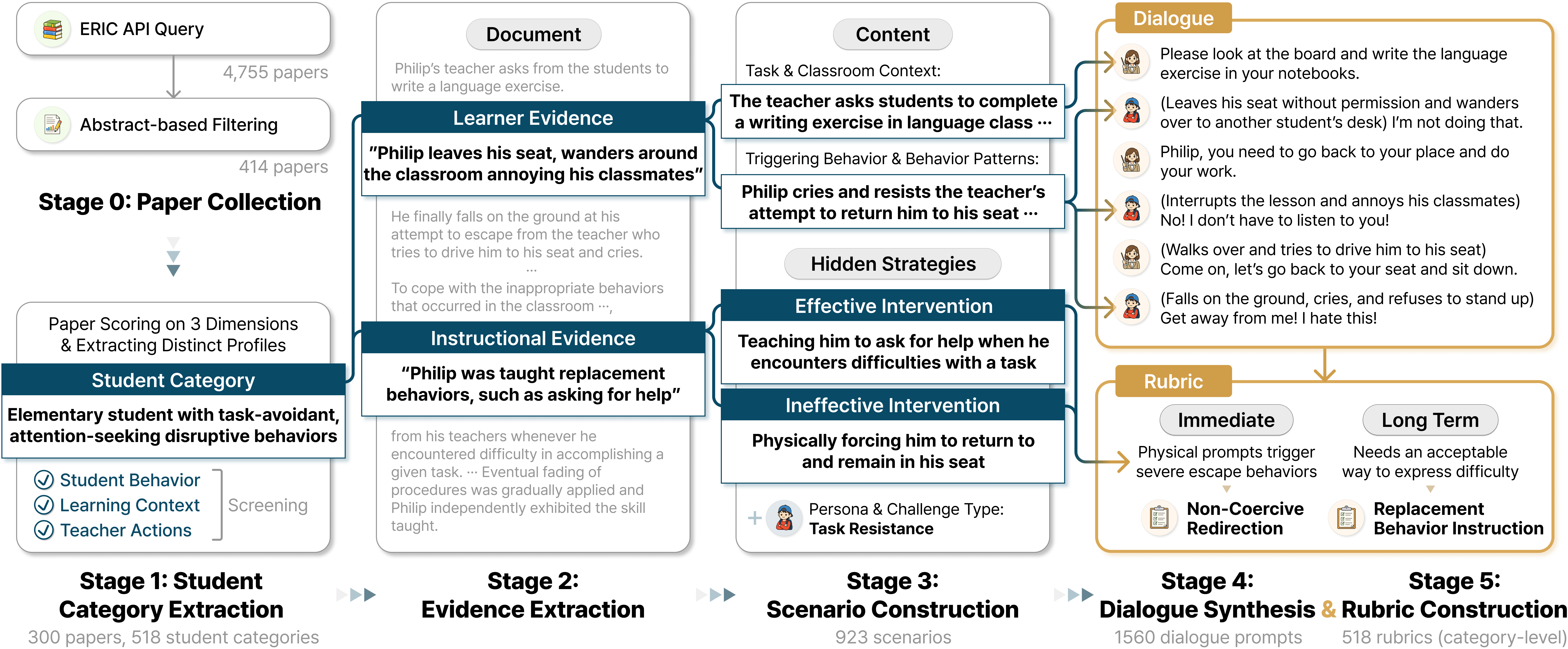}
    \caption{Overview of the \dataset{} construction pipeline. Based on extracted evidence, we compose case-grounded tutor--student dialogues. Examples are adapted from \citet{theodoridou2013functional}.}
    \label{fig:pipeline}
\end{figure*}

Stages 1--4 convert the collected papers into tutor--student dialogues paired with a hidden answer key of effective and ineffective interventions. Figure~\ref{fig:pipeline} provides an overview of this pipeline.

\paragraph{Stage 1: Student Category Extraction.} Starting from the initial set of papers, the LLM rates how concretely each paper documents three dimensions essential to our benchmark---\textit{student behavior}, \textit{learning context}, and \textit{teacher actions}---on a 1--5 scale (prompt in Appendix~\ref{app:stage1_screen}). Papers scoring below 3 on any dimension are filtered out. We set this cutoff after inspecting samples and finding them too general to ground a concrete dialogue.

For papers that pass screening, the model extracts \textit{student categories}---distinct student profiles within the paper, each with a label, rationale, demographics, and supporting excerpts for the three dimensions above (prompt in Appendix~\ref{app:stage1_category}). We use each category as a unit for scenario construction, since a paper may document multiple students or groups with different behaviors, contexts, or instructional needs. Categories are then filtered by the same dimensions at a stricter threshold (3 or below), since downstream dialogue synthesis requires per-category evidence to be concrete on its own.

\paragraph{Stage 2: Evidence Extraction.} Next, the LLM reads each paper from Stage 1 twice---once for learner evidence, once for instructional evidence---tagging each verbatim span with the categories it applies to. The first read extracts learner evidence along four dimensions---\textit{observable behaviors}, \textit{cognitive states}, \textit{affective states}, and \textit{difficulty triggers}---along with verbatim student quotes (prompt in Appendix~\ref{app:stage2_learner}). These dimensions adapt Pintrich's self-regulated learning framework~\cite{pintrich2000role, panadero2017review}, replacing its contextual axis with difficulty triggers to surface what specifically prompted the student's struggle.

The second read takes the paper and the prior learner spans as input, and extracts spans of instructional evidence, recording \textit{when} the strategy applies, \textit{how} it was enacted, and its documented \textit{outcome} (prompt in Appendix~\ref{app:stage2_instruction}). Each instructional span is tagged as \textit{effective} or \textit{ineffective}, given a temporal scope of \textit{immediate}, \textit{long-term}, or \textit{both}.

\paragraph{Stage 3: Scenario Construction.} The LLM then composes one or more concrete \textit{scenarios} per category from the extracted evidence (prompt in Appendix~\ref{app:stage3_scenario}). We construct multiple scenarios when evidence spans distinct tasks or classroom contexts, rather than collapsing them into a generic scenario.

Each scenario is organized as two parallel blocks: a \textit{content} block (task, classroom context, triggering behavior, related behavior patterns) that frames what a tutor sees in the moment, paired with a \textit{hidden strategies} block (immediate effective, long-term effective, and ineffective) that is later used to derive the evaluation rubric. Each element is linked back to Stage 2 evidence spans. We retain scenarios only when all content fields are specified, and at least one effective strategy is documented.

At the category level, the model produces two additional artifacts. The \textit{student persona} captures the learner's recurring patterns, triggers, and what tends to work or fail, ensuring consistent student turns across all scenarios in the category. Each category is also assigned a \textit{primary challenge type} (task resistance, emotional distress, attention/engagement failure, social-interpersonal difficulty, or academic struggle) using a dialogue-based taxonomy of challenge types (see Appendix~\ref{app:taxonomy} for details), allowing us to examine model responses across qualitatively different learner difficulties.

\paragraph{Stage 4: Dialogue Synthesis.} In the final stage, the LLM converts each scenario into a 6-turn tutor--student \textit{dialogue} (i.e., three tutor and three student turns) that ends with the student exhibiting the challenging behavior (prompt in Appendix~\ref{app:stage4_dialogue}). Given the student category, persona, and scenario content, the model generates a coherent dialogue in which every student turn is grounded in one or more excerpts. The hidden strategies block is excluded and tutor turns are constrained to be neutral. 

The model also generates a one-sentence \textit{minimal context} containing only the task and setting, leaving the student's needs to be inferred from the interaction. The full scenario content is preserved as \textit{full context} for downstream grading.

We use six turns to provide sufficient interaction context around the challenging behavior, as tutor responses may depend on how the interaction has developed across multiple exchanges. Limiting dialogue length also keeps synthesized interactions grounded in case evidence, avoiding unsupported details and repetition. To assess this choice, we additionally test 4- and 8-turn variants and observe consistent results (see Appendix~\ref{app:turn_ablation} for details).

\subsubsection{Open-Ended Response Evaluation} \label{sec:rubric}

\begin{figure}[h]
    \centering
    \includegraphics[width=\linewidth]{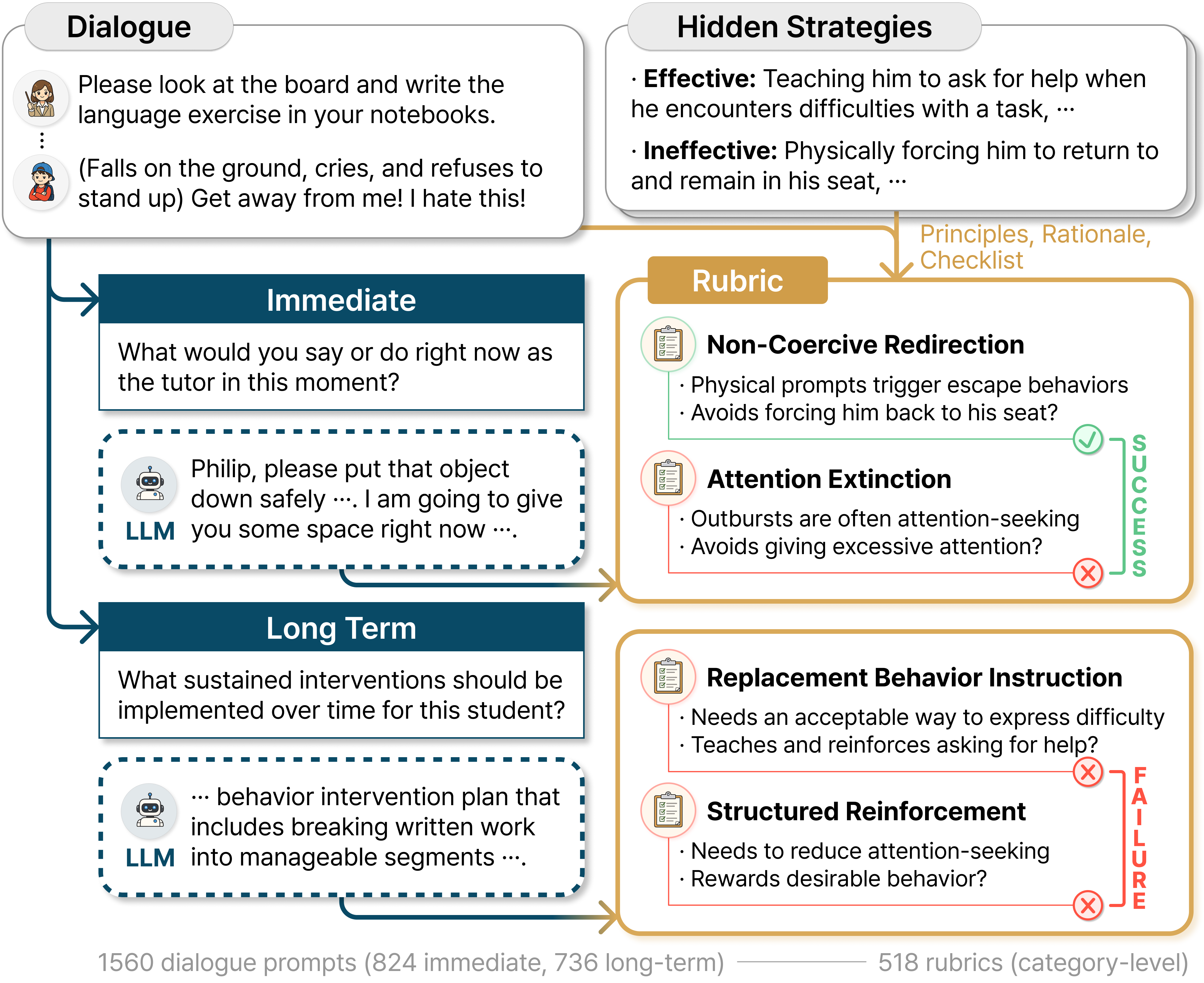}
    \caption{Rubric construction and evaluation process.}
    \label{fig:rubric}
\end{figure}

\paragraph{Stage 5: Rubric Construction.} To evaluate tutor responses, we create case-specific rubrics by student category and response type (immediate or long-term) (prompt in Appendix~\ref{app:stage5_rubric}). Rather than scoring responses on fixed dimensions, each rubric is built from effective intervention strategies documented for that student. We define rubrics at the category level because scenarios within a category share a learner profile and intervention evidence.

Given the student persona, challenge type, synthesized dialogues, and hidden strategies from the category's scenarios, the model produces a checklist of criteria, each capturing an evidence-supported response principle with a rationale and check condition. Ineffective strategies are included as negative anchors, and a brief student description is produced as auxiliary context. Figure~\ref{fig:rubric} shows the rubric construction and evaluation process.

Each criterion is judged independently against its check condition and scored as satisfied or not (0--1). The criteria represent alternative evidence-supported approaches to the same student difficulty, rather than requirements to be combined in a single response. Requiring multiple criteria could favor broad and abstract responses over those that clearly apply a single approach. Thus, we count a response as successful if it satisfies at least one criterion.

\subsection{MCQ Selection Task} \label{sec:mcq}

To complement open-ended evaluation, we convert scenarios from \S\ref{sec:pipeline} into 4-option MCQs. Each question pairs one effective strategy from the target scenario with three from other scenarios as distractors, testing whether models identify the strategy that best fits the student and dialogue. MCQ accuracy thus reflects the ability to select interventions supported by case-specific evidence, without implying that alternative strategies are ineffective.

\begin{figure}[h]
    \centering
    \includegraphics[width=\linewidth]{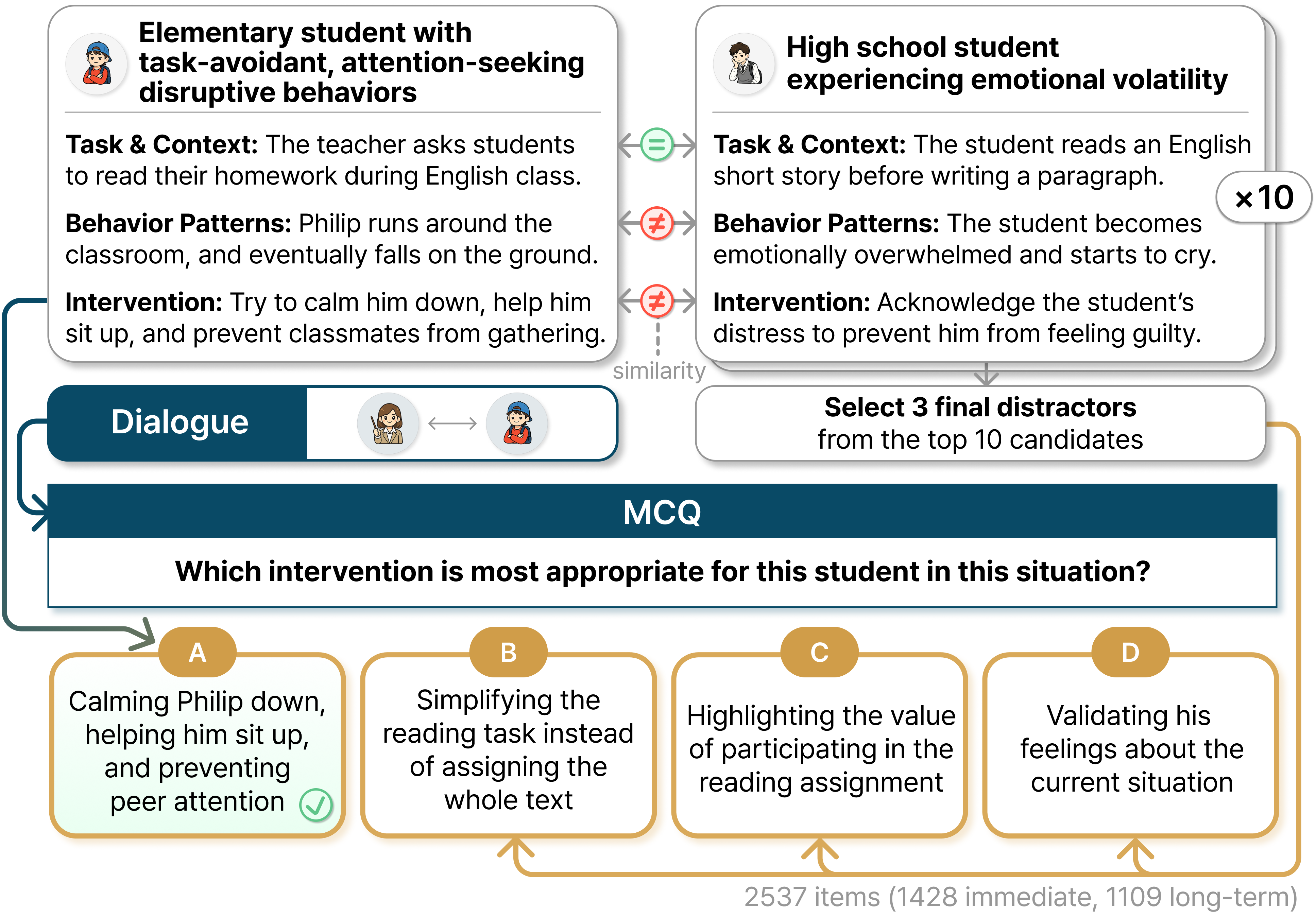}
    \caption{4-option MCQ construction with cross-scenario distractors. Examples from \citet{sander2010qualitative}, \citet{lichtinger2015employing}, \citet{newman2012home}.}
    \label{fig:mcq}
\end{figure}

Figure~\ref{fig:mcq} illustrates the MCQ construction process. We retrieve candidate distractors by embedding similarity, select three with an LLM judge, and rewrite all options into a uniform surface form. We also construct a binary variant pairing one effective and one ineffective strategy from the same scenario as a sanity check. Full construction details for both variants are provided in Appendix~\ref{app:mcq_construction}.

\subsection{Data Statistics} \label{sec:stats}

\paragraph{Overall.} Starting from 4{,}755 ERIC candidates, the pipeline produces 923 scenarios from 300 source papers and 518 student categories ($M=1.78$ scenarios per category, $SD=0.77$). The final benchmark includes 1{,}560 open-ended items: 824 immediate and 736 long-term. Category-level rubrics contain $2.90$ immediate criteria ($SD=0.91$) and $2.45$ long-term criteria ($SD=0.87$) on average. The MCQ set includes 2{,}537 items: 1{,}428 immediate and 1{,}109 long-term. Additional details on student demographics and corresponding model performance are provided in Appendix~\ref{app:demographics}.

\paragraph{Challenge-Type Distribution.} Our taxonomy of student challenges (Appendix~\ref{app:taxonomy}) covers the dataset scenarios as follows: Task Resistance (28.7\%), Attention/Engagement Failure (20.4\%), Academic Struggle (18.4\%), Emotional Distress (13.7\%), and Social-Interpersonal Difficulty (18.9\%).

\subsection{Data Validation} \label{sec:validation}
We conducted a human validation study to assess whether outputs from our pipeline were grounded in the source papers. We randomly sampled 30 papers and recruited 12 students (three undergraduates, nine graduates) from South Korea through online university forums, all with prior experience in reading and analyzing academic papers. Each reviewer evaluated five papers, with every paper independently assessed by two reviewers so that the validation did not depend on a single evaluator. 

Reviewers first read and summarized the source paper, then evaluated key outputs from each pipeline stage, providing a rationale for every judgment (see Appendix~\ref{app:validation_interface} for the study interface). On a three-point scale (No$=$1, Partly$=$2, Yes$=$3), mean ratings were 2.62 ($SD=0.56$) for student categories (Stage 1), 2.42 ($SD=0.56$) for evidence (Stage 2), 2.57 ($SD=0.57$) for scenarios and 2.48 ($SD=0.54$) for instructional strategies (Stage 3), 2.60 ($SD=0.60$) for dialogues (Stage 4), and 2.69 ($SD=0.46$) for rubrics (Stage 5). Challenge types were evaluated on a binary scale (No$=$0, Yes$=$1), yielding a mean rating of 0.76 ($SD=0.43$). Further details on human validation results are provided in Appendix~\ref{app:human_validation}. Overall, these results suggest that our pipeline outputs align with human judgments of what is supported by the source literature.
\section{Experiments}

We evaluate 11 state-of-the-art models on \dataset{} across three tasks: open-ended response generation, MCQ selection, and reference-based response generation. Experimental settings and model details are provided in Appendix~\ref{app:models}.

\paragraph{Open-Ended Response.} Each model produces a free-form response to each open-ended item, either an immediate next-turn response or a long-term intervention plan (prompts in Appendix~\ref{app:open_response}). Immediate items ask what the tutor should say or do next, while long-term items ask what sustained interventions should be implemented over several weeks. In both cases, the model is shown only the scenario's minimal context and the dialogue up to the challenging student utterance, requiring it to infer the student's needs and respond appropriately.

We evaluate each response with \texttt{gemini-3.1-pro-preview} using the rubric from \S\ref{sec:rubric} (prompt in Appendix~\ref{app:rubric_grade}). The evaluator receives the rubric, student description, dialogue, the scenario's full context, and the model response, and assigns each criterion a binary score (1 if satisfied, 0 otherwise). A response is marked successful if it satisfies at least one criterion. We also report the average criterion satisfaction rate in Appendix~\ref{app:normalized_score}. Since Gemini models are used for both construction and evaluation, we additionally test for model-family bias by varying the construction models (Appendix~\ref{app:pipeline_ablation}) and evaluation judge (Appendix~\ref{app:judge_ablation}), finding consistent results for both.

\paragraph{MCQ Selection.} For each MCQ from \S\ref{sec:mcq}, we provide the same minimal context and dialogue used in the open-ended setting, along with four answer options, and ask the model to select one (prompt in Appendix~\ref{app:mcq_response}). A response counts as correct if it matches the correct option, which is grounded in the scenario's effective strategy.

\begin{figure}[h]
    \centering
    \includegraphics[width=\linewidth]{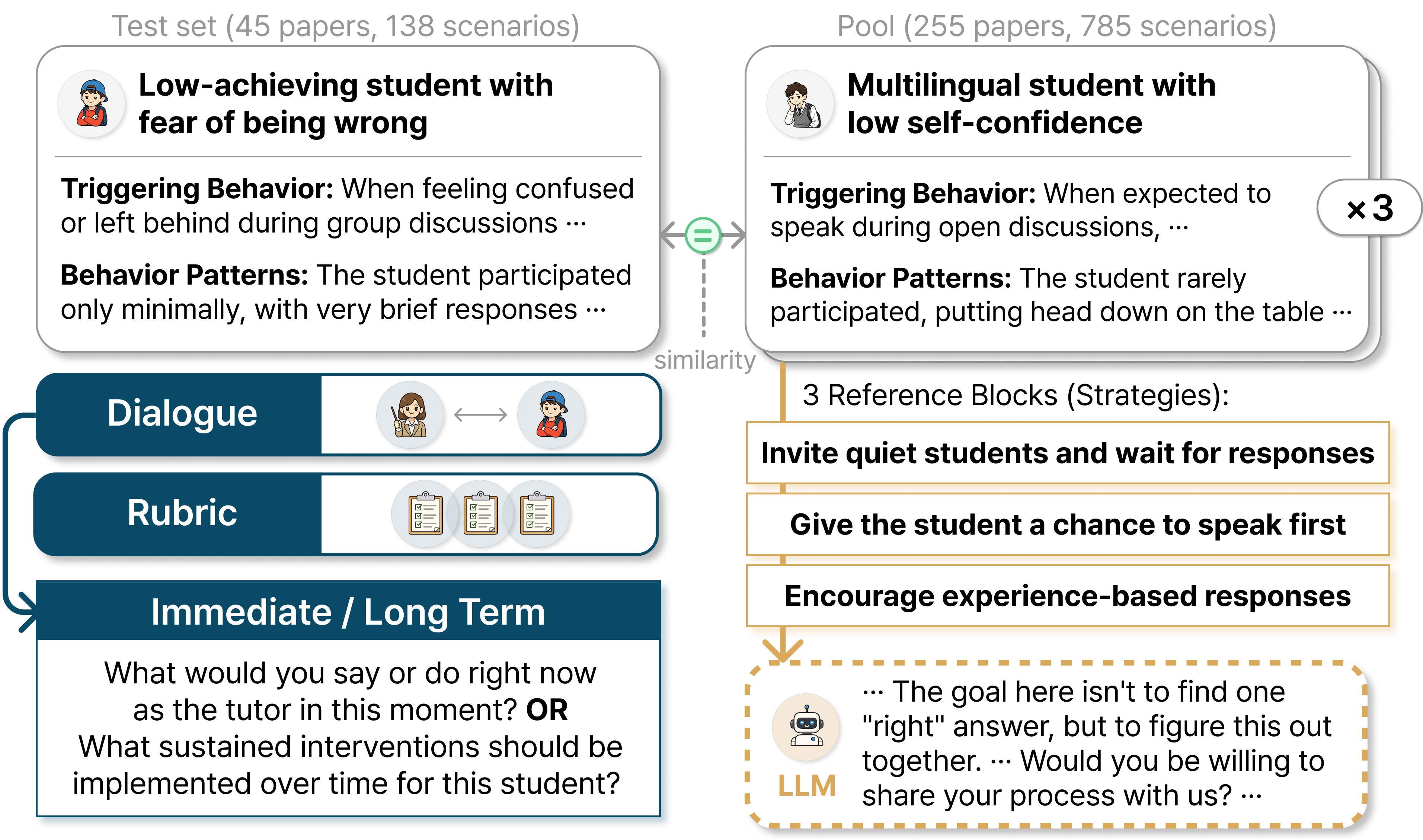}
    \caption{We use strategies from similar cases as references. Examples from \citet{lack2014low}, \citet{collett2023relationships}, \citet{alghamdi2016supporting}, \citet{cassidy2018philosophy}.}
    \label{fig:reference}
\end{figure}

\paragraph{Reference-Based Response.} We also create a reference-based variant to test whether models can produce better open-ended responses when relevant guidance is available. For each test dialogue, we retrieve the most similar scenarios from a separate pool and add their intervention strategies to the prompt as a reference (Figure~\ref{fig:reference}). We evaluate the resulting responses with the same rubrics as the dialogue-only setting, enabling a paired comparison. Full construction details are in Appendix~\ref{app:reference_details}.

\section{Evaluation Results}

\subsection{Open-Ended Response}

\paragraph{Overall Performance.} Table~\ref{tab:quant_open} reports the success rate of open-ended responses (i.e., the proportion satisfying at least one criterion in the rubric). Overall performance remains limited even for the strongest models. Proprietary models occupy the top ranks, with Claude Opus 4.7 performing best at 55.8\% and GPT-5.5 following at 54.7\%. Among open-weight models, DeepSeek V4 Pro performs best at 47.0\%, while Llama 4 Maverick is lowest at 31.7\%. These results indicate that generating pedagogically grounded responses for students with challenging behaviors remains difficult.

\begin{table}[h]
\small
\centering
\caption{Success rate of open-ended response (\%)}
\label{tab:quant_open}
\begin{tabular}{@{}p{2.3cm}ccc@{}}
\toprule
Model & \makecell{Immediate\\\scriptsize($N{=}824$)} & \makecell{Long-Term\\\scriptsize($N{=}736$)} & \makecell{Overall\\\scriptsize($N{=}1{,}560$)} \\
\midrule
Claude Opus 4.7   & 67.0 & 43.3 & 55.8 \\
Claude Haiku 4.5  & 59.0 & 40.1 & 50.1 \\
Gemini 3.1 Pro    & 67.5 & 35.2 & 52.2 \\
Gemini 3 Flash    & 58.7 & 32.2 & 46.2 \\
GPT-5.5           & 67.2 & 40.6 & 54.7 \\
GPT-5.4 Mini      & 49.2 & 33.3 & 41.7 \\
\midrule
Llama 4 Maverick  & 43.0 & 19.0 & 31.7 \\
Gemma 4 31B       & 52.4 & 29.1 & 41.4 \\
DeepSeek V4 Pro   & 55.6 & 37.4 & 47.0 \\
GPT-OSS 120B      & 52.9 & 34.6 & 44.3 \\
Qwen 3.5-397B     & 51.7 & 27.4 & 40.3 \\
\bottomrule
\end{tabular}
\end{table}

A consistent gap appears between immediate and long-term responses. All models perform better on immediate responses, with the largest gap for Gemini 3.1 Pro (67.5\% vs. 35.2\%) and the smallest for GPT-5.4 Mini (49.2\% vs. 33.3\%). This suggests that models are better at producing plausible in-the-moment responses, but struggle to formulate sustained interventions tailored to student needs.

\begin{figure}[h]
    \centering
    \includegraphics[width=\columnwidth]{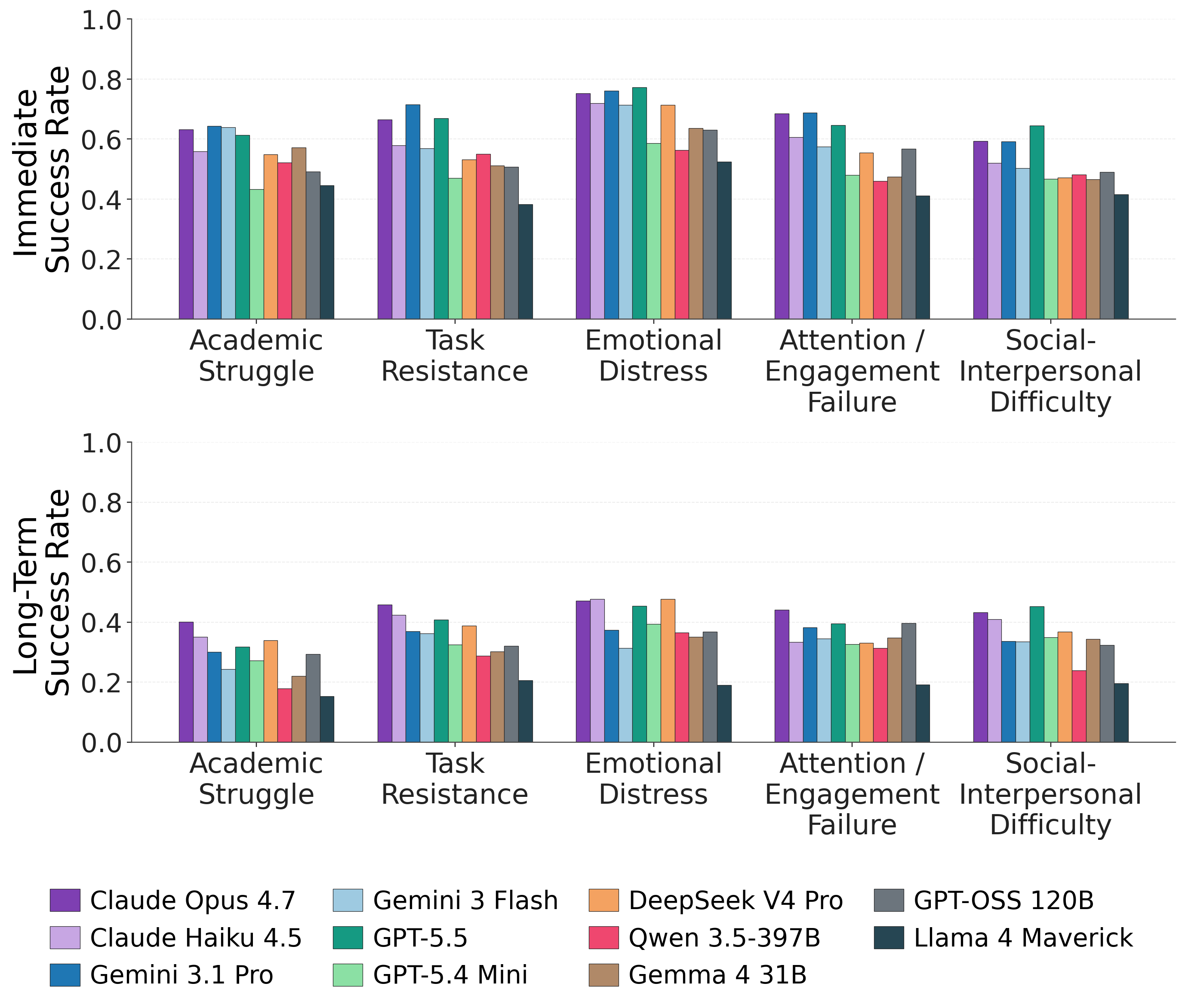}
    \caption{Success rate for open-ended response across models broken down by challenge type.}
    \label{fig:open-response}
\end{figure}

Figure~\ref{fig:open-response} breaks down open-ended performance by challenge type. For immediate responses, social-interpersonal difficulty has the lowest performance and ranks last for 6 of 11 models. For long-term interventions, academic struggle has the lowest performance and ranks last for all 11 models.

\paragraph{Diversity and Case-Specificity.} To better understand model-level differences, we additionally test whether stronger models produce less templated and more case-specific responses. We measure lexical diversity using the distinct-$n$ ratio---the fraction of unique $n$-grams in each model's responses---and compute Spearman correlations with success rates.

\begin{figure}[h]
    \centering
    \includegraphics[width=0.8\columnwidth]{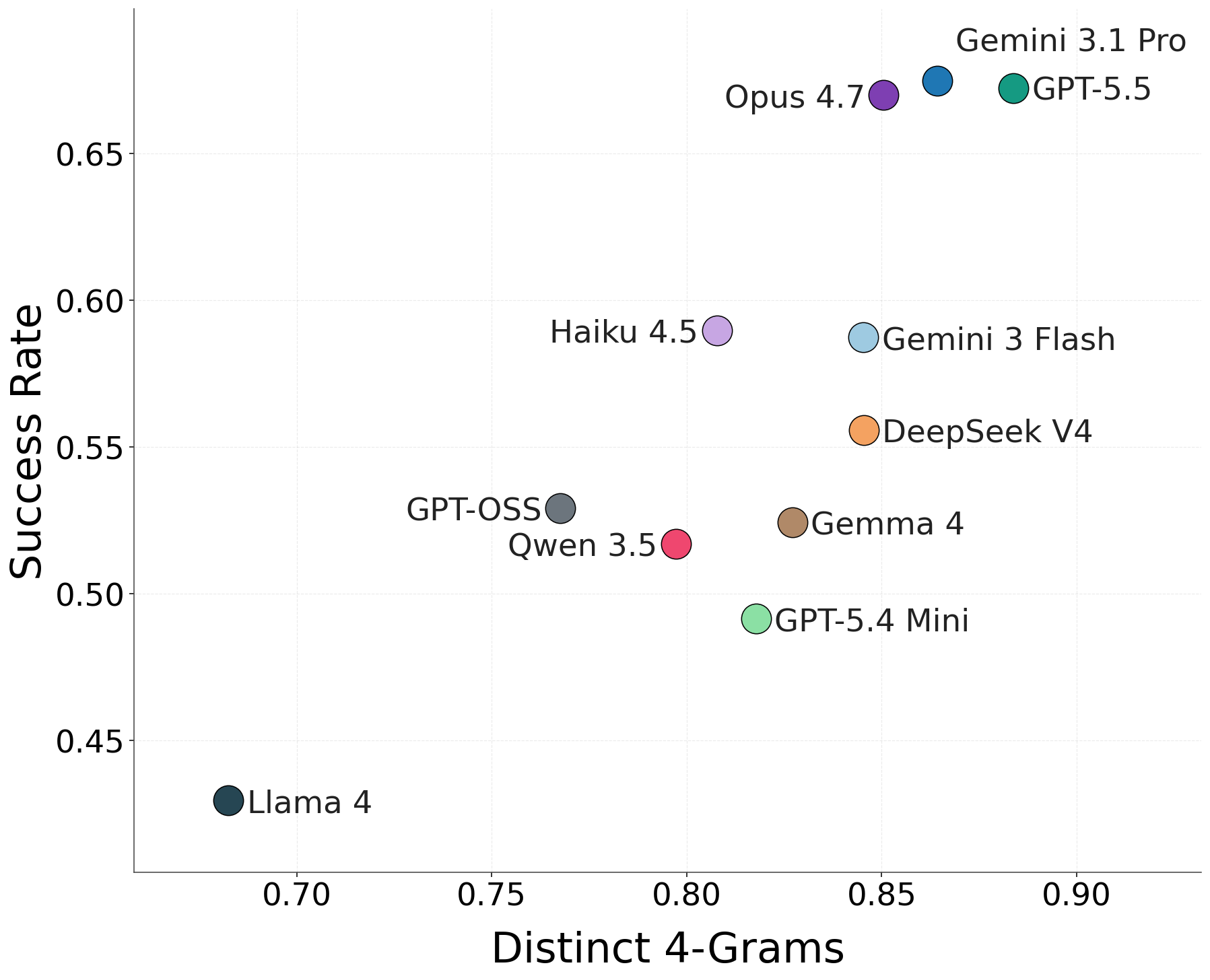}
    \caption{Correlation between distinct 4-grams ratio and immediate response success rate across models.}
    \label{fig:diversity_immediate}
\end{figure}

For immediate responses, 4-gram diversity shows a positive correlation with success rates ($\rho=0.800$, $p<0.01$), as shown in Figure~\ref{fig:diversity_immediate}. This suggests that higher-performing models tend to vary their responses more across cases rather than rely on repeated phrasing. For example, Llama 4 Maverick's most frequent 4-gram (``take a deep breath'') appears in 50.4\% of its immediate responses, whereas the most frequent 4-grams of Claude Opus 4.7, GPT-5.5, and Gemini 3.1 Pro appear in only 6--9\%. The relationship is not significant for long-term interventions ($p>0.05$). Results for 3- and 5-grams are in Appendix~\ref{app:diversity}.

We complement lexical diversity with case specificity. For each response, we use \texttt{text-embedding-3-large}~\cite{openai2024embedding} to retrieve the nearest response from another model and check whether it comes from the same dialogue case. We define the resulting case-recovery rate as a measure of case specificity and compute its Spearman correlation with success rates. Case-recovery rate is positively correlated with immediate-response success ($\rho=0.691$, $p<0.05$; see Appendix~\ref{app:specificity}), suggesting that stronger models produce more case-specific responses. The correlation is not significant for long-term interventions.

\paragraph{Failure Analysis.} \label{sec:oe_failure_modes}

We qualitatively analyze a sample of 50 open-ended cases where all models failed (25 immediate, 25 long-term). We identify four recurring failure patterns, shown in Table~\ref{tab:oe_failure_modes}.

\begin{table*}[h]
\small
\centering
\caption{Recurring failure modes in open-ended response evaluation missed by all models.}
\begin{tabularx}{\linewidth}{@{}lX@{}}
\toprule
\textbf{Failure Mode} & \textbf{Description} \\
\midrule
Empathy/Praise
  & Comforts the student when the response should challenge their belief or behavior. \\
Generic Scaffolding
  & Applies a standard support plan without identifying the student's central difficulty. \\
Over-Intervention
  & Steps in to correct or redirect when the tutor should step back instead. \\
Task Redesign Missed
  & Helps the student cope with a task that itself should be changed. \\
\bottomrule
\end{tabularx}
\label{tab:oe_failure_modes}
\end{table*}

\textit{Empathy/Praise.}
A student refuses to present because she believes her work is wrong and calls herself ``stupid''~\cite{macmath2010research}. The case-prescribed response is to verify that her work is accurate before presenting. Models instead reassure her that mistakes are acceptable, leaving her belief that the work is wrong unaddressed.

\textit{Generic Scaffolding.}
Another student with language-comprehension difficulties interprets ``5 times more'' additively and writes 10~\cite{bruno2024strategies}. The case-prescribed response is to use simpler terms such as ``double'' or ``triple.'' Models instead correct the answer, explain multiplication, or ask why he added the numbers, addressing the error but not the underlying language difficulty.

\textit{Over-Intervention.}
A normally quiet student becomes nervous while co-teaching a science lesson~\cite{adjapong2015rethinking}. The case-prescribed response is for the teacher to take a student role and intervene only for content issues. Models instead keep guiding him to breathe, speak louder, or begin with a suggested sentence, reducing his opportunity to lead independently.

\textit{Task Redesign Missed.}
A student refuses an assigned reading task and becomes disruptive during independent work~\cite{guardino2010changing}. The case-prescribed response is to replace the task with a simple activity such as a crossword puzzle, coloring, or journaling. All models instead focus on calming the student and returning them to the original task, leaving the rejected task unchanged.

\subsection{MCQ Selection}

Table~\ref{tab:quat_mcq} reports accuracy on the MCQ task. Scores are higher and less variable than in the open-ended task, suggesting that models can recognize an appropriate intervention when provided as an option, even when they struggle to formulate one from scratch. Claude Opus 4.7 performs best at 63.0\%, followed by Gemini 3.1 Pro at 62.6\% and GPT-5.5 at 62.4\%. The top open-weight model, Gemma 4 31B, reaches 57.3\%, while GPT-OSS 120B is the lowest at 49.2\%, still above the 25\% chance level.

\begin{table}[h]
\centering
\small
\caption{Accuracy on 4-option MCQ cases (\%).}
\label{tab:quat_mcq}
\begin{tabular}{@{}p{2.3cm}ccc@{}}
\toprule
Model & \makecell{Immediate\\\scriptsize($N{=}1{,}428$)} & \makecell{Long-Term\\\scriptsize($N{=}1{,}109$)} & \makecell{Overall\\\scriptsize($N{=}2{,}537$)} \\
\midrule
Claude Opus 4.7   & 63.2 & 62.7 & 63.0 \\
Claude Haiku 4.5  & 53.9 & 55.9 & 54.8 \\
Gemini 3.1 Pro    & 61.6 & 63.8 & 62.6 \\
Gemini 3 Flash    & 55.5 & 56.9 & 56.1 \\
GPT-5.5           & 62.7 & 62.0 & 62.4 \\
GPT-5.4 Mini      & 51.9 & 54.6 & 53.1 \\
\midrule
Llama 4 Maverick  & 51.9 & 51.4 & 51.7 \\
Gemma 4 31B       & 55.3 & 59.8 & 57.3 \\
DeepSeek V4 Pro   & 52.4 & 54.7 & 53.4 \\
GPT-OSS 120B      & 48.3 & 50.2 & 49.2 \\
Qwen 3.5-397B     & 55.9 & 56.0 & 55.9 \\
\bottomrule
\end{tabular}
\end{table}

Challenge-type patterns partly mirror the open-ended setting. For immediate responses, social-interpersonal difficulty is the weakest MCQ category for all 11 models, reinforcing its difficulty in open-ended generation. Academic struggle shows the opposite pattern, performing strongly in MCQ despite being the weakest category for long-term responses (see Appendix~\ref{app:mcq_challenge}). We also observe systematic convergence in MCQ errors (see Appendix~\ref{app:mcq_failure}).

\subsection{Reference-Based Response}

\begin{figure}[h]
    \centering
    \includegraphics[width=0.9\columnwidth]{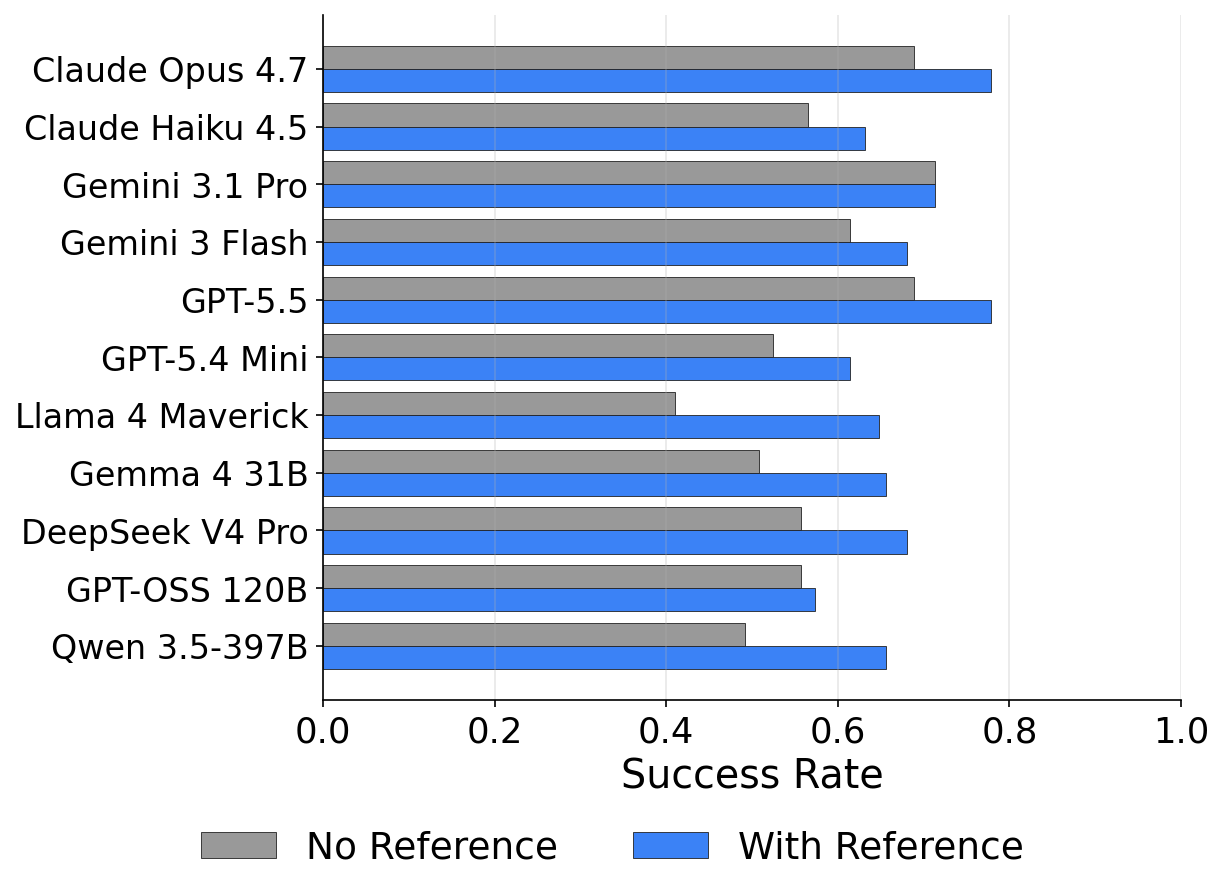}
    \caption{Success rate across models for immediate response, with and without references.}
    \label{fig:reference_immediate}
\end{figure}

We compare responses generated from the dialogue alone with those generated with strategies from similar prior cases as references. Figure~\ref{fig:reference_immediate} shows the effect on immediate responses. Most models benefit, with 10 of 11 improving and an average gain of 9.9 points overall. Llama 4 Maverick improves the most, from 41.0\% to 64.8\%, while Gemini 3.1 Pro does not improve (71.3\% in both conditions).

The reference effect does not extend to long-term plans (see Appendix~\ref{app:ref_longterm} for details). This suggests that providing references is more effective when they directly guide the next tutor move.
\section{Conclusion}

We introduce \dataset{}, a case-grounded benchmark for evaluating whether LLM tutors can support students with challenging behaviors. \dataset{} converts qualitative case studies into tutor--student dialogues and case-specific evaluation criteria, avoiding idealized learner simulations and fixed pedagogical rubrics. Across 923 dialogues and 11 state-of-the-art LLMs, we find that current models remain limited in generating support aligned with documented interventions, even when they often recognize appropriate strategies in multiple-choice settings. These results highlight the need for AI tutors that move beyond plausible responses toward context-sensitive support grounded in individual student needs. Overall, \dataset{} provides a resource for developing tutors that address not only cognitive errors but also students' affective and behavioral difficulties.
\newpage
\section*{Limitations}

Our work has several limitations. First, although \dataset{} is grounded in qualitative case-study literature, its construction primarily relies on the Gemini model family, which may introduce model-specific interpretations into the generated scenarios and rubrics. While human validation (\S\ref{sec:validation}) and robustness analyses with alternative construction models (Appendix ~\ref{app:pipeline_ablation}) and evaluation judges (Appendix ~\ref{app:judge_ablation}) support the consistency of our findings, future work could explore combining different model families across pipeline stages.

Second, the benchmark has limited coverage of some student populations and educational contexts. Although we draw on accessible, contemporary ERIC case studies with detailed evidence about student behavior, learning context, and teacher actions, the resulting categories are concentrated in particular regions and age groups (Appendix~\ref{app:demographics}). Future work could broaden this coverage by incorporating cases from additional databases, disciplines, countries, age groups, or classroom settings.

Third, \dataset{} evaluates LLM tutors' responses to challenging behaviors in text-based dialogues, enabling controlled evaluation independently of behavior detection. Although the dialogues describe actions and nonverbal cues, text may not fully capture their complexity in real interactions. Future work could extend this approach to interactive, multimodal settings where tutors must recognize and respond to challenging behaviors.

Lastly, we construct student categories separately within each paper rather than grouping similar cases across the corpus. This preserves case-specific intervention evidence but limits analysis of recurring patterns in student challenges and support strategies. Future work could cluster related categories to examine these patterns across cases.
\section*{Ethical Considerations}

The human validation study was reviewed and approved by the Institutional Review Board (IRB) of our institution prior to data collection. All participants provided informed consent before participation and were informed that they could withdraw at any time. As the study was conducted remotely, participants could contact the authors with any questions or concerns throughout the process. Validating each paper took approximately 90 minutes, and participants received 120,000 KRW ($\approx$ 89.68 USD) in compensation for their time.

Source documents were identified through ERIC (Education Resources Information Center), a U.S. Department of Education database for education research literature. We used only publicly accessible full-text materials, including documents hosted by ERIC and papers reachable through DOI or publisher links listed in ERIC records. Copyright remains with the original authors and publishers, and the materials were used only for non-commercial academic research. Our released benchmark consists only of LLM-synthesized tutor--student dialogues, rubrics, and multiple-choice items. We do not redistribute the full source texts or release intermediate extraction artifacts. Source excerpts used as examples in the paper are explicitly cited. For provenance, we provide source-paper metadata for the documents used in dataset construction.

Since \dataset{} focuses on students with learning and behavioral difficulties, some cases may include descriptions of distress, refusal, aggression, or other challenging behaviors. These behaviors are included only when grounded in the source literature and are used to evaluate supportive tutoring responses, not to stereotype or label students. We frame such behaviors as context-dependent student needs rather than as fixed traits.

We also acknowledge the use of AI assistants in preparing this work. We used ChatGPT\footnote{\url{https://chatgpt.com/}} and Claude\footnote{\url{https://claude.ai}} for writing assistance, as well as for generating icons used in figures. We also used Codex\footnote{\url{https://openai.com/codex/}} and Claude Code\footnote{\url{https://code.claude.com/docs}} for coding assistance in implementing the benchmark construction pipeline.


\newpage
\bibliography{custom}

@misc{jurenka2025towards,
    title={Towards Responsible Development of Generative AI for Education: An Evaluation-Driven Approach}, 
    author={Irina Jurenka and Markus Kunesch and Kevin R. McKee and Daniel Gillick and Shaojian Zhu and Sara Wiltberger and Shubham Milind Phal and Katherine Hermann and Daniel Kasenberg and Avishkar Bhoopchand and Ankit Anand and Miruna Pîslar and Stephanie Chan and Lisa Wang and Jennifer She and Parsa Mahmoudieh and Aliya Rysbek and Wei-Jen Ko and Andrea Huber and Brett Wiltshire and Gal Elidan and Roni Rabin and Jasmin Rubinovitz and Amit Pitaru and Mac McAllister and Julia Wilkowski and David Choi and Roee Engelberg and Lidan Hackmon and Adva Levin and Rachel Griffin and Michael Sears and Filip Bar and Mia Mesar and Mana Jabbour and Arslan Chaudhry and James Cohan and Sridhar Thiagarajan and Nir Levine and Ben Brown and Dilan Gorur and Svetlana Grant and Rachel Hashimshoni and Laura Weidinger and Jieru Hu and Dawn Chen and Kuba Dolecki and Canfer Akbulut and Maxwell Bileschi and Laura Culp and Wen-Xin Dong and Nahema Marchal and Kelsie Van Deman and Hema Bajaj Misra and Michael Duah and Moran Ambar and Avi Caciularu and Sandra Lefdal and Chris Summerfield and James An and Pierre-Alexandre Kamienny and Abhinit Mohdi and Theofilos Strinopoulous and Annie Hale and Wayne Anderson and Luis C. Cobo and Niv Efron and Muktha Ananda and Shakir Mohamed and Maureen Heymans and Zoubin Ghahramani and Yossi Matias and Ben Gomes and Lila Ibrahim},
    year={2025},
    eprint={2407.12687},
    archivePrefix={arXiv},
    primaryClass={cs.CY},
    url={https://arxiv.org/abs/2407.12687}, 
}

@article{baker2010frustrated,
    title={Better to be frustrated than bored: The incidence, persistence, and impact of learners’ cognitive–affective states during interactions with three different computer-based learning environments},
    author={Ryan S.J.d. Baker and Sidney K. D'Mello and Ma.Mercedes T. Rodrigo and Arthur C. Graesser},
    journal={International Journal of Human-Computer Studies},
    volume={68},
    number={4},
    pages={223--241},
    year={2010},
    issn={1071-5819},
    doi={https://doi.org/10.1016/j.ijhcs.2009.12.003}
}

@article{dmello2013autotutor,
    title={AutoTutor and affective autotutor: Learning by talking with cognitively and emotionally intelligent computers that talk back},
    author={D'Mello, Sidney and Graesser, Art},
    year={2013},
    issue_date={December 2012},
    publisher={Association for Computing Machinery},
    address={New York, NY, USA},
    volume={2},
    number={4},
    issn={2160-6455},
    journal={ACM Transactions on Interactive Intelligent Systems},
    month=jan,
    articleno={23},
    numpages={39},
    doi={10.1145/2395123.2395128},
    url={https://doi.org/10.1145/2395123.2395128}
}

@inproceedings{karumbaiah2023spatiotemporal,
    title = {A Spatiotemporal Analysis of Teacher Practices in Supporting Student Learning and Engagement in an AI-Enabled Classroom},
    author = {Karumbaiah, Shamya and Borchers, Conrad and Shou, Tianze and Falhs, Ann-Christin and Liu, Pinyang and Nagashima, Tomohiro and Rummel, Nikol and Aleven, Vincent},
    year = {2023},
    isbn = {978-3-031-36271-9},
    publisher = {Springer Nature Switzerland},
    address = {Cham, Switzerland},
    doi = {https://doi.org/10.1007/978-3-031-36272-9_37},
    booktitle = {Artificial Intelligence in Education},
    pages = {450--462},
    numpages = {13}
}

@article{sugai2006pbis,
    title = {A Promising Approach for Expanding and Sustaining School-Wide Positive Behavior Support},
    author = {George Sugai and Robert R. Horner},
    journal = {School Psychology Review},
    volume = {35},
    number = {2},
    pages = {245--259},
    year = {2006},
    publisher = {Routledge},
    doi = {10.1080/02796015.2006.12087989},
    url ={https://doi.org/10.1080/02796015.2006.12087989}
}

@article{pepe2013comparison,
    title={Comparison of occupational stress in response to challenging behaviours between general and special education primary teachers in Northern Italy},
    author={Pepe, Alessandro and Addimando, Loredana},
    journal={International Journal of Special Education},
    volume={28},
    number={1},
    pages={14--26},
    year={2013},
    publisher={ERIC},
    url={https://eric.ed.gov/?id=EJ1013679}
}

@article{patnaik2023challenging,
    title={Challenging Behaviours in Primary Classrooms: Examining Causal Attributions and Strategies Used by Primary School Teachers in India},
    author={Patnaik, Susmita and Subban, Pearl},
    journal={Australasian Journal of Special and Inclusive Education},
    volume={47},
    number={2},
    pages={78--95},
    year={2023},
    publisher={Cambridge University Press},
    doi={10.1017/jsi.2023.8}
}

@inproceedings{markel2023gpteach,
    title = {GPTeach: Interactive TA Training with GPT-based Students},
    author = {Markel, Julia M. and Opferman, Steven G. and Landay, James A. and Piech, Chris},
    year = {2023},
    isbn = {9798400700255},
    publisher = {Association for Computing Machinery},
    address = {New York, NY, USA},
    url = {https://doi.org/10.1145/3573051.3593393},
    doi = {10.1145/3573051.3593393},
    booktitle = {Proceedings of the Tenth ACM Conference on Learning @ Scale},
    pages = {226--236},
    numpages = {11},
    location = {Copenhagen, Denmark},
    series = {L@S '23}
}

@inproceedings{liu2024socraticlm,
    title = {SocraticLM: Exploring Socratic Personalized Teaching with Large Language Models},
    author = {Liu, Jiayu and Huang, Zhenya and Xiao, Tong and Sha, Jing and Wu, Jinze and Liu, Qi and Wang, Shijin and Chen, Enhong},
    booktitle = {Advances in Neural Information Processing Systems},
    doi = {10.52202/079017-2721},
    editor = {A. Globerson and L. Mackey and D. Belgrave and A. Fan and U. Paquet and J. Tomczak and C. Zhang},
    pages = {85693--85721},
    publisher = {Curran Associates Inc.},
    url = {https://proceedings.neurips.cc/paper_files/paper/2024/file/9bae399d1f34b8650351c1bd3692aeae-Paper-Conference.pdf},
    volume = {37},
    year = {2024}
}

@article{park2017function,
    title={Function-Based Approach to Designing an Instructional Environment},
    author={Park, Kristy and Pinkelman, Sarah},
    journal={Beyond Behavior},
    volume={26},
    number={3},
    pages={124--134},
    year={2017},
    doi={10.1177/1074295617729813}
}

@inproceedings{zheng2023judging,
    author = {Zheng, Lianmin and Chiang, Wei-Lin and Sheng, Ying and Zhuang, Siyuan and Wu, Zhanghao and Zhuang, Yonghao and Lin, Zi and Li, Zhuohan and Li, Dacheng and Xing, Eric and Zhang, Hao and Gonzalez, Joseph E. and Stoica, Ion},
    booktitle = {Advances in Neural Information Processing Systems},
    doi = {10.52202/075280-2020},
    editor = {A. Oh and T. Naumann and A. Globerson and K. Saenko and M. Hardt and S. Levine},
    pages = {46595--46623},
    publisher = {Curran Associates Inc.},
    title = {Judging LLM-as-a-Judge with MT-Bench and Chatbot Arena},
    url = {https://proceedings.neurips.cc/paper_files/paper/2023/file/91f18a1287b398d378ef22505bf41832-Paper-Datasets_and_Benchmarks.pdf},
    volume = {36},
    year = {2023}
}

@inproceedings{hendrycks2021measuring,
    title={Measuring Massive Multitask Language Understanding},
    author={Dan Hendrycks and Collin Burns and Steven Basart and Andy Zou and Mantas Mazeika and Dawn Song and Jacob Steinhardt},
    booktitle={International Conference on Learning Representations},
    year={2021},
    url={https://openreview.net/forum?id=d7KBjmI3GmQ}
}

@inproceedings{lu2024generative,
    author = {Lu, Xinyi and Wang, Xu},
    title = {Generative Students: Using LLM-Simulated Student Profiles to Support Question Item Evaluation},
    year = {2024},
    isbn = {9798400706332},
    publisher = {Association for Computing Machinery},
    address = {New York, NY, USA},
    url = {https://doi.org/10.1145/3657604.3662031},
    doi = {10.1145/3657604.3662031},
    booktitle = {Proceedings of the Eleventh ACM Conference on Learning @ Scale},
    pages = {16--27},
    numpages = {12},
    location = {Atlanta, GA, USA},
    series = {L@S '24}
}

@article{qi2026simulating,
    title={Simulating student learning behaviors with LLM-based role-playing agents: A data-driven and cognitively inspired framework},
    author={Qi, Changyong and Zheng, Longwei and He, Anna and Xu, Haoxin and Jia, Linzhao and Wei, Yuang and Jiang, Bingqian and Gu, Xiaoqing},
    journal={Expert Systems with Applications},
    volume = {304},
    pages = {130753},
    year = {2026},
    issn = {0957-4174},
    doi = {https://doi.org/10.1016/j.eswa.2025.130753},
}

@inproceedings{jin2025teachtune,
    title = {TeachTune: Reviewing Pedagogical Agents Against Diverse Student Profiles with Simulated Students},
    author = {Jin, Hyoungwook and Yoo, Minju and Park, Jeongeon and Lee, Yokyung and Wang, Xu and Kim, Juho},
    year = {2025},
    isbn = {9798400713941},
    publisher = {Association for Computing Machinery},
    address = {New York, NY, USA},
    url = {https://doi.org/10.1145/3706598.3714054},
    doi = {10.1145/3706598.3714054},
    booktitle = {Proceedings of the 2025 CHI Conference on Human Factors in Computing Systems},
    articleno = {1073},
    numpages = {28},
    location = {Yokohama, Japan},
    series = {CHI '25}
}

@article{marquez2026simulating,
    title = {Simulating students with large language models: A review of architecture, mechanisms, and role modeling in education with generative AI},
    journal = {Computer Science Review},
    volume = {62},
    pages = {101008},
    year = {2026},
    issn = {1574-0137},
    doi = {https://doi.org/10.1016/j.cosrev.2026.101008},
    url = {https://www.sciencedirect.com/science/article/pii/S1574013726001164},
    author = {Luis Marquez-Carpintero and Alberto Lopez-Sellers and Miguel Cazorla}
}

@inproceedings{scarlatos2026simulated,
    title = "Simulated Students in Tutoring Dialogues: Substance or Illusion?",
    author = "Scarlatos, Alexander and Lee, Jaewook  and Woodhead, Simon and Lan, Andrew",
    editor = "Liakata, Maria  and
      Moreira, Viviane P.  and
      Zhang, Jiajun  and
      Jurgens, David",
    booktitle = "Proceedings of the 64th Annual Meeting of the {A}ssociation for {C}omputational {L}inguistics (Volume 1: Long Papers)",
    month = jul,
    year = "2026",
    address = "San Diego, California, United States",
    publisher = "Association for Computational Linguistics",
    url = "https://aclanthology.org/2026.acl-long.1960/",
    doi = "10.18653/v1/2026.acl-long.1960",
    pages = "42349--42385",
    ISBN = "979-8-89176-390-6"
}

@inproceedings{wang2025coser,
    title={Co{SER}: Coordinating {LLM}-Based Persona Simulation of Established Roles},
    author={Xintao Wang and Heng Wang and Yifei Zhang and Xinfeng Yuan and Rui Xu and Jen-tse Huang and Siyu Yuan and Haoran Guo and Jiangjie Chen and Shuchang Zhou and Wei Wang and Yanghua Xiao},
    booktitle={Forty-second International Conference on Machine Learning},
    year={2025},
    url={https://openreview.net/forum?id=BOrR7YqKUt}
}

@inproceedings{zhang2019bertscore,
    title={BERTScore: Evaluating Text Generation with BERT},
    author={Tianyi Zhang and Varsha Kishore and Felix Wu and Kilian Q. Weinberger and Yoav Artzi},
    booktitle={International Conference on Learning Representations},
    year={2020},
    url={https://openreview.net/forum?id=SkeHuCVFDr}
}

@inproceedings{tack2022ai,
    title = {The {AI} Teacher Test: Measuring the Pedagogical Ability of Blender and {GPT-3} in Educational Dialogues},
    address = {Durham, United Kingdom},
    author = {Tack, Ana{\"\i}s and Piech, Chris},
    booktitle = {Proceedings of the 15th International Conference on Educational Data Mining},
    doi = {10.5281/zenodo.6853187},
    editor = {Antonija Mitrovic and Nigel Bosch},
    isbn = {978-1-7336736-3-1},
    month = {July},
    pages = {522--529},
    publisher = {International Educational Data Mining Society},
    year = {2022}
}

@misc{google2025gemini3flash,
    author       = {{Google DeepMind}},
    title        = {{Gemini 3 Flash Model Card}},
    year         = {2025},
    url = {https://storage.googleapis.com/deepmind-media/Model-Cards/Gemini-3-Flash-Model-Card.pdf},
    note         = {Accessed: 2026-10-12}
}

@misc{google2026gemini31pro,
    author       = {{Google DeepMind}},
    title        = {{Gemini 3.1 Pro Model Card}},
    year         = {2026},
    url = {https://storage.googleapis.com/deepmind-media/Model-Cards/Gemini-3-1-Pro-Model-Card.pdf},
    note         = {Accessed: 2026-10-12}
}

@article{theodoridou2013functional,
    author = {Zoe Theodoridou and Athanasios Koutsoklenis},
    title ={Functional Behavioral Assessment for a Boy With Duchenne Muscular Dystrophy and Problem Behavior: A Case Study From Greece},
    journal = {Assessment for Effective Intervention},
    volume = {39},
    number = {1},
    pages = {54--64},
    year = {2013},
    doi = {10.1177/1534508413499771}
}

@incollection{pintrich2000role,
    title = {Chapter 14 - The Role of Goal Orientation in Self-Regulated Learning},
    editor = {Monique Boekaerts and Paul R. Pintrich and Moshe Zeidner},
    booktitle = {Handbook of Self-Regulation},
    publisher = {Academic Press},
    address = {San Diego},
    pages = {451-502},
    year = {2000},
    isbn = {978-0-12-109890-2},
    doi = {https://doi.org/10.1016/B978-012109890-2/50043-3},
    url = {https://www.sciencedirect.com/science/article/pii/B9780121098902500433},
    author = {Paul R. Pintrich}
}

@ARTICLE{panadero2017review,
    AUTHOR={Panadero, Ernesto},
    TITLE={A Review of Self-regulated Learning: Six Models and Four Directions for Research},
    JOURNAL={Frontiers in Psychology},
    VOLUME={8},
    YEAR={2017},
    URL={https://www.frontiersin.org/journals/psychology/articles/10.3389/fpsyg.2017.00422},
    DOI={10.3389/fpsyg.2017.00422},
    ISSN={1664-1078}
}

@article{sander2010qualitative,
    author = {Janay B. Sander and Jill D. Sharkey and Roger Olivarri and Diane A. Tanigawa and Tory Mauseth},
    title = {A Qualitative Study of Juvenile Offenders, Student Engagement, and Interpersonal Relationships: Implications for Research Directions and Preventionist Approaches},
    journal = {Journal of Educational and Psychological Consultation},
    volume = {20},
    number = {4},
    pages = {288--315},
    year = {2010},
    publisher = {Routledge},
    doi = {10.1080/10474412.2010.522878}
}

@article{lichtinger2015employing,
    title={Employing a case study approach to capture motivation and self-regulation of young students with learning disabilities in authentic educational contexts},
    author={Lichtinger, Einat and Kaplan, Avi},
    journal={Metacognition and Learning},
    volume={10},
    number={1},
    pages={119--149},
    year={2015},
    publisher={Springer},
    doi = {https://doi.org/10.1007/s11409-014-9131-1}
}

@mastersthesis{newman2012home,
    title={Home/Hospital Instruction: Instructional Approach to Working with Students with Major Depressive Disorder},
    author={Newman, Cara},
    school={Dominican University of California},
    year={2012},
    publisher={ERIC},
    url={https://eric.ed.gov/?id=ED531809}
}

@article{lack2014low,
    title={Low- and High-Achieving Sixth-Grade Students’ Access to Participation during Mathematics Discourse},
    author={Lack, Brian and Swars, Susan Lee and Meyers, Barbara},
    journal={The Elementary School Journal},
    volume={115},
    number={1},
    pages={97--123},
    year={2014},
    publisher={The University of Chicago Press},
    doi={10.1086/676947}
}

@article{collett2023relationships,
    author = {Jennifer Collett},
    title = {Relationships between Reading Identities and Dialogue: Perspectives from Fifth-grade Multilingual Learners},
    journal = {Reading \& Writing Quarterly},
    volume = {39},
    number = {4},
    pages = {259--276},
    year = {2023},
    publisher = {Routledge},
    doi = {10.1080/10573569.2022.2103053}
}

@article{alghamdi2016supporting,
  title={Supporting low-achieving EFL Learners: Expectations, procedure and significance of remedial sessions at a Saudi university},
  author={Alghamdi, Fatimah M A and Siddiqui, Ozma},
  journal={Journal of Education and Training Studies},
  volume={4},
  number={12},
  pages={204--212},
  year={2016},
  publisher={Redfame Publishing},
  doi={https://doi.org/10.11114/jets.v4i12.2028}
}

@article{cassidy2018philosophy,
  title={Philosophy with children, self-regulation and engaged participation for children with emotional-behavioural and social communication needs},
  author={Cassidy, Claire and Marwick, Helen and Deeney, Lynn and McLean, Gillian},
  journal={Emotional and Behavioural Difficulties},
  volume={23},
  number={1},
  pages={81--96},
  year={2018},
  publisher = {Routledge},
  doi={10.1080/13632752.2017.1388654},
  url={https://doi.org/10.1080/13632752.2017.1388654}
}

@misc{openai2024embedding,
  author       = {{OpenAI}},
  title        = {{New embedding models and API updates}},
  year         = {2024},
  url = {https://openai.com/index/new-embedding-models-and-api-updates/},
  note         = {Accessed: 2026-10-12}
}

@article{macmath2010research,
    author = {MacMath, Sheryl and Roberts, Jillian and Wallace, John and Chi, Xiaohong},
    title = {RESEARCH SECTION: Curriculum integration and at-risk students: a Canadian case study examining student learning and motivation},
    journal = {British Journal of Special Education},
    volume = {37},
    number = {2},
    pages = {87--94},
    doi = {https://doi.org/10.1111/j.1467-8578.2009.00454.x},
    year = {2010}
}

@article{bruno2024strategies,
  title={Strategies for solving multiplicative problems using a conceptual model-based problem-solving approach. A case study with a student with autism spectrum disorder},
  author={Bruno, Alicia and Polo-Blanco, Irene and Van Vaerenbergh, Steven and Fern{\'a}ndez-Cobos, Ra{\'u}l and Gonz{\'a}lez-L{\'o}pez, Mar{\'\i}a Jos{\'e}},
  journal={ZDM Mathematics Education},
  doi = {https://doi.org/10.1007/s11858-024-01568-w},
  volume={56},
  pages={1239--1256},
  year={2024},
  publisher={Springer}
}

@article{adjapong2015rethinking,
  title={Rethinking Pedagogy in Urban Spaces: Implementing Hip-Hop Pedagogy in the Urban Science Classroom},
  author={Adjapong, Edmund S and Emdin, Christopher},
  journal={Journal of Urban Learning, Teaching, and Research},
  volume={11},
  pages={66--77},
  year={2015},
  publisher={ERIC},
  url={https://eric.ed.gov/?id=EJ1071416}
}

@article{guardino2010changing,
  title={Changing Behaviors by Changing the Classroom Environment},
  author={Guardino, Caroline A and Fullerton, Elizabeth},
  journal = {TEACHING Exceptional Children},
  volume={42},
  number={6},
  pages={8--13},
  year={2010},
  doi={10.1177/004005991004200601},
  url={https://doi.org/10.1177/004005991004200601}
}

@misc{openai2025gpt41,
  author       = {{OpenAI}},
  title        = {{Introducing GPT‑4.1 in the API}},
  year         = {2025},
  url = {https://openai.com/index/gpt-4-1/},
  note         = {Accessed: 2026-10-12}
}

@book{achenbach1991child,
  title={Manual for the child behavior checklist/4-18 and 1991 profile},
  author={Achenbach, Thomas M},
  year={1991},
  publisher={University of Vermont, Department of Psychiatry}
}

@misc{anthropic2026claudeopus47,
  author       = {Anthropic},
  title        = {{System Card: Claude Opus 4.7}},
  year         = {2026},
  url = {https://anthropic.com/claude-opus-4-7-system-card},
  note         = {Accessed: 2026-10-12}
}

@misc{anthropic2025claudehaiku45,
  author       = {Anthropic},
  title        = {{System Card: Claude Haiku 4.5}},
  year         = {2025},
  url = {https://anthropic.com/claude-haiku-4-5-system-card},
  note         = {Accessed: 2026-10-12}
}

@misc{openai2026gpt55,
  author       = {{OpenAI}},
  title        = {{GPT-5.5 System Card}},
  year         = {2026},
  url = {https://deploymentsafety.openai.com/gpt-5-5/gpt-5-5.pdf},
  note         = {Accessed: 2026-10-12}
}

@misc{openai2026gpt54mini,
  author       = {{OpenAI}},
  title        = {{Introducing GPT‑5.4 mini and nano}},
  year         = {2026},
  url = {https://openai.com/index/introducing-gpt-5-4-mini-and-nano/},
  note         = {Accessed: 2026-10-12}
}

@misc{xu2026deepseek,
      title={DeepSeek-V4: Towards Highly Efficient Million-Token Context Intelligence}, 
      author={DeepSeek-AI and Anyi Xu and Bangcai Lin and Bing Xue and Bingxuan Wang and Bingzheng Xu and Bochao Wu and Bowei Zhang and Chaofan Lin and Chen Dong and Chenchen Ling and Chengda Lu and Chenggang Zhao and Chengqi Deng and Chengyu Hou and Chenhao Xu and Chenze Shao and Chong Ruan and Conner Sun and Damai Dai and Daya Guo and Dejian Yang and Deli Chen and Donghao Li and Dongjie Ji and Erhang Li and Fang Wei and Fangyun Lin and Fangzhou Yuan and Feiyu Xia and Fucong Dai and Guangbo Hao and Guanting Chen and Guoai Cao and Guolai Meng and Guowei Li and Han Yu and Han Zhang and Hanwei Xu and Hao Li and Haofen Liang and Haoling Zhang and Haoming Luo and Haoran Wei and Haotian Yuan and Haowei Zhang and Haowen Luo and Haoyu Chen and Haozhe Ji and Hengqing Zhang and Honghui Ding and Hongxuan Tang and Huanqi Cao and Huazuo Gao and Hui Qu and Hui Zeng and J Yang and JQ Zhu and Jia Luo and Jia Song and Jia Yu and Jialiang Huang and Jialu Cai and Jian Liang and Jiangting Zhou and Jiasheng Ye and Jiashi Li and Jiaxin Xu and Jiewen Hu and Jieyu Yang and Jin Chen and Jin Yan and Jingchang Chen and Jingli Zhou and Jingting Xiang and Jingyang Yuan and Jingyuan Cheng and Jingzi Zhou and Jinhua Zhu and Jiping Yu and Joseph Sun and Jun Ran and Junguang Jiang and Junjie Qiu and Junlong Li and Junmin Zheng and Junxiao Song and Kai Dong and Kaige Gao and Kang Guan and Kexing Zhou and Kezhao Huang and Kuai Yu and Lean Wang and Lecong Zhang and Lei Wang and Leyi Xia and Li Zhang and Liang Zhao and Lihua Guo and Lingxiao Luo and Linwang Ma and Linyan Zhu and Litong Wang and Liyu Cai and Liyue Zhang and Longhao Chen and MS Di and MY Xu and Max Mei and Miaojun Wang and Mingchuan Zhang and Minghua Zhang and Minghui Tang and Mingming Li and Mingxu Zhou and Minmin Han and Ning Wang and Panpan Huang and Panpan Wang and Peixin Cong and Peiyi Wang and Peng Zhang and Qiancheng Wang and Qihao Zhu and Qingyang Li and Qinyu Chen and Qiushi Du and Qiwei Jiang and Rui Tian and Ruifan Xu and Ruijie Lu and Ruiling Xu and Ruiqi Ge and Ruisong Zhang and Ruizhe Pan and Runji Wang and Runqian Chen and Runqiu Yin and Runxin Xu and Ruomeng Shen and Ruoyu Zhang and Ruyi Chen and SH Liu and Shanghao Lu and Shangmian Sun and Shangyan Zhou and Shanhuang Chen and Shaofei Cai and Shaoheng Nie and Shaoqing Wu and Shaoyuan Chen and Shengding Hu and Shengyu Liu and Shiqiang Hu and Shirong Ma and Shiyu Wang and Shuiping Yu and Shunfeng Zhou and Shuting Pan and Shuying Yu and Songyang Zhou and Tao Ni and Tao Yun and Tian Jin and Tian Pei and Tian Ye and Tianle Lin and Tianran Ji and Tianyi Cui and Tianyuan Yue and Tingting Yu and Tun Wang and W Zhang and WL Xiao and Wangding Zeng and Wei An and Weilin Zhao and Wen Liu and Wenfeng Liang and Wenjie Pang and Wenjing Luo and Wenjing Yao and Wenjun Gao and Wenkai Yang and Wenlve Huang and Wenqing Hou and Wentao Zhang and Wenting Ma and Xi Gao and Xiang He and Xiangwen Wang and Xianzu Wang and Xiao Bi and Xiaodong Liu and Xiaohan Wang and Xiaokang Chen and Xiaokang Zhang and Xiaotao Nie and Xiaowen Sun and Xiaoxiang Wang and Xin Cheng and Xin Liu and Xin Xie and Xingchao Liu and Xingchen Liu and Xingkai Yu and Xingyou Li and Xinyu Yang and Xinyu Zhang and Xu Chen and Xuanyu Wang and Xuecheng Su and Xueyin Chen and Xuheng Lin and Xuwei Fu and YC Yan and YQ Wang and YW Ma and Yanfeng Luo and Yang Zhang and Yanhong Xu and Yanru Ma and Yanwen Huang and Yao Li and Yao Li and Yao Xu and Yao Zhao and Yaofeng Sun and Yaohui Wang and Yi Qian and Yi Shao and Yi Yu and Yichao Zhang and Yifan Ding and Yifan Shi and Yijia Wu and Yiliang Xiong and Yiling Ma and Ying He and Ying Tang and Ying Zhou and Yingjia Luo and Yinmin Zhong and Yishi Piao and Yisong Wang and Yixiang Zhang and Yixiao Chen and Yixuan Tan and Yixuan Wei and Yiyang Ma and Yiyuan Liu and Yonglun Yang and Yongqiang Guo and Yongtong Wu and Yu Wu and YuKun Li and Yuan Cheng and Yuan Ou and Yuanfan Xu and Yuanhao Li and Yuduan Wang and Yuehan Yang and Yuer Xu and Yuhan Wu and Yuhao Meng and Yuheng Zou and Yukun Zha and Yunfan Xiong and Yupeng Chen and Yuping Lin and Yuqian Cao and Yuqian Wang and Yushun Zhang and Yuting Yan and Yutong Lin and Yuxian Gu and Yuxiang Luo and Yuxiang You and Yuxuan Liu and Yuxuan Zhou and Yuyang Zhou and Yuzhen Huang and ZF Wu and Zehao Wang and Zehua Zhao and Zehui Ren and Zekai Zhang and Zhangli Sha and Zhe Fu and Zhe Ju and Zhean Xu and Zhenda Xie and Zhengyan Zhang and Zheren Gao and Zhewen Hao and Zhibin Gou and Zhicheng Ma and Zhigang Yan and Zhihong Shao and Zhixian Huang and Zhixuan Chen and Zhiyu Wu and Zhizhou Ren and Zhongyu Wu and Zhuoshu Li and Zhuping Zhang and Zian Xu and Zihao Wang and Zihua Qu and Zihui Gu and Zijia Zhu and Zilin Li and Zipeng Zhang and Ziwei Xie and Ziyi Gao and Ziyi Wan and Zizheng Pan and Zongqing Yao},
      year={2026},
      eprint={2606.19348},
      archivePrefix={arXiv},
      primaryClass={cs.CL},
      url={https://arxiv.org/abs/2606.19348}, 
}

@misc{google2026gemma4,
  author       = {{Google DeepMind}},
  title        = {{Gemma 4 model card}},
  year         = {2026},
  url = {https://ai.google.dev/gemma/docs/core/model_card_4},
  note         = {Accessed: 2026-10-12}
}

@misc{openai2025gptoss,
  author       = {{OpenAI}},
  title        = {{gpt-oss-120b \& gpt-oss-20b Model Card}},
  year         = {2025},
  url = {https://deploymentsafety.openai.com/gpt-oss},
  note         = {Accessed: 2026-10-12}
}

@misc{meta2025llama4,
  author       = {{Meta}},
  title        = {{The Llama 4 herd: The beginning of a new era of natively multimodal AI innovation}},
  year         = {2025},
  url = {https://ai.meta.com/blog/llama-4-multimodal-intelligence/},
  note         = {Accessed: 2026-10-12}
}

@misc{qwen3.5,
    title  = {{Qwen3.5}: Towards Native Multimodal Agents},
    author = {{Qwen Team}},
    month  = {February},
    year   = {2026},
    url    = {https://qwen.ai/blog?id=qwen3.5}
}

@misc{anthropic2026claudesonnet5.5,
  author       = {Anthropic},
  title        = {{System Card: Claude Sonnet 5.5}},
  year         = {2026},
  url = {https://anthropic.com/claude-sonnet-5-5-system-card},
  note         = {Accessed: 2026-10-12}
}

@inproceedings{macina-etal-2023-mathdial,
    title = "{M}ath{D}ial: A Dialogue Tutoring Dataset with Rich Pedagogical Properties Grounded in Math Reasoning Problems",
    author = "Macina, Jakub  and
      Daheim, Nico  and
      Chowdhury, Sankalan  and
      Sinha, Tanmay  and
      Kapur, Manu  and
      Gurevych, Iryna  and
      Sachan, Mrinmaya",
    editor = "Bouamor, Houda  and
      Pino, Juan  and
      Bali, Kalika",
    booktitle = "Findings of the {A}ssociation for {C}omputational {L}inguistics: {EMNLP} 2023",
    month = dec,
    year = "2023",
    address = "Singapore",
    publisher = "Association for Computational Linguistics",
    url = "https://aclanthology.org/2023.findings-emnlp.372/",
    doi = "10.18653/v1/2023.findings-emnlp.372",
    pages = "5602--5621"
}

@inproceedings{macina-etal-2025-mathtutorbench,
    title = "{M}ath{T}utor{B}ench: A Benchmark for Measuring Open-ended Pedagogical Capabilities of {LLM} Tutors",
    author = "Macina, Jakub  and
      Daheim, Nico  and
      Hakimi, Ido  and
      Kapur, Manu  and
      Gurevych, Iryna  and
      Sachan, Mrinmaya",
    editor = "Christodoulopoulos, Christos  and
      Chakraborty, Tanmoy  and
      Rose, Carolyn  and
      Peng, Violet",
    booktitle = "Proceedings of the 2025 Conference on Empirical Methods in Natural Language Processing",
    month = nov,
    year = "2025",
    address = "Suzhou, China",
    publisher = "Association for Computational Linguistics",
    url = "https://aclanthology.org/2025.emnlp-main.11/",
    doi = "10.18653/v1/2025.emnlp-main.11",
    pages = "204--221",
    ISBN = "979-8-89176-332-6"
}

@inproceedings{sonkar-etal-2023-class,
    title = "{CLASS}: A Design Framework for Building Intelligent Tutoring Systems Based on Learning Science principles",
    author = "Sonkar, Shashank  and
      Liu, Naiming  and
      Mallick, Debshila  and
      Baraniuk, Richard",
    editor = "Bouamor, Houda  and
      Pino, Juan  and
      Bali, Kalika",
    booktitle = "Findings of the {A}ssociation for {C}omputational {L}inguistics: {EMNLP} 2023",
    month = dec,
    year = "2023",
    address = "Singapore",
    publisher = "Association for Computational Linguistics",
    url = "https://aclanthology.org/2023.findings-emnlp.130/",
    doi = "10.18653/v1/2023.findings-emnlp.130",
    pages = "1941--1961"
}

@inproceedings{liu-etal-2024-personality,
    title = "Personality-aware Student Simulation for Conversational Intelligent Tutoring Systems",
    author = "Liu, Zhengyuan  and
      Yin, Stella Xin  and
      Lin, Geyu  and
      Chen, Nancy F.",
    editor = "Al-Onaizan, Yaser  and
      Bansal, Mohit  and
      Chen, Yun-Nung",
    booktitle = "Proceedings of the 2024 Conference on Empirical Methods in Natural Language Processing",
    month = nov,
    year = "2024",
    address = "Miami, Florida, USA",
    publisher = "Association for Computational Linguistics",
    url = "https://aclanthology.org/2024.emnlp-main.37/",
    doi = "10.18653/v1/2024.emnlp-main.37",
    pages = "626--642"
}

@inproceedings{shridhar-etal-2022-automatic,
    title = "Automatic Generation of Socratic Subquestions for Teaching Math Word Problems",
    author = "Shridhar, Kumar  and
      Macina, Jakub  and
      El-Assady, Mennatallah  and
      Sinha, Tanmay  and
      Kapur, Manu  and
      Sachan, Mrinmaya",
    editor = "Goldberg, Yoav  and
      Kozareva, Zornitsa  and
      Zhang, Yue",
    booktitle = "Proceedings of the 2022 Conference on Empirical Methods in Natural Language Processing",
    month = dec,
    year = "2022",
    address = "Abu Dhabi, United Arab Emirates",
    publisher = "Association for Computational Linguistics",
    url = "https://aclanthology.org/2022.emnlp-main.277/",
    doi = "10.18653/v1/2022.emnlp-main.277",
    pages = "4136--4149"
}

@inproceedings{martynova-etal-2025-llms,
    title = "Can {LLM}s Effectively Simulate Human Learners? Teachers' Insights from Tutoring {LLM} Students",
    author = "Martynova, Daria  and
      Macina, Jakub  and
      Daheim, Nico  and
      Yalcin, Nilay  and
      Zhang, Xiaoyu  and
      Sachan, Mrinmaya",
    editor = {Kochmar, Ekaterina  and
      Alhafni, Bashar  and
      Bexte, Marie  and
      Burstein, Jill  and
      Horbach, Andrea  and
      Laarmann-Quante, Ronja  and
      Tack, Ana{\"i}s  and
      Yaneva, Victoria  and
      Yuan, Zheng},
    booktitle = "Proceedings of the 20th Workshop on Innovative Use of {NLP} for Building Educational Applications ({BEA} 2025)",
    month = jul,
    year = "2025",
    address = "Vienna, Austria",
    publisher = "Association for Computational Linguistics",
    url = "https://aclanthology.org/2025.bea-1.8/",
    doi = "10.18653/v1/2025.bea-1.8",
    pages = "100--117",
    ISBN = "979-8-89176-270-1"
}

@inproceedings{feng-etal-2020-doc2dial,
    title = "doc2dial: A Goal-Oriented Document-Grounded Dialogue Dataset",
    author = "Feng, Song  and
      Wan, Hui  and
      Gunasekara, Chulaka  and
      Patel, Siva  and
      Joshi, Sachindra  and
      Lastras, Luis",
    editor = "Webber, Bonnie  and
      Cohn, Trevor  and
      He, Yulan  and
      Liu, Yang",
    booktitle = "Proceedings of the 2020 Conference on Empirical Methods in Natural Language Processing ({EMNLP})",
    month = nov,
    year = "2020",
    address = "Online",
    publisher = "Association for Computational Linguistics",
    url = "https://aclanthology.org/2020.emnlp-main.652/",
    doi = "10.18653/v1/2020.emnlp-main.652",
    pages = "8118--8128"
}

@inproceedings{feng-etal-2021-multidoc2dial,
    title = "{M}ulti{D}oc2{D}ial: Modeling Dialogues Grounded in Multiple Documents",
    author = "Feng, Song  and
      Patel, Siva Sankalp  and
      Wan, Hui  and
      Joshi, Sachindra",
    editor = "Moens, Marie-Francine  and
      Huang, Xuanjing  and
      Specia, Lucia  and
      Yih, Scott Wen-tau",
    booktitle = "Proceedings of the 2021 Conference on Empirical Methods in Natural Language Processing",
    month = nov,
    year = "2021",
    address = "Online and Punta Cana, Dominican Republic",
    publisher = "Association for Computational Linguistics",
    url = "https://aclanthology.org/2021.emnlp-main.498/",
    doi = "10.18653/v1/2021.emnlp-main.498",
    pages = "6162--6176"
}

@inproceedings{wang-etal-2024-book2dial,
    title = "{B}ook2{D}ial: Generating Teacher Student Interactions from Textbooks for Cost-Effective Development of Educational Chatbots",
    author = "Wang, Junling  and
      Macina, Jakub  and
      Daheim, Nico  and
      Pal Chowdhury, Sankalan  and
      Sachan, Mrinmaya",
    editor = "Ku, Lun-Wei  and
      Martins, Andre  and
      Srikumar, Vivek",
    booktitle = "Findings of the {A}ssociation for {C}omputational {L}inguistics: {ACL} 2024",
    month = aug,
    year = "2024",
    address = "Bangkok, Thailand",
    publisher = "Association for Computational Linguistics",
    url = "https://aclanthology.org/2024.findings-acl.578/",
    doi = "10.18653/v1/2024.findings-acl.578",
    pages = "9707--9731"
}

@inproceedings{papineni-etal-2002-bleu,
    title = "{B}leu: a Method for Automatic Evaluation of Machine Translation",
    author = "Papineni, Kishore  and
      Roukos, Salim  and
      Ward, Todd  and
      Zhu, Wei-Jing",
    editor = "Isabelle, Pierre  and
      Charniak, Eugene  and
      Lin, Dekang",
    booktitle = "Proceedings of the 40th Annual Meeting of the {A}ssociation for {C}omputational {L}inguistics",
    month = jul,
    year = "2002",
    address = "Philadelphia, Pennsylvania, USA",
    publisher = "Association for Computational Linguistics",
    url = "https://aclanthology.org/P02-1040/",
    doi = "10.3115/1073083.1073135",
    pages = "311--318"
}

@inproceedings{wang-etal-2024-bridging,
    title = "Bridging the Novice-Expert Gap via Models of Decision-Making: A Case Study on Remediating Math Mistakes",
    author = "Wang, Rose  and
      Zhang, Qingyang  and
      Robinson, Carly  and
      Loeb, Susanna  and
      Demszky, Dorottya",
    editor = "Duh, Kevin  and
      Gomez, Helena  and
      Bethard, Steven",
    booktitle = "Proceedings of the 2024 Conference of the {N}orth {A}merican Chapter of the {A}ssociation for {C}omputational {L}inguistics: Human Language Technologies (Volume 1: Long Papers)",
    month = jun,
    year = "2024",
    address = "Mexico City, Mexico",
    publisher = "Association for Computational Linguistics",
    url = "https://aclanthology.org/2024.naacl-long.120/",
    doi = "10.18653/v1/2024.naacl-long.120",
    pages = "2174--2199"
}

@inproceedings{maurya-etal-2025-unifying,
    title = "Unifying {AI} Tutor Evaluation: An Evaluation Taxonomy for Pedagogical Ability Assessment of {LLM}-Powered {AI} Tutors",
    author = "Maurya, Kaushal Kumar  and
      Srivatsa, Kv Aditya  and
      Petukhova, Kseniia  and
      Kochmar, Ekaterina",
    editor = "Chiruzzo, Luis  and
      Ritter, Alan  and
      Wang, Lu",
    booktitle = "Proceedings of the 2025 Conference of the Nations of the {A}mericas Chapter of the {A}ssociation for {C}omputational {L}inguistics: Human Language Technologies (Volume 1: Long Papers)",
    month = apr,
    year = "2025",
    address = "Albuquerque, New Mexico",
    publisher = "Association for Computational Linguistics",
    url = "https://aclanthology.org/2025.naacl-long.57/",
    doi = "10.18653/v1/2025.naacl-long.57",
    pages = "1234--1251",
    ISBN = "979-8-89176-189-6"
}

\newpage
\appendix
\clearpage

\section{Model Fallbacks in the Pipeline} \label{app:model_fallback}

Throughout the benchmark construction pipeline, we prioritized the Gemini model specified in \S\ref{sec:data_construction} for consistency. Failed calls were retried up to five times before falling back to an OpenAI model. Here, we document these fallback cases.

In Stage 2 (Evidence Extraction), a small subset of papers ($N=7$, 2.3\% of 302 papers before subsequent filtering to 300) was reprocessed with \texttt{gpt-5.4-mini} due to output truncation ($N=5$), recitation blocking ($N=1$), or a schema error ($N=1$). In Stage 5 (Rubric Construction), a single student category ($N=1$ of 518, 0.2\%) was processed with \texttt{gpt-5.5} after the original Gemini call was blocked by its safety filter. Follow-up tests suggest that the block may have been related to the combination of the student's young age, disability, and descriptions of severe distress behaviors.

Given the small proportion of affected cases, we expect this fallback model use to have only a limited impact on the overall benchmark data.

\section{ERIC Search Descriptors} \label{app:eric-descriptors}

Table~\ref{tab:eric-descriptors} lists the ERIC descriptors used to retrieve candidate papers for dataset construction.

\begin{table}[h]
\centering
\caption{Descriptors used to retrieve candidate papers.}
\small
\begin{tabular}{@{}p{0.23\linewidth}p{0.70\linewidth}@{}}
\toprule
\textbf{Category} & \textbf{Descriptors} \\
\midrule
\multirow{35}{=}{\textbf{Student} \\ (35 Total)}
& Behavior Problems \\
& Behavior Disorders \\
& Self Destructive Behavior \\
& Aggression \\
& Emotional Problems \\
& Emotional Disturbances \\
& Anxiety Disorders \\
& Depression (Psychology) \\
& School Phobia \\
& Withdrawal (Psychology) \\
& Fear of Success \\
& Mental Disorders \\
& Personality Problems \\
& Psychosomatic Disorders \\
& Attention Deficit Disorders \\
& Attention Deficit Hyperactivity Disorder \\
& Autism Spectrum Disorders \\
& Developmental Disabilities \\
& Intellectual Disability \\
& Neurodevelopmental Disorders \\
& Neurological Impairments \\
& Developmental Delays \\
& Learning Problems \\
& Learning Disabilities \\
& Low Achievement \\
& Academic Failure \\
& Reading Difficulties \\
& Reading Failure \\
& Writing Difficulties \\
& Mathematics Anxiety \\
& Slow Learners \\
& At Risk Students \\
& Communication Problems \\
& Communication Disorders \\
& Language Impairments \\
\midrule
\multirow{4}{=}{\textbf{Methodology} \\ (4 Total)}
& Case Studies \\
& Observation \\
& Qualitative Research \\
& Classroom Research \\
\bottomrule
\end{tabular}
\label{tab:eric-descriptors}
\end{table}

\section{Abstract-Based Paper Filtering Procedure} \label{app:abstract_filtering}
The candidate papers are initially screened by \texttt{gpt-4.1-mini}~\cite{openai2025gpt41} using the title and abstract. We use this as a cost-efficient, high-recall first pass, since abstracts provide enough signal to remove clearly unfit papers while leaving ambiguous cases to later pipeline stages.

The filter applies four criteria, filtering out papers that fail to meet our benchmark requirements: (1) papers without direct student observation or interview, which lack grounded learner behavior; (2) quantitative studies, which report aggregate effects instead of the individual student responses we need; (3) papers focused on educational policy or measurement-tool development, which describe students only as targets of analysis; and (4) papers without practical guidance for teachers, which lack the prescriptive strategies needed for evaluation. A paper is retained only if rejected by none of the four. The full prompt is in Appendix~\ref{app:stage0_abstract}.

We additionally exclude excessively long papers ($\geq$ 30 pages) and those that cannot be accessed or downloaded, as we observed that long papers tend to mistakenly introduce literature-review content as learner evidence in downstream extraction.

\section{Dialogue-Based Taxonomy of Learner Challenges} \label{app:taxonomy}
To distinguish the challenge type associated with each student category, we designed a taxonomy that identifies the observable challenge a tutor must respond to in the tutoring interaction. The taxonomy consists of five categories: \textit{Task Resistance}, \textit{Emotional Distress}, \textit{Attention/Engagement Failure}, \textit{Social-Interpersonal Difficulty}, and \textit{Academic Struggle}. It was designed and refined by an author with a doctoral degree in education.

We use CBCL/ASEBA~\citep{achenbach1991child} as a reference for defining our categories. CBCL/ASEBA is an empirical framework that characterizes children's functioning through standardized behavior descriptions. Since our benchmark is dialogue-based, we retain only constructs observable in tutoring interactions and reorganize them by the type of challenge they present to the tutor.

Specifically, the Internalizing grouping is adapted as \textit{Emotional Distress}; Attention Problems as \textit{Attention/Engagement Failure}; Social Problems and Aggressive Behavior as \textit{Social-Interpersonal Difficulty}; and Externalizing behaviors directed at the learning task as \textit{Task Resistance}. \textit{Academic Struggle} adapts the School scale for content-level difficulty such as confusion, uncertainty, repeated errors, misconceptions, and failure to respond.

\section{Impact of Dialogue Turns on Evaluation Results} \label{app:turn_ablation}

We examine whether evaluation results depend on the six-turn dialogue setting used in our benchmark. For this analysis, we sampled 25\% of student categories ($N=130$), stratified by the five challenge types, and regenerated their dialogues with four and eight turns. We used an adapted version of the original dialogue-synthesis prompt that allows the number of turns to vary (prompt in Appendix~\ref{app:turn_ablation_prompt}). We then repeated the open-ended evaluation for each dialogue setting using the same rubrics.

Changing the number of dialogue turns had little effect on the evaluation results. Overall success rates were 45.8\%, 46.6\%, and 46.8\% for four-, six-, and eight-turn dialogues,  respectively. Model rankings also remained highly consistent, with Spearman correlation being $\rho=0.916$ ($p<0.001$) for the four- versus six-turn comparison and $\rho=0.911$ ($p<0.001$) for the six- versus eight-turn comparison. These results suggest that our findings are not sensitive to the number of dialogue turns.

\section{MCQ Construction Details} \label{app:mcq_construction}

\subsection{4-Option MCQs}

We provide additional details on distractor retrieval, selection, and option adaptation for the MCQ task.

We construct distractors in three steps: candidate retrieval, distractor selection, and surface adaptation. As a scenario may contain multiple effective strategies, each strategy is treated as a separate correct option and receives its own distractor set.

For retrieval, we embed each scenario's learning context, learner behavior, and intervention descriptions using \texttt{text-embedding-3-large}~\cite{openai2024embedding}. Each candidate intervention is scored as:
\begin{align*}
\text{score} = \text{ctx\_sim} \, &\times (1 - \text{behav\_sim}) \\
  &\times (1 - \text{max\_interv\_sim}),
\end{align*}
where \texttt{ctx\_sim} and \texttt{behav\_sim} are cosine similarities between the target and candidate learning contexts and learner behaviors, and \texttt{max\_interv\_sim} is the maximum similarity between the candidate intervention and any of the target's effective interventions. This favors candidates that fit the target's classroom context but address a different learner problem with a different intervention mechanism. We exclude strategies from the same paper or student category and keep only the highest-scoring intervention per scenario, ensuring that distractors come from distinct cases. The top 10 candidates are forwarded to the next step.

However, embedding similarity alone cannot reliably identify contextual incompatibilities or paraphrased versions of valid target interventions. Hence, we use \texttt{gemini-3.1-pro-preview}~\cite{google2026gemini31pro} as an LLM judge~\cite{zheng2023judging} to select three final distractors from the top 10 candidates, prioritizing plausible but incorrect tutor actions that are structurally compatible with the target setting (prompt in Appendix~\ref{app:mcq_distract}).

Finally, the same model rewrites all four options---including the correct one---into the target dialogue's vocabulary, length, and tone while preserving each option's underlying mechanism (prompt in Appendix~\ref{app:mcq_adapt}). This prevents models from identifying the correct answer through lexical or stylistic match with the dialogue. Options are shuffled with a fixed seed and labeled A--D, and each MCQ is presented with the dialogue and a prompt asking the model to choose the most appropriate intervention (see Appendix~\ref{app:mcq_response}).

\subsection{Binary MCQs}
We additionally construct a binary MCQ variant as a sanity check. Unlike the 4-option MCQs in \S\ref{sec:mcq}, where all options are effective strategies and the task is to identify the best fit, the binary variant presents one effective and one ineffective strategy from the target scenario. Since the ineffective option failed for the same student in the source case, this variant tests whether models can reject a clearly inappropriate response. High accuracy therefore helps rule out this basic failure mode and supports interpreting the 4-option task as the harder problem of choosing among plausible effective strategies.

\subsubsection{Construction} \label{app:binary_method}

For each scenario and response type (\textit{immediate} or \textit{long-term}), we create one item when the scenario contains at least one matching effective strategy and at least one ineffective strategy. When multiple options are available on either side, we use \texttt{gemini-3.1-pro-preview}~\cite{google2026gemini31pro} as an LLM judge~\cite{zheng2023judging} to select the final effective--ineffective option pair (prompt in Appendix~\ref{app:binary_distract}). The judge prioritizes pairs with matching scope, concrete tutor actions, and comparable granularity, where the effective option directly addresses the dialogue and the ineffective option is a plausible misstep rather than an unrelated action. If each side has only one eligible strategy, the pair is fixed, and the judge is skipped.

Since effective and ineffective strategies may differ in style in how they are written, the same model then rewrites the two options to match in vocabulary, length, and tone while preserving their meanings (prompt in Appendix~\ref{app:binary_adapt}). This step is constrained to make minimal edits, avoid unsupported additions, and keep the ineffective option substantively ineffective. The two options are then assigned to A and B using a fixed random seed. 

\subsubsection{Results}

\begin{figure}[h]
    \centering
    \includegraphics[width=\columnwidth]{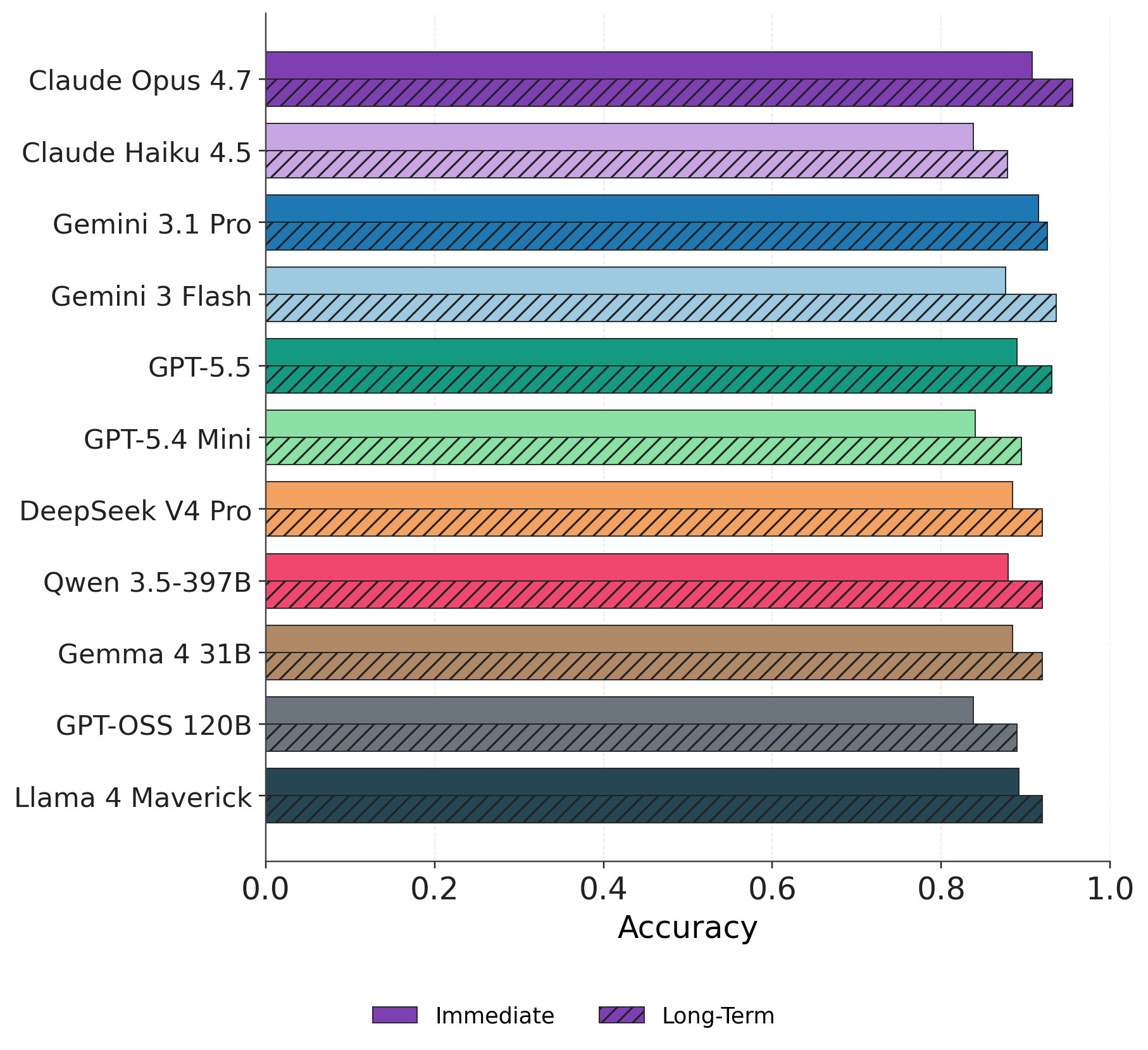}
    \caption{Binary MCQ accuracy across 11 models.}
    \label{fig:first_fullwidth}
\end{figure}

Across 11 models and 751 binary MCQs, accuracy ranges from 85.8\% to 93.1\%, which is far above the 50\% chance baseline (Figure~\ref{fig:first_fullwidth}). This suggests that models can reliably distinguish a documented effective intervention from an ineffective one when the contrast is explicit. Therefore, the lower performance in the 4-option MCQ setting is not due to a basic inability to reject ineffective responses, but to the harder task of selecting the best-fitting intervention among plausible alternatives.

\section{Student Demographic Information} \label{app:demographics}

\subsection{Demographic Distribution}
We summarize demographic information for 518 student categories from 300 source papers. During Stage 1 of our pipeline, the LLM extracted country and education level at the paper level, and age and gender at the student-category level. For analysis, we grouped countries into regions and ages into ranges. Country is reported for 99.8\% of categories, education level for all categories, gender for 74.9\%, and age for 60.4\%. We discuss the potential implications of these distributions in Limitations.

\paragraph{Geographic distribution.}
Most categories are from North America (61.4\%) and Europe (20.5\%), followed by Asia (5.8\%), the Middle East (5.4\%), Oceania (5.0\%), Latin America (1.4\%), and Africa (0.4\%). The United States accounts for 58.9\% of categories, followed by the United Kingdom (9.7\%), Australia (4.4\%), and Canada (2.5\%).

\paragraph{Education level.}
Primary-school students account for 41.9\% of categories, followed by lower secondary (20.5\%), upper secondary (19.3\%), early childhood (7.1\%), and tertiary education (5.4\%). The remaining 5.8\% come from settings outside a single school level, including special-education schools and adult or alternative programs.

\paragraph{Age.}
Students aged 6--11 account for 25.9\% of all categories, followed by ages 12--14 (13.5\%), 15--18 (11.8\%), 0--5 (6.8\%), and 19 or older (2.5\%).

\paragraph{Gender.}
Male students appear in 47.7\% of categories, female students in 24.3\%, and mixed-gender groups in 2.7\%, with a single category (0.2\%) describing a non-binary student.

\subsection{Performance by Demographic Groups}

Table~\ref{tab:demographic_performance} presents open-ended response success rates across demographic groups, averaged over 11 LLMs. Performance showed similar trends across groups, with immediate responses generally outperforming long-term interventions. Given the uneven group sizes, comparisons across demographic groups should be interpreted with caution.

\begin{table}[t]
\centering
\small
\caption{Open-ended success rates (\%) by geographic region, education level, and gender, averaged across 11 LLMs. $N$ denotes the number of student categories.}
\label{tab:demographic_performance}
\setlength{\tabcolsep}{2pt}
\begin{tabular*}{\columnwidth}{@{\extracolsep{\fill}}llrrrr}
\toprule
Group & Category & $N$ & Imm. & LT & Overall \\
\midrule
Overall & -- & 518 & 56.7 & 33.8 & 45.9 \\
\midrule
\multirow[t]{8}{*}{Region}
 & North America & 318 & 53.5 & 32.5 & 43.3 \\
 & Europe & 106 & 64.5 & 33.1 & 50.3 \\
 & Asia & 30 & 57.5 & 43.3 & 51.2 \\
 & Middle East & 28 & 64.5 & 38.8 & 52.9 \\
 & Oceania & 26 & 53.7 & 34.8 & 45.0 \\
 & Latin America & 7 & 45.5 & 31.4 & 38.7 \\
 & Africa & 2 & 51.5 & 81.8 & 63.6 \\
 & Not reported & 1 & 72.7 & 0.0 & 36.4 \\
\midrule
\multirow[t]{7}{*}{Education}
 & Early childhood & 37 & 48.8 & 29.8 & 40.5 \\
 & Primary & 217 & 57.5 & 32.0 & 45.2 \\
 & Lower secondary & 106 & 55.9 & 33.5 & 45.8 \\
 & Upper secondary & 100 & 57.4 & 37.9 & 47.8 \\
 & Secondary & 2 & 50.0 & 86.4 & 68.2 \\
 & Tertiary & 28 & 61.6 & 36.6 & 48.0 \\
 & Other/Unspecified & 28 & 59.7 & 33.1 & 49.3 \\
\midrule
\multirow[t]{5}{*}{Gender}
 & Male & 247 & 56.8 & 33.0 & 45.2 \\
 & Female & 126 & 54.3 & 33.8 & 45.0 \\
 & Mixed & 14 & 54.5 & 26.4 & 42.0 \\
 & Non-binary & 1 & 27.3 & 18.2 & 22.7 \\
 & Not reported & 130 & 58.9 & 36.1 & 48.5 \\
\bottomrule
\end{tabular*}
\end{table}

\section{Human Validation Interface} \label{app:validation_interface}

\begin{figure*}[h]
    \centering
    \includegraphics[width=\textwidth]{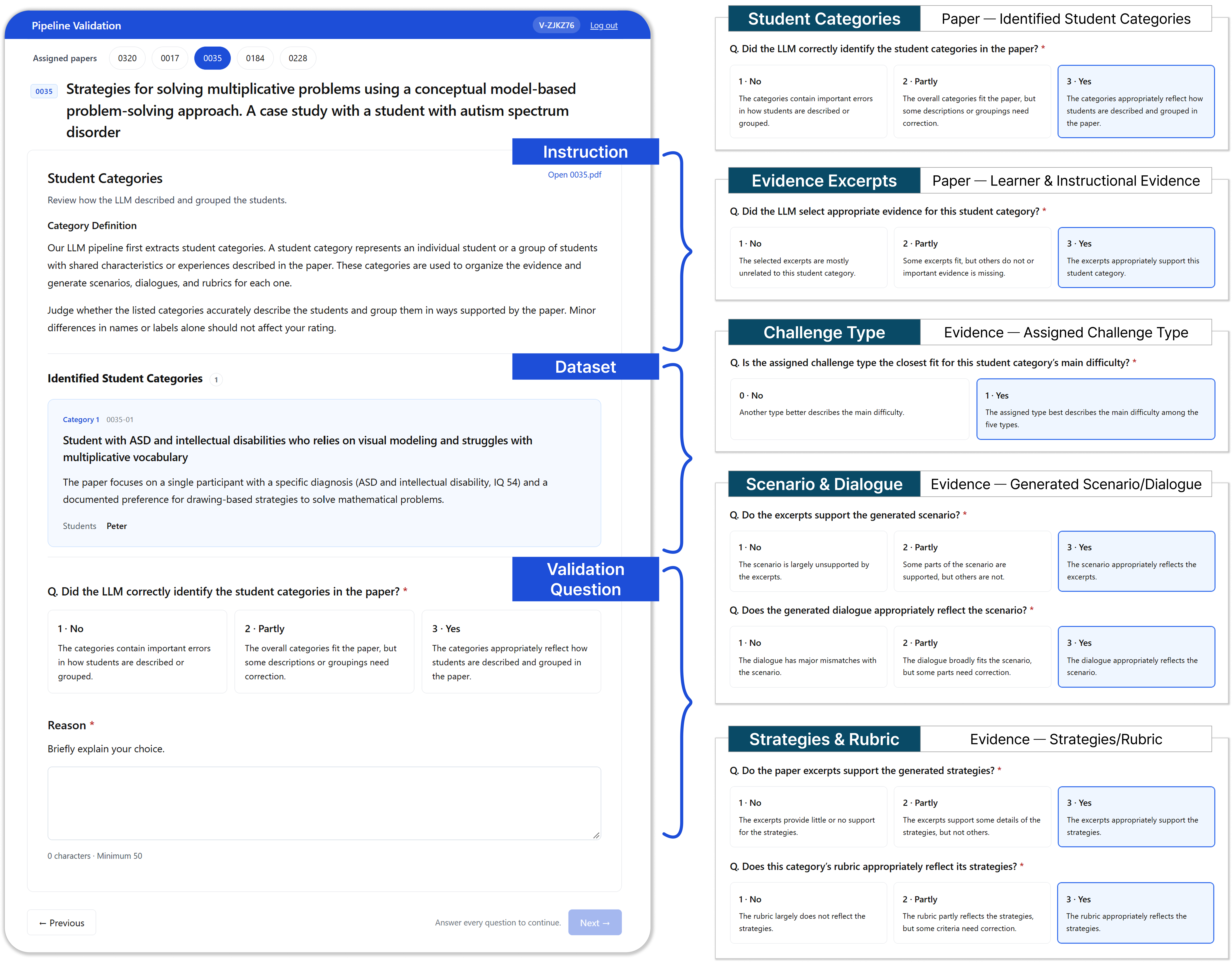}
    \caption{Human validation interface across five pipeline stages. The left panel shows an example evaluation screen, while the right summarizes validation questions. Example from~\citet{bruno2024strategies}.}
    \label{fig:interface}
\end{figure*}

Figure~\ref{fig:interface} shows the interface used for human validation. For each assigned paper, participants reviewed the outputs from each pipeline stage and rated how well they were supported by the source paper. At each step, the interface provided the evaluation criterion and its definition, the corresponding pipeline output, and a validation question with response options. Participants selected a rating and provided a written rationale for their judgment.

We designed the validation interface to encourage informed and deliberate judgments. For each assigned paper, participants were first required to read the source paper and write a summary of at least 30 words before validating its pipeline outputs. The interface also required a written rationale of at least 50 characters for each rating to encourage participants to articulate the reasoning behind their judgments and discourage arbitrary ratings.

\section{Further Details on Human Validation}
\label{app:human_validation}
Our human validation study covered 30 of the 300 source papers and 57 of the 518 student categories. Each paper was independently evaluated by two non-author reviewers across all pipeline stages. We present the distribution of human validation ratings and exact inter-rater agreement in Table~\ref{tab:human_validation_distribution}. 

\begin{table}[h]
\centering
\small
\caption{Human validation ratings and exact inter-rater agreement (\%). Challenge types used a binary (0/1), and other outputs used a three-point scale (1--3). Agreement indicates identical ratings between evaluators.}
\label{tab:human_validation_distribution}
\setlength{\tabcolsep}{4pt}
\begin{tabular*}{\columnwidth}{@{\extracolsep{\fill}}lrrrr}
\toprule
Output & Yes & Partly & No & Agreement \\
\midrule
Student Categories & 65.0 & 31.7 & 3.3 & 73.3 \\
Evidence Excerpts & 45.6 & 50.9 & 3.5 & 47.4 \\
Challenge Types & 76.3 & -- & 23.7 & 84.2 \\
Scenarios & 61.1 & 34.8 & 4.0 & 61.6 \\
Instructional Strategies & 50.5 & 47.5 & 2.0 & 43.4 \\
Dialogues & 65.7 & 28.3 & 6.1 & 55.6 \\
Rubrics & 69.3 & 30.7 & 0.0 & 59.6 \\
\midrule
Overall & 59.1 & 37.5 & 3.4 & 58.2 \\
\bottomrule
\end{tabular*}
\end{table}

Across 882 three-point judgments, 96.6\% were rated \textit{Yes} or \textit{Partly}, with only 3.4\% receiving \textit{No}. 76.3\% of 114 binary judgments also confirmed the assigned challenge types. Exact agreement between evaluators was 58.2\% overall. Agreement was highest for challenge types (84.2\%) and lowest for instructional strategies (43.4\%). While these results suggest that the generated outputs of our pipeline were generally well-grounded in the source papers, we further examine reviewers' rationales for both \textit{Partly} and \textit{No} judgments to understand the
concerns underlying lower ratings.

\paragraph{Student Categories (Stage 1).}
Most judgments of student categories were \textit{Yes} (65.0\%), suggesting that the extracted categories captured relevant students and their characteristics. Partial ratings concerned omitted students, overlapping categories, or imprecise descriptions. One reviewer noted, ``the student groupings are appropriate, but Jennifer's description includes a misattributed motivation.'' Reviewers also disagreed on whether students from the same study should be grouped together or separated based on their distinct characteristics. Overall, most concerns reflected differences in categorization and interpretation, with relatively few judgments identifying substantive grounding issues.

\paragraph{Evidence Excerpts (Stage 2).}
Evidence excerpts received the highest proportion of \textit{Partly} ratings (50.9\%). Reviewers raised concerns about the relevance and specificity of individual excerpts. One reviewer noted, ``Rather than being incorrect, some excerpts only partially describe the category or include passages that are too broad, making their relevance less clear.'' Reviewers also noted excerpts drawn from other students' cases or prior literature. Since reviewers evaluated entire evidence sets, their rationales for \textit{Partly} ratings often focused on issues with a few individual excerpts, while the remaining evidence was considered appropriate.

\paragraph{Challenge Types (Stage 3).}
Most judgments (76.3\%) confirmed the assigned challenge types, suggesting that they generally captured students' primary difficulties. However, reviewers noted that many students exhibited multiple, overlapping challenges that extended beyond the assigned type.

\paragraph{Scenarios (Stage 3).}
Most judgments (61.1\%) were \textit{Yes}, suggesting that scenarios were grounded in students' documented behaviors and contexts. Partial ratings often concerned combining evidence from different moments or contexts into a single situation. One reviewer noted, ``The scenario is largely supported, but it combines several related moments from Jimmy's case into one situation.'' Reviewers also identified unsupported contextual details or occasional misinterpretations of student behaviors. Overall, these concerns centered on preserving the original context when synthesizing documented observations into coherent scenarios.

\paragraph{Instructional Strategies (Stage 3).}
Most judgments were \textit{Yes} (50.5\%) or \textit{Partly} (47.5\%), suggesting that the extracted strategies were supported by the source papers. Partial ratings often concerned descriptions of individual strategies' contexts and outcomes. One reviewer noted, ``The actions are appropriate, but some context and outcomes seem to have been added to fit the scenario.'' Reviewers also identified general recommendations presented as interventions for specific students. These issues arose mainly when adapting documented strategies and their outcomes to individual scenarios.

\paragraph{Dialogues (Stage 4).}
Most judgments (65.7\%) were \textit{Yes}, suggesting that the generated dialogues reflected the intended scenarios and student behaviors. Partial ratings often concerned unnatural exchanges, even when the key elements of the scenarios were preserved. One reviewer noted, ``The main points are captured, but the lines feel more like separate monologues than a natural conversation.'' Reviewers also identified exaggerated student behaviors or tutor responses that were more directive than intended. The dialogues remained well-grounded in the scenarios, with reviewers' feedback primarily addressing conversational naturalness.

\paragraph{Rubrics (Stage 5).}
Most judgments (69.3\%) were \textit{Yes}, with no \textit{No} ratings, suggesting that rubrics well-reflected the extracted instructional strategies. Partial ratings often concerned overly specific requirements or rationales that extended beyond the source evidence. One reviewer noted, ``The six criteria align well with the strategies, but some of the rationales contain unsupported claims.'' Reviewers identified cases where context-specific findings were turned into broader rules. Overall, the rubrics captured the core instructional guidance, with some interpretation involved in translating the strategies into scenario-specific criteria.

\vspace{0.2cm}
Overall, human validation indicated that our pipeline outputs were well-grounded in the source papers. Reviewers' rationales also highlighted opportunities to improve how source evidence is interpreted and adapted into case-specific outputs, extending beyond grounding alone. Nevertheless, the evaluations supported the integrity of the resulting dataset, with most outputs meeting human reviewers' judgments and expectations.

\section{Models} \label{app:models}
We evaluate a diverse set of 11 state-of-the-art models on \dataset{}: six proprietary (\texttt{claude-opus-4.7}~\cite{anthropic2026claudeopus47}, \texttt{claude-haiku-4.5}~\cite{anthropic2025claudehaiku45}, \texttt{gemini-3.1-pro-preview}~\cite{google2026gemini31pro}, \texttt{gemini-3-flash-preview}~\cite{google2025gemini3flash}, \texttt{gpt-5.5}~\cite{openai2026gpt55}, \texttt{gpt-5.4-mini}~\cite{openai2026gpt54mini}) and five open-weight (\texttt{DeepSeek-V4-Pro}~\cite{xu2026deepseek}, \texttt{gemma-4-31B-it}~\cite{google2026gemma4}, \texttt{gpt-oss-120b}~\cite{openai2025gptoss}, \texttt{Llama-4-Maverick-17B-128E-Instruct}~\cite{meta2025llama4}, \texttt{Qwen3.5-397B-A17B}~\cite{qwen3.5}). 

We accessed each model through its respective deployed API. Claude models and Llama were served through Vertex AI\footnote{\url{https://docs.cloud.google.com/vertex-ai/docs}}, Gemini models and Gemma through the Google Gemini API\footnote{\url{https://ai.google.dev/gemini-api/docs}}, GPT models through the OpenAI API\footnote{\url{https://developers.openai.com/api/docs}}, DeepSeek and GPT-OSS through Together AI\footnote{\url{https://docs.together.ai/intro}}, and Qwen through OpenRouter\footnote{\url{https://openrouter.ai/docs/quickstart}}. In all settings, models were queried with their default API settings to reflect their standard deployed configurations.

\section{Average Criterion Satisfaction Rate} \label{app:normalized_score}
In addition to the binary success rate, we report the average criterion satisfaction rate, calculated as the number of satisfied criteria divided by the total number of criteria for each response.

\begin{table}[h]
\small
\centering
\caption{Average criterion satisfaction rate (\%)}
\label{tab:quant_open_criteria}
\begin{tabular}{@{}p{2.3cm}ccc@{}}
\toprule
Model & \makecell{Immediate\\\scriptsize($N{=}824$)} & \makecell{Long-Term\\\scriptsize($N{=}736$)} & \makecell{Overall\\\scriptsize($N{=}1{,}560$)} \\
\midrule
Claude Opus 4.7   & 35.8 & 20.9 & 28.8 \\
Claude Haiku 4.5  & 29.1 & 18.4 & 24.1 \\
Gemini 3.1 Pro    & 35.7 & 16.0 & 26.4 \\
Gemini 3 Flash    & 29.1 & 14.7 & 22.3 \\
GPT-5.5           & 35.7 & 19.4 & 28.0 \\
GPT-5.4 Mini      & 23.4 & 15.1 & 19.5 \\
\midrule
Llama 4 Maverick  & 18.9 &  8.2 & 13.9 \\
Gemma 4 31B       & 25.2 & 13.5 & 19.7 \\
DeepSeek V4 Pro   & 27.4 & 18.5 & 23.2 \\
GPT-OSS 120B      & 25.9 & 16.3 & 21.4 \\
Qwen 3.5-397B     & 24.5 & 12.0 & 18.6 \\
\bottomrule
\end{tabular}
\end{table}

Table~\ref{tab:quant_open_criteria} reports the average criterion satisfaction rates across the 11 models. Claude Opus 4.7 performs best overall (28.8\%), followed by GPT-5.5 (28.0\%) and Gemini 3.1 Pro (26.4\%). Consistent with the success rate evaluation, the same models lead overall, and all 11 models perform better on immediate responses than on long-term plans. These results indicate that the overall performance patterns are preserved under this alternative metric.

\begin{figure}[h]
    \centering
    \includegraphics[width=\columnwidth]{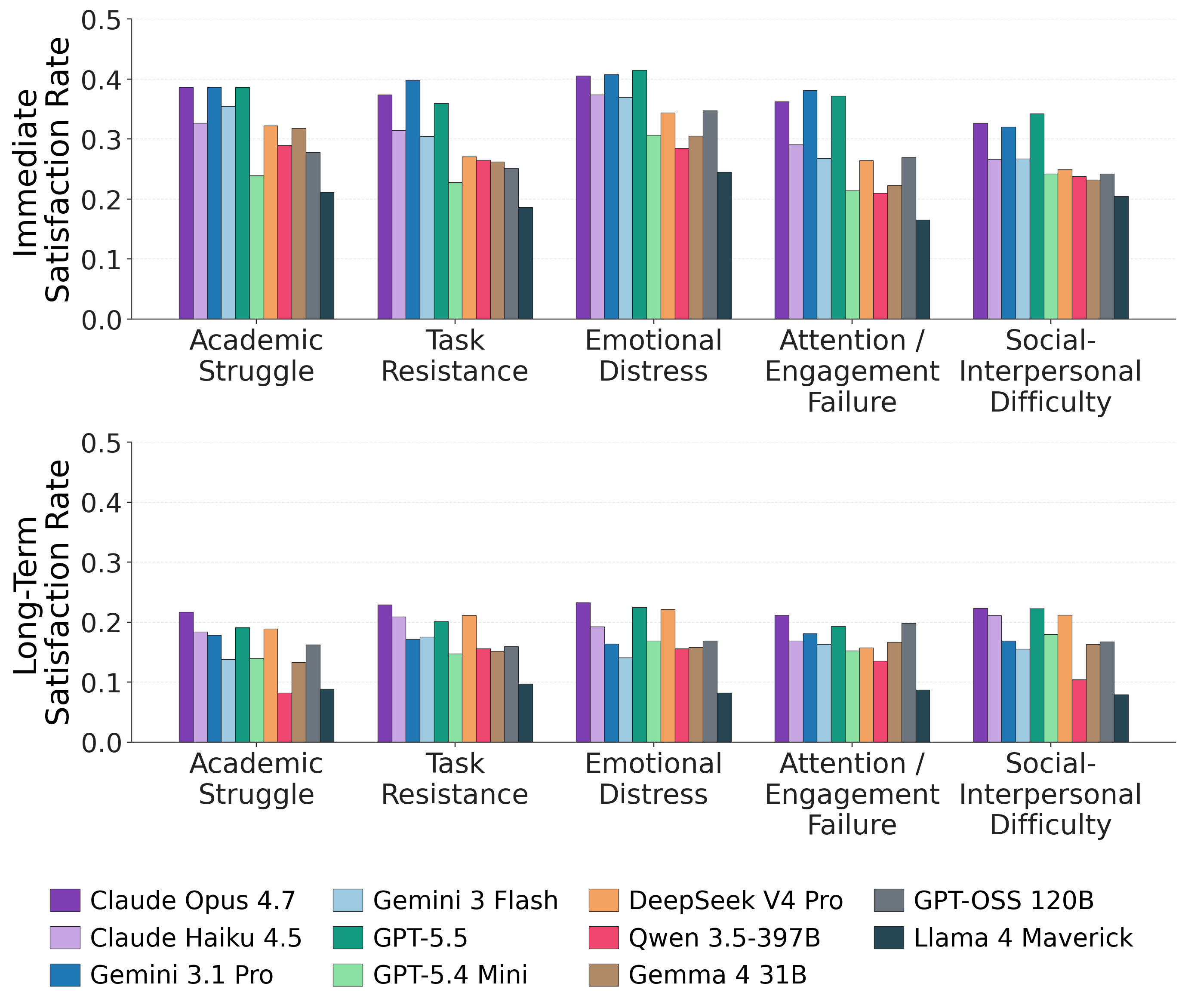}
    \caption{Average criterion satisfaction rates across models broken down by challenge type.}
    \label{fig:normalized_by_ct}
\end{figure}

Figure~\ref{fig:normalized_by_ct} further breaks down the results by challenge type. For immediate responses, task resistance and social-interpersonal difficulty generally show lower satisfaction rates, while emotional distress performs relatively well. For long-term interventions, academic struggle and social-interpersonal difficulty tend to show lower rates. Overall, social-interpersonal difficulty remains challenging across both settings, while the other lower-performing challenge types differ between immediate and long-term interventions.

\section{Impact of Construction Model Choice on Evaluation Results} \label{app:pipeline_ablation}
We examine whether evaluation results depend on the choice of models used for benchmark construction. Using a 25\% ($N=75$) stratified sample of source papers, we rerun the full construction and evaluation pipeline with GPT (\texttt{gpt-5.4-mini}, \texttt{gpt-5.5}) and Claude (\texttt{claude-sonnet-5.5}~\cite{anthropic2026claudesonnet5.5}, \texttt{claude-opus-4.7}) instead of Gemini, while keeping the prompts and the set of 11 evaluated models fixed. As each pipeline generates its own evaluation items, this yields 6{,}721 evaluated responses for GPT, 4{,}928 for Claude, and 3{,}850 for Gemini on the same source papers.

Model rankings remained highly consistent across construction models, with Spearman correlation being $\rho=0.943$ ($p<0.001$) for the Gemini- versus Claude-built comparison and $\rho=0.925$ ($p<0.001$) for the Gemini- versus GPT-built comparison. These results suggest that our findings are not sensitive to the choice of construction model.

\section{Impact of Judge Model Choice on Evaluation Results} \label{app:judge_ablation}
We examine whether the open-ended evaluation depends on the choice of judge model. Using the 25\% ($N=130$) stratified sample of student categories as in Appendix~\ref{app:turn_ablation}, we re-evaluate responses from all 11 models with \texttt{gpt-5.5} and \texttt{claude-opus-4.7}, while keeping the rubrics and evaluation prompt fixed. This yields 4{,}323 evaluated responses and 12{,}617 criterion judgments for each judge.

\begin{table}[h]
\centering
\caption{Pairwise agreement and ranking consistency across judges ($p<0.001$ for all Spearman correlations).}
\small
\setlength{\tabcolsep}{3pt}
\begin{tabular}{lccc}
\toprule
Judge Pair & \makecell{Case-level \\ $\kappa$} & \makecell{Criterion-level \\ $\kappa$} & \makecell{Spearman \\ $\rho$} \\
\midrule
Gemini vs. GPT & 0.75 & 0.77 & 0.96 \\
Gemini vs. Claude & 0.76 & 0.77 & 0.97 \\
GPT vs. Claude & 0.75 & 0.76 & 0.96 \\
\bottomrule
\end{tabular}
\label{tab:judge_robustness}
\end{table}

Table~\ref{tab:judge_robustness} shows high agreement across judges at both the case and criterion levels. Model rankings are also highly consistent across all judge pairs, with Spearman correlations of 0.96--0.97, all statistically significant at $p<0.001$. These results indicate that the relative performance of the evaluated models is  preserved across judge choices.

\section{Reference-Based Response Details} \label{app:reference_details}

We construct the reference pool using a paper-level split stratified by challenge type, targeting 15\% of papers for testing. This yields a \textit{pool} of 255 papers and 785 scenarios and a \textit{test} set of 45 papers and 138 scenarios, while ensuring that no test dialogue receives a reference from its source paper. We embed each scenario's challenging-behavior trigger and behavior patterns with \texttt{text-embedding-3-large}~\cite{openai2024embedding} and rank pool scenarios by cosine similarity. For each test dialogue, we retrieve the ten nearest pool scenarios, filter to those with an effective strategy for the target response type (immediate or long-term), and keep the three closest matches. Their matching effective and ineffective strategies are concatenated into a reference block, framed as examples from similar but different students that the responder may adapt, combine, or ignore (prompt in Appendix~\ref{app:reference_response}). We evaluate the resulting responses with the same rubrics used in the open-ended experiment, enabling a direct paired comparison against the dialogue-only condition.

\section{Lexical Diversity} \label{app:diversity}

\begin{figure}[h]
    \centering
    \includegraphics[width=0.9\columnwidth]{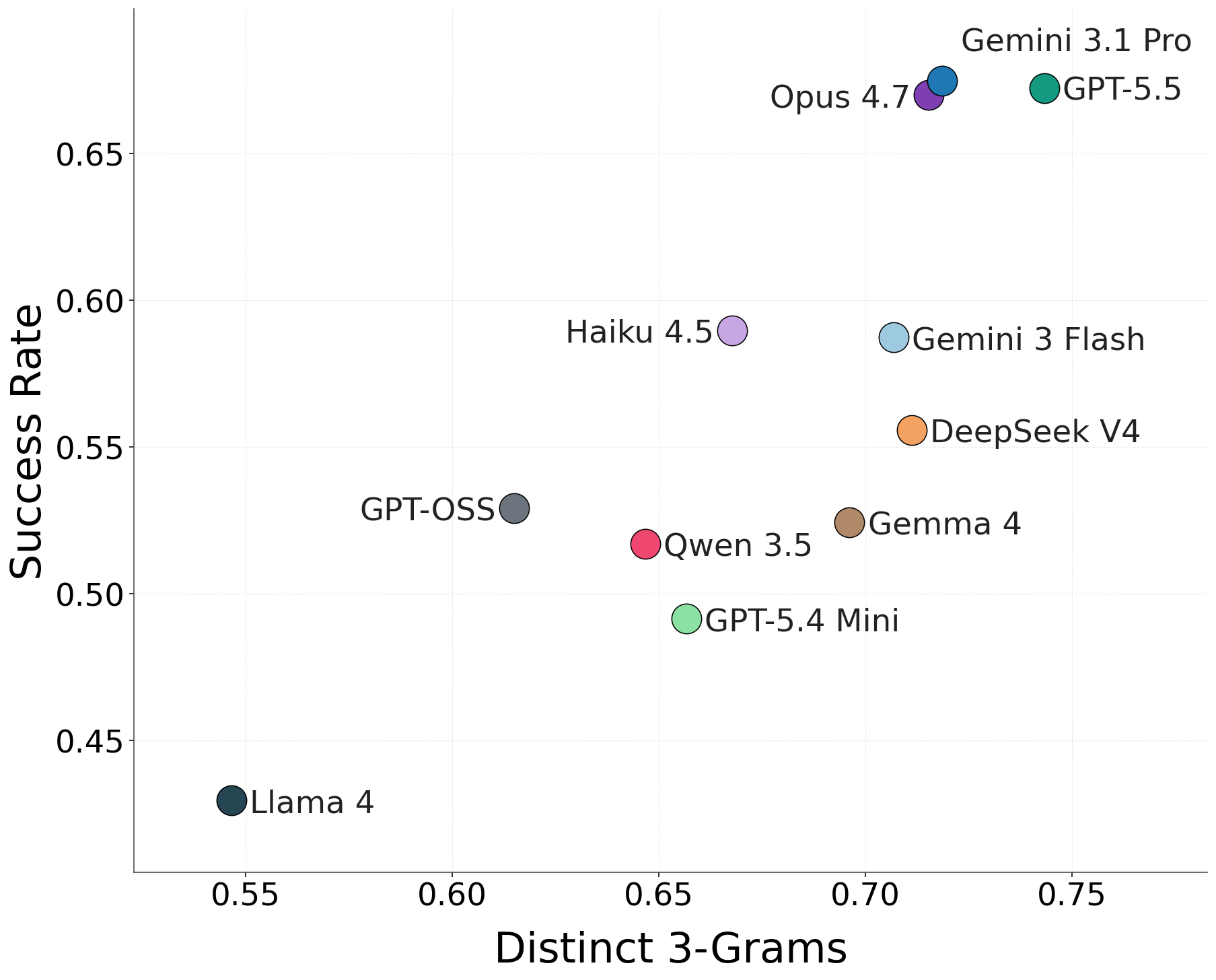}
    \caption{Correlation between distinct-$n$ ($n=3$) and immediate response success rate across models.}
    \label{fig:diversity_immediate_3gram}
\end{figure}

\begin{figure}[h]
    \centering
    \includegraphics[width=0.9\columnwidth]{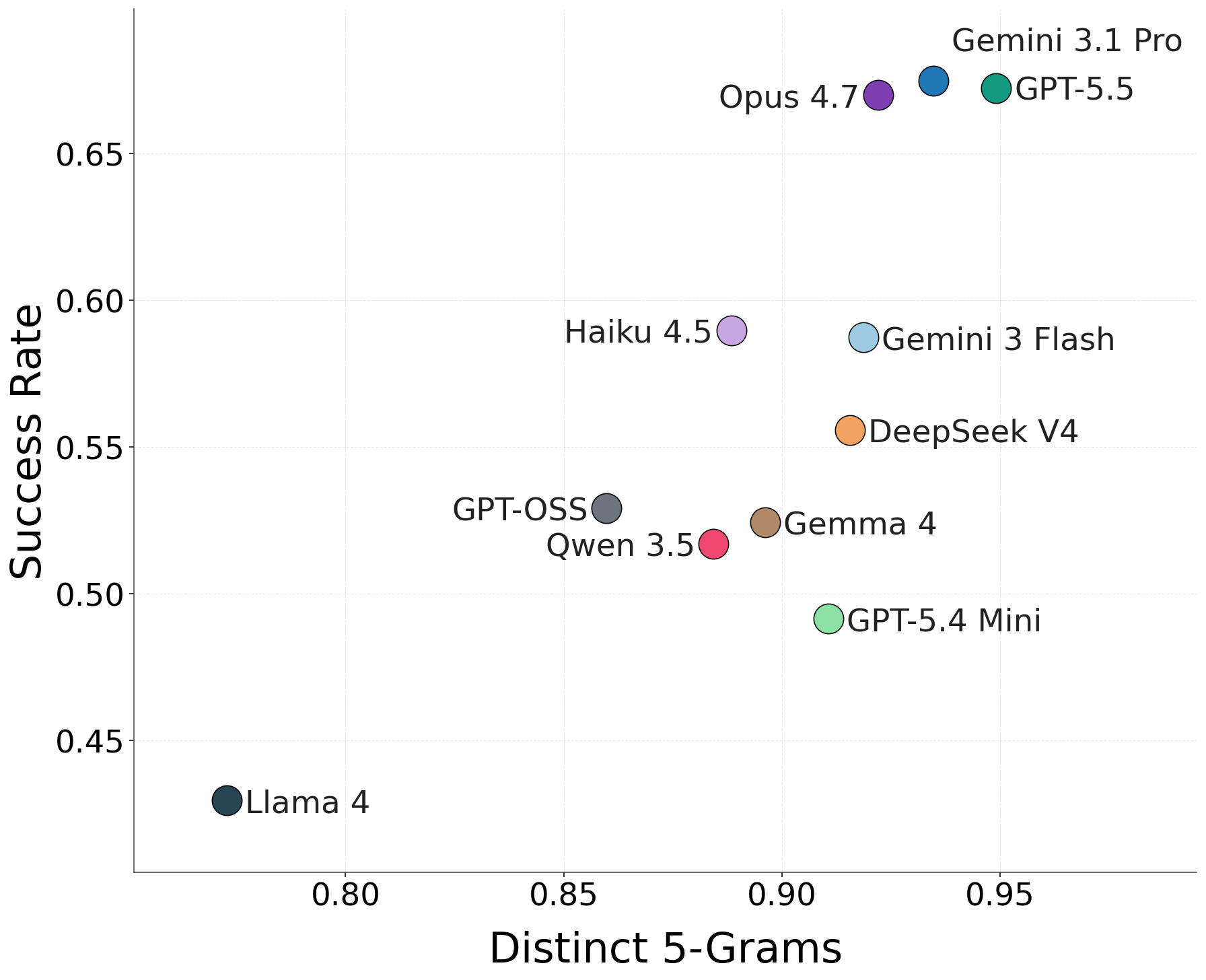}
    \caption{Correlation between distinct-$n$ ($n=5$) and immediate response success rate across models.}
    \label{fig:diversity_immediate_5gram}
\end{figure}

Figures~\ref{fig:diversity_immediate_3gram} and~\ref{fig:diversity_immediate_5gram} show the 3- and 5-gram variants of the distinct-$n$ analysis for immediate responses. Spearman correlations show the same pattern as the 4-gram result (Figure~\ref{fig:diversity_immediate}), with $\rho=0.855$ for 3-grams and $\rho=0.791$ for 5-grams (both $p<0.01$), supporting the association between higher lexical diversity and immediate-response success rates.

For long-term interventions, only the 3-gram correlation is significant ($\rho=0.627$, $p<0.05$).

\section{Case Specificity} \label{app:specificity}

\begin{figure}[h]
    \centering
    \includegraphics[width=0.85\columnwidth]{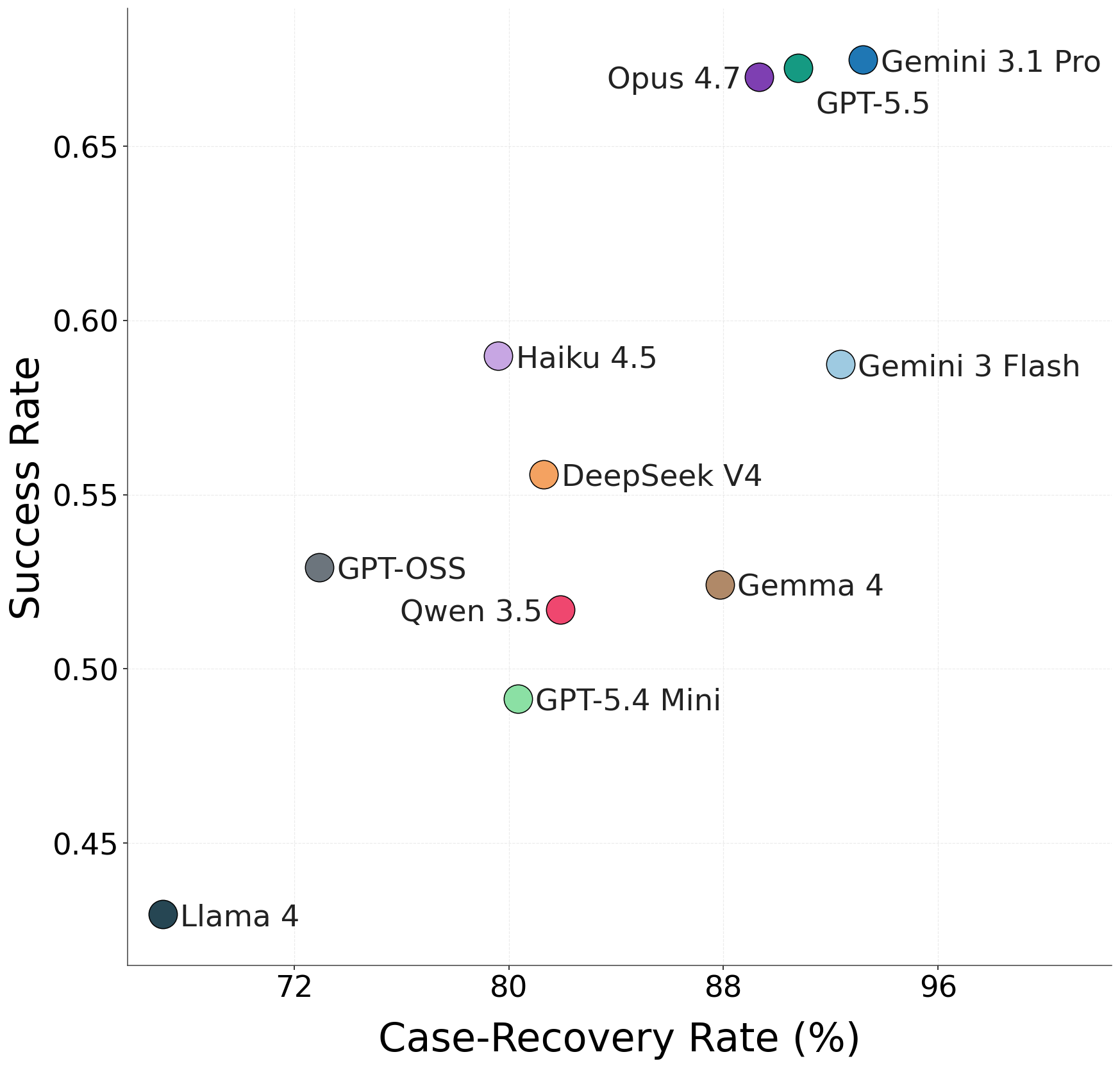}
    \caption{Correlation between case-recovery rate and immediate response success rate across models.}
    \label{fig:casefit_immediate}
\end{figure}

Figure~\ref{fig:casefit_immediate} shows the relationship between case-recovery rate and immediate-response success rates. Spearman correlation shows a positive association across models ($\rho=0.691$, $p<0.05$), suggesting that stronger models produce responses that are more specific to the current dialogue.

\section{MCQ Selection by Challenge Type} \label{app:mcq_challenge}

\begin{figure}[h]
    \centering
    \includegraphics[width=\columnwidth]{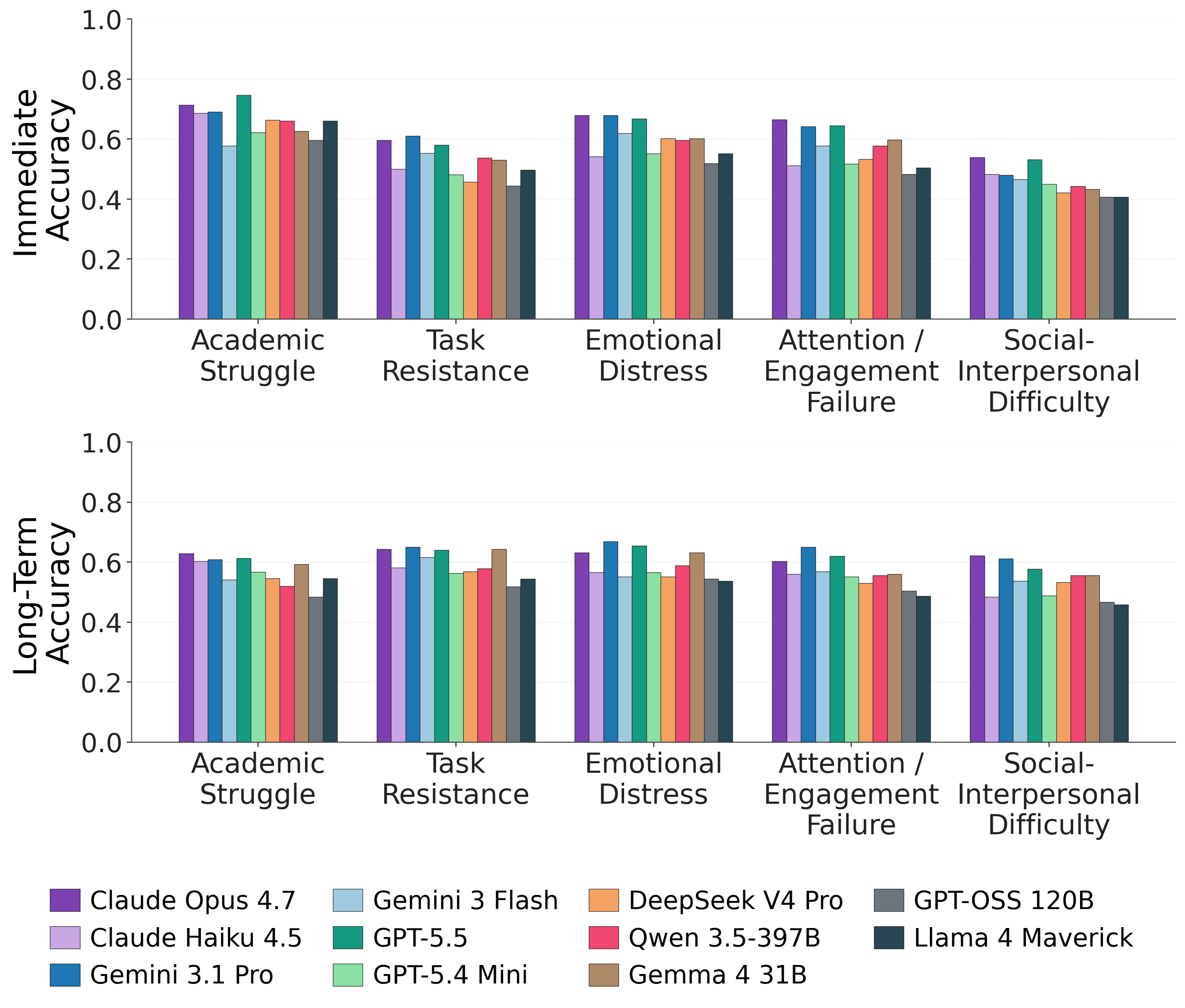}
    \caption{MCQ accuracy across models broken down by the challenge type of students.}
    \label{fig:mcq_result}
\end{figure}

Figure~\ref{fig:mcq_result} shows MCQ accuracy by challenge type. For immediate responses, social-interpersonal difficulty is the weakest category for all 11 models, while academic struggle is strongest for 10 of 11. For long-term plans, differences across challenge types are smaller, although social-interpersonal difficulty remains weakest for 7 of 11 models.

The contrast with open-ended evaluation is especially notable for these two categories. Academic struggle is the weakest category for all 11 models in long-term open-ended responses, but performs relatively well in MCQ selection. Social-interpersonal difficulty, in contrast, remains challenging in both settings. This suggests that academic strategies may be easier to recognize than generate, whereas social-interpersonal challenges remain difficult even when options are provided.

\section{MCQ Failure Analysis} \label{app:mcq_failure}

\begin{figure}[h]
    \centering
    \includegraphics[width=\columnwidth]{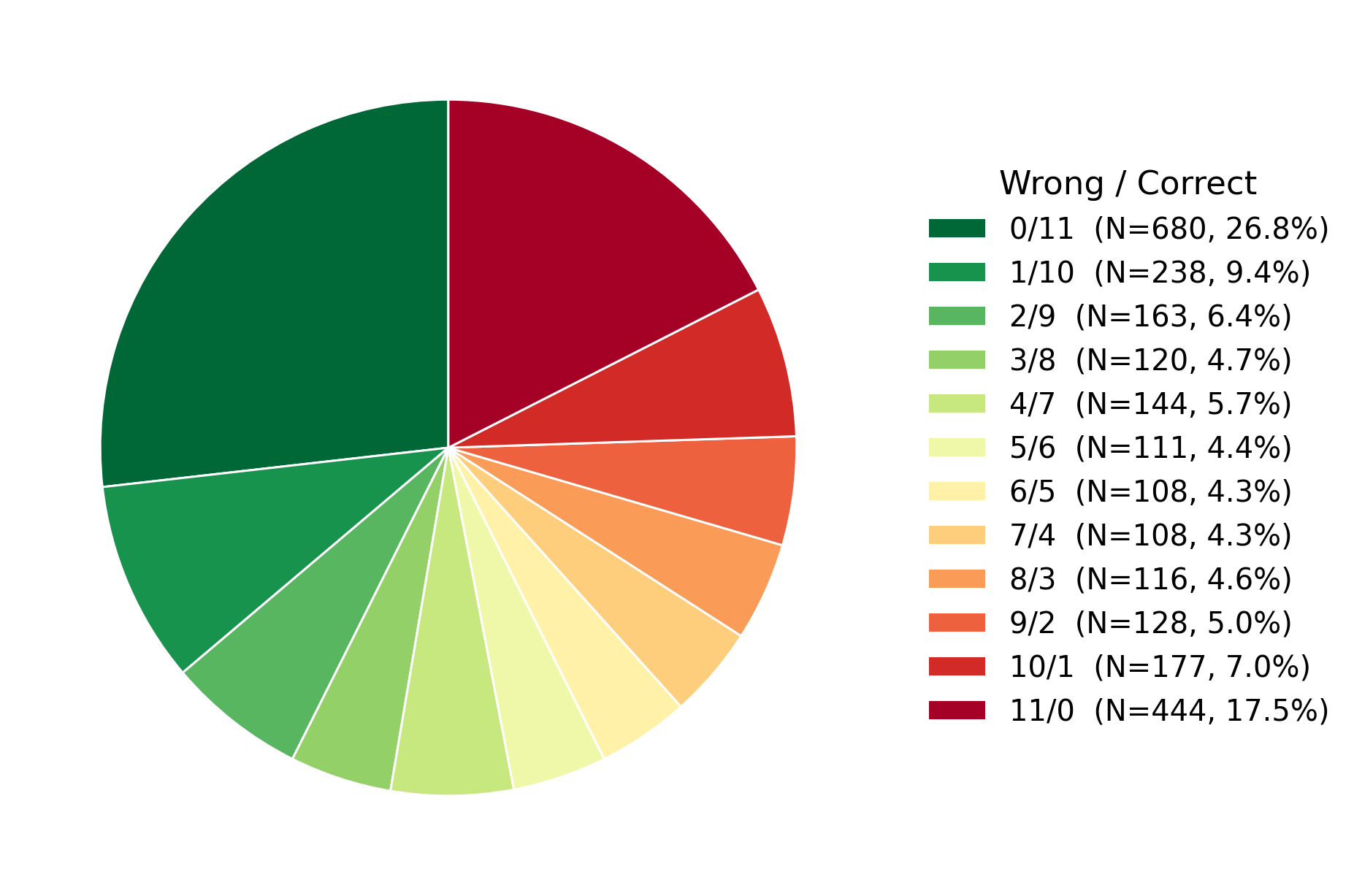}
    \caption{Distribution of model agreement on MCQ outcomes across all 2{,}537 items.}
    \label{fig:mcq_pi}
\end{figure}

Figure~\ref{fig:mcq_pi} summarizes model agreement across MCQ questions. All 11 models answer 26.8\% of questions correctly and miss 17.5\% of questions, indicating substantial agreement on both easy and difficult cases. Errors also show systematic convergence. Among questions missed by all models, all 11 select the same distractor in 40.3\% of cases.

\begin{figure}[h]
    \centering
    \includegraphics[width=\columnwidth]{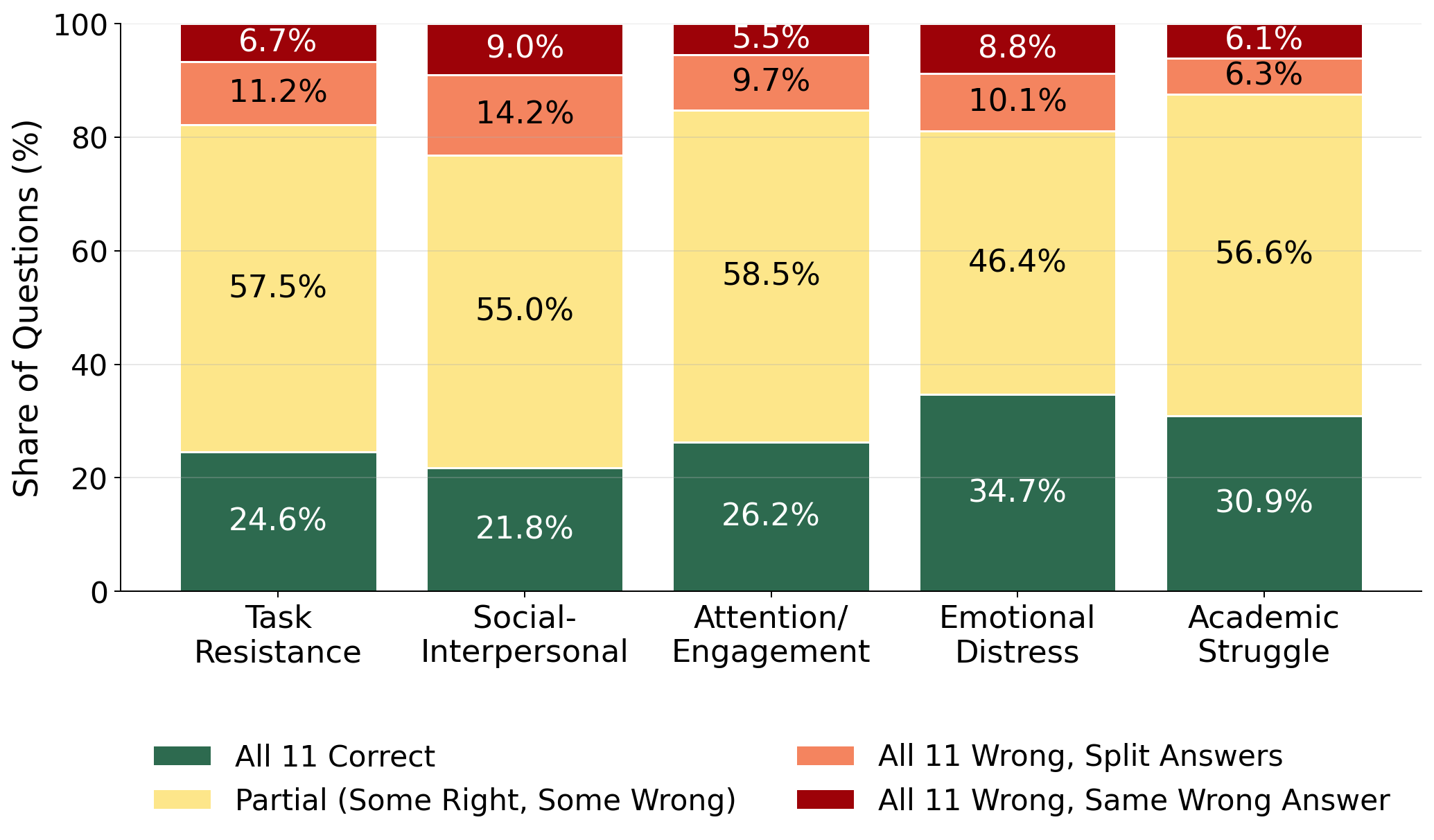}
    \caption{Model-agreement breakdown of MCQ outcomes by challenge type.}
    \label{fig:mcq_stacked}
\end{figure}

Figure~\ref{fig:mcq_stacked} breaks down these outcomes by challenge type. Social-interpersonal difficulty has the lowest all-correct share and the highest all-wrong share. It also has the highest share of questions where all models select the same wrong answer. These patterns suggest that social-interpersonal cases are not only more difficult, but also more likely to elicit the same errors across models.

\section{Reference-Based Response} \label{app:ref_longterm}

\begin{figure}[h]
    \centering
    \includegraphics[width=0.9\columnwidth]{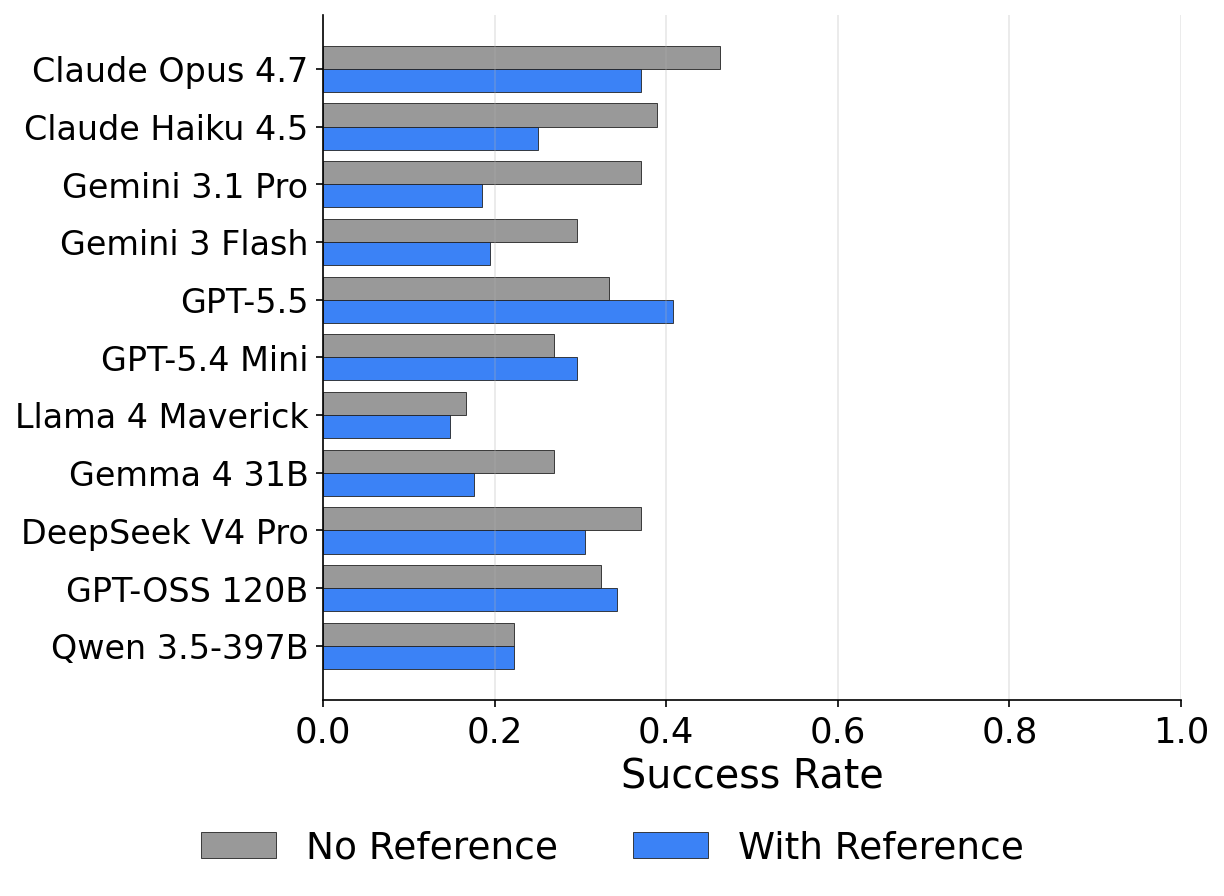}
    \caption{Success rate across models for long-term intervention plans, with and without references.}
    \label{fig:reference_longterm}
\end{figure}

Figure~\ref{fig:reference_longterm} compares long-term intervention performance with and without reference cases. In contrast to immediate responses, adding references often reduces performance for long-term intervention planning. Only 3 of 11 models improve their performance, 7 decline, and one is unchanged, leading to an average drop of 5.2 percentage points.

\section{Prompts Used in Dataset Construction}

Each prompt is shown with its system and user components separated by the marker \texttt{<System Prompt/User Prompt>}.



\vspace{0.3em}
\subsection{Abstract-Based Paper Filtering} \label{app:stage0_abstract}
\vspace{0.3em}
\begin{tcolorbox}[
    breakable, enhanced,
    width=\columnwidth,
    top=1pt, left=1pt, right=1pt, bottom=1pt,
    colback=white,
    fontupper=\ttfamily\small,
    fonttitle=\bfseries\rmfamily\small,
    subtitle style={
        fonttitle=\bfseries\rmfamily\small,
        fontupper=\bfseries\rmfamily\small,
    },
    title={Prompt}
]
You are a research screener selecting qualitative education papers that describe how challenging students behave in classroom settings and what teachers can do about it. Respond ONLY with a valid JSON object, no markdown, no extra text.
\\ \\
<System Prompt/User Prompt>
\\ \\
Read the abstract of the following education research paper. For each criterion below, answer Y (= filter out) or N (= keep). A paper is accepted only if ALL four answers are N.
\\ \\
Title: \{title\}
\\
Abstract: \{abstract\}
\\ \\ 
F1. Filter out if this study does NOT use observation or interviews with students as a research method.
\\
→ Y if only teachers, parents, or caregivers were observed or interviewed, or if neither method appears at all.
\\
→ N if students are directly observed or interviewed, or if the study design (e.g. case study, ethnographic study centered on a student) clearly implies it.
\\ \\ 
F2. Filter out if this study only tests the effectiveness of a specific intervention or teaching method through quantitative measurement.
\\
→ Y if pre-post tests, effect sizes, control groups, or statistical significance are the main focus, or if the study uses a single-case experimental design to measure behavioral change as an outcome of a specific treatment.
\\
→ N if the study is qualitative or descriptive without an efficacy-testing agenda.
\\ \\
F3. Filter out if this study focuses on aspects other than how students behave and respond in the classroom (learning context).
\\ 
→ Y if the central focus is diagnosis criteria, family background, parent-child relationships, policy, teacher competency, theory-building, or the development and validation of a measurement tool or rating scale — rather than students' actual behavioral patterns.
\\
→ N if concrete student behavior and reactions are the core of the study.
\\ \\
F4. Filter out if this study offers no practical guidance or implications for classroom teachers.
\\ 
→ Y if there is no indication — explicit or implied — that the findings offer any takeaway for how teachers should understand or respond to these students in a classroom setting.
\\
→ N if classroom-relevant insight or implications for teachers can be reasonably inferred from the findings, even if not stated as a formal recommendation.
\\ \\
Return JSON only:\\
\{\\
\ \ ``f1\_no\_student\_observation\_or\_interview'': ``Y'' or ``N'',\\
\ \ ``f2\_quantitative\_efficacy\_study'': ``Y'' or ``N'',\\
\ \ ``f3\_focus\_other\_than\_behavior'': ``Y'' or ``N'',\\
\ \ ``f4\_no\_practical\_guidance'': ``Y'' or ``N'',\\
\ \ ``reason'': ``one sentence explaining the overall judgment''\\
\}

\end{tcolorbox}


\vspace{0.3em}
\subsection{Paper Screening} \label{app:stage1_screen}
\vspace{0.3em}
\begin{tcolorbox}[
breakable, enhanced, top=1pt, left=1pt, right=1pt, bottom=1pt, colback=white, fontupper=\ttfamily, fonttitle=\bfseries\small,
title={Prompt}
]
\small
You are a research assistant that evaluates whether an educational paper contains enough concrete evidence to simulate a tutor-student interaction.
\\ \\ 
Score the paper on 3 dimensions using a 1–5 scale. Score 1 means the dimension is not described at all.
\\ \\
-------------------------------- \\
\#\#\# DIMENSIONS \\
-------------------------------- \\
\\ 
**1. Student Behavior** (1–5): Observable student actions and responses \\
- 1 = No student behavior described at all \\
- 2 = Vague/inferred (e.g., ``seemed frustrated'', ``was disengaged'') \\
- 3 = General behavior (e.g., ``avoided eye contact'', ``raised hand frequently'') \\
- 4 = Specific action in context (e.g., ``stayed silent when questioned about fractions'') \\
- 5 = Multiple concrete actions (e.g., ``pushed the worksheet aside, crossed arms, and looked away when asked to continue'') \\
\\
**2. Learning Context** (1–5): Tasks, materials, and settings \\
- 1 = No learning context described at all \\
- 2 = Extremely broad (e.g., ``during math instruction'') \\
- 3 = General task type (e.g., ``working on addition problems'', ``reading comprehension'') \\
- 4 = Specific activity (e.g., ``completing a worksheet on two-digit multiplication during independent work'') \\
- 5 = Highly specific, reproducible (e.g., ``solving 3/4 ÷ 1/2 using fraction strips during small-group instruction'') \\
\\
**3. Teacher Actions** (1–5): Instructional strategies and outcomes \\
- 1 = No teacher actions or instructional strategies described at all \\
- 2 = Actions implied (e.g., ``teacher helped the student'') \\
- 3 = General strategy (e.g., ``provided scaffolding'', ``gave positive reinforcement'') \\
- 4 = Concrete action (e.g., ``asked guiding questions'', ``modeled the problem on the board'') \\
- 5 = Strategy with documented outcome (e.g., ``after breaking the problem into smaller steps, the student successfully solved three similar problems independently'') \\
\\
-------------------------------- \\
\#\#\# OUTPUT FORMAT \\
-------------------------------- \\
\\
Output ONLY valid JSON. No explanations, no markdown. \\
\\ 
\{ \\
  ``student\_behavior'': \{ ``score'': 1-5, ``reason'': ``one sentence'' \}, \\
  ``learning\_context'': \{ ``score'': 1-5, ``reason'': ``one sentence'' \}, \\
  ``teacher\_actions'':  \{ ``score'': 1-5, ``reason'': ``one sentence'' \}, \\
  ``passed'': true if all three scores are $\geq$ 3, else false \\
\} \\
\\
<System Prompt/User Prompt> \\
\\
-------------------------------- \\
\#\#\# DOCUMENT TO SCREEN \\
-------------------------------- \\
\\
\{TEXT\} \\
\\
-------------------------------- \\
\end{tcolorbox}

\vspace{0.3em}
\subsection{Student Category Extraction} \label{app:stage1_category}
\vspace{0.3em}
\begin{tcolorbox}[
breakable, enhanced, top=1pt, left=1pt, right=1pt, bottom=1pt, colback=white, fontupper=\ttfamily, fonttitle=\bfseries\small,
title={Prompt}
]
\small
You are a research assistant specialized in extracting key information from educational studies. \\
\\
Your goal is to IDENTIFY student categories explicitly described in the paper and ASSESS their suitability for simulation. \\
\\
-------------------------------- \\
\#\#\# CRITICAL RULES \\
-------------------------------- \\
1. Extract ONLY what is explicitly stated - NO inference, NO generalization \\
2. ALL excerpts MUST be VERBATIM from the document - NO paraphrasing \\
3. If not stated, use empty string (``'') or null \\
4. Output ONLY valid JSON - NO explanations, NO markdown \\
\\
-------------------------------- \\
\#\#\# TASK OVERVIEW \\
-------------------------------- \\
\\
**STEP 1: Extract Basic Information** \\
- Metadata (title, author, year, description, research method) \\
- Research scope (country, education level, discipline) \\
\\
**STEP 2: Identify Student Categories** \\
- If paper describes students as one group → Create ONE category \\
- If paper describes individual students with DIFFERENT characteristics (e.g., avoidance vs. rushing) or DIFFERENT approaches needed (e.g., wait time vs. immediate redirection) → Create SEPARATE categories \\
  - If paper explicitly names student types (e.g., ``daydreaming students'', ``students with high anxiety'') → Use those \\
- If paper describes individual students with similar characteristics and same approach works for them → ONE category is enough \\
- Avoid broad labels: Each category name must include specific behavioral/cognitive/affective characteristics (e.g., ``Students who shut down when corrected'' NOT ``Challenging students'') \\
\\
**STEP 3: Assess Extraction Eligibility** \\
For EACH category, evaluate 3 dimensions: \\
\\
  **1. Student Behavior** (1-5): Observable actions and responses \\
  - 1 = No student behavior described at all \\
  - 2 = Vague/inferred (e.g., ``seemed frustrated'', ``was disengaged'') \\
  - 3 = General behavior (e.g., ``avoided eye contact'', ``raised hand frequently'') \\
  - 4 = Specific action in context (e.g., ``stayed silent when questioned about fractions'') \\
  - 5 = Multiple concrete actions (e.g., ``pushed the worksheet aside, crossed arms, and looked away when asked to continue'') \\
\\
  **2. Learning Context** (1-5): Tasks, materials, settings \\
  - 1 = No learning context described at all \\
  - 2 = Extremely broad (e.g., ``during math instruction'') \\
  - 3 = General task type (e.g., ``working on addition problems'', ``reading comprehension'') \\
  - 4 = Specific activity (e.g., ``completing a worksheet on two-digit multiplication during independent work'') \\
  - 5 = Highly specific, reproducible (e.g., ``solving 3/4 ÷ 1/2 using fraction strips during small-group instruction on Wednesday'') \\
\\
  **3. Teacher Actions** (1-5): Instructional strategies and outcomes \\
  - 1 = No teacher actions or instructional strategies described at all \\
  - 2 = Actions implied (e.g., ``teacher helped the student'') \\
  - 3 = General strategy (e.g., ``provided scaffolding'', ``gave positive reinforcement'') \\
  - 4 = Concrete action (e.g., ``asked guiding questions'', ``modeled the problem-solving process on the board'') \\
  - 5 = Strategy with documented outcome (e.g., ``after breaking the problem into smaller steps, the student successfully solved three similar problems independently'') \\

For each dimension: Write assessment → Provide supporting verbatim excerpt (only ONE most representative excerpt) → Assign score \\
\\
**Score 1 means the dimension is not described in the paper at all.** \\
Use 1 only when there is truly no evidence; use 2 when evidence exists but is very weak. \\
\\
-------------------------------- \\
\#\#\# OUTPUT STRUCTURE \\
-------------------------------- \\
\\
\{ \\
  ``id'': ``\{ID\}'', \\
\\
  ``metadata'': \{ \\
    ``title'': ``paper title'', \\
    ``author'': ``author names or `unknown' '', \\
    ``year'': year as number or null, \\
    ``description'': ``brief description of the paper's focus and purpose'', \\
    ``research\_method'': ``1-2 sentences on study design'' \\
  \}, \\
\\
  ``research\_scope'': \{ \\
    ``country'': ``country where study was conducted or empty string'', \\
    ``education\_level'': ``one of: early\_childhood, elementary\_school, middle\_school, high\_school, undergraduate, other'', \\
    ``discipline'': ``one of: mathematics, science, language\_arts, foreign\_language, social\_studies, engineering, teacher\_education, interdisciplinary, or other'' \\
  \}, \\
\\
  ``category\_list'': [ \\
    \{ \\
      ``category\_id'': ``\{ID\}-01 (use the paper id `\{ID\}' as prefix, then -01, -02, etc. for each category)'', \\
      ``name'': ``Generalized name of the student type (e.g., `Low-engagement adolescents with behavioral challenges')'', \\
      ``label'': [``identifier used in the paper (e.g., `Participant 1', `Tom'); otherwise make up a single unique label (e.g. `Label 1')''], \\
      ``categorization\_reasoning'': ``Why this category exists'', \\
\\
      ``demographics'': \{ \\
        ``age'': number or null, \\
        ``gender'': ``male | female | non\_binary | empty string'', \\
        ``background'': ``educational or socioeconomic background if explicitly stated; otherwise empty string'' \\
      \}, \\
\\
      ``extraction\_eligibility'': \{ \\ 
        ``student\_behavior'': \{ \\
          ``assessment'': ``What behavioral evidence exists in the paper? How concrete and detailed is it? (2-3 sentences)'', \\
          ``excerpt'': ``verbatim excerpt'', \\
          ``score'': 1-5 \\
        \}, \\
\\
        ``learning\_context'': \{ \\
          ``assessment'': ``What context evidence exists? How specific are the tasks/problems, materials, and settings? (2-3 sentences)'', \\
          ``excerpt'': ``verbatim excerpt'', \\
          ``score'': 1-5 \\
        \}, \\
\\
        ``teacher\_actions'': \{ \\
          ``assessment'': ``What instructional strategies are documented? How concrete are they? Are outcomes documented? (2-3 sentences)'', \\
          ``excerpt'': ``verbatim excerpt'', \\
          ``score'': 1-5 \\
        \} \\
      \} \\
    \} \\
  ], \\
\\
  ``irrelevant\_pages'': \{ \\
    ``page\_numbers'': [], \\
    ``sections'': [] \\
  \} \\
\} \\
\\
\\
<System Prompt/User Prompt> \\
\\
-------------------------------- \\
\#\#\# DOCUMENT TO ANALYZE \\
-------------------------------- \\
\\
\{TEXT\} \\
\\
-------------------------------- \\
\end{tcolorbox}


\vspace{0.3em}
\subsection{Learner Evidence Extraction} \label{app:stage2_learner}
\vspace{0.3em}
\begin{tcolorbox}[
breakable, enhanced, top=1pt, left=1pt, right=1pt, bottom=1pt, colback=white, fontupper=\ttfamily, fonttitle=\bfseries\small,
title={Prompt}
]
\small
You are a research assistant specialized in extracting key information from educational studies. \\ 
\\
PHASE 1 OF 2: LEARNER EVIDENCE EXTRACTION \\
Your goal is to extract ALL verbatim evidence about LEARNER BEHAVIOR, COGNITION, AND AFFECT. \\
\\
-------------------------------- \\
\#\#\# CRITICAL RULES \\
-------------------------------- \\
\\
1. Extract EXTENSIVELY - MAXIMUM COVERAGE of learner evidence \\
2. ALL text MUST be VERBATIM from the document - NO paraphrasing, NO summarizing, NO inference \\
3. EVERY excerpt MUST include source location (page/section) \\
4. ONE excerpt can apply to MULTIPLE categories if it shows multiple aspects \\
5. Output ONLY valid JSON - NO explanations, NO markdown \\
\\
-------------------------------- \\
\#\#\# INPUT CONTEXT \\
-------------------------------- \\
\\
ALL STUDENT CATEGORIES: \\
\{CATEGORIES\} \\
\\
- Use ``name'' and ``label'' field to identify which category\_id(s) each excerpt refers to. \\
- Your task is to extract evidence for ALL of them. \\
\\
-------------------------------- \\
\#\#\# TASK OVERVIEW \\
-------------------------------- \\

**STEP 1: Scan and Tag** \\
Read the entire paper and mentally note ALL locations containing STUDENT-CENTERED content: \\
  - Student behaviors (actions, responses, engagement patterns, physical movements) \\
  - Cognitive states (thinking processes, misconceptions, understanding levels) \\
  - Affective states (emotions, motivation, confidence, attitudes) \\
  - Difficulty triggers (conditions/situations causing struggle) \\
  - Student quotes (any direct utterances) \\
\\
DO NOT tag TEACHER-CENTERED content: \\
  - Pure intervention descriptions without student behavior (``Teachers implemented X'') → Skip for PHASE 1 \\
\\
**STEP 2: Extract Evidence** \\
For EACH tagged location: \\
  - Copy excerpt with 2-3 surrounding sentences VERBATIM (word-for-word from paper) \\
  - Focus on passages that primarily describe the STUDENT (even if teacher actions are mentioned as context) \\
  - Identify all applicable category\_id(s) based on labels \\
  - Classify excerpt\_type: behavior | cognitive | affective | difficulty\_trigger \\
  - Describe task\_context: Include (1) specific task/problem, (2) materials, (3) instructional setting
    (e.g., ``solving word problem `If train leaves at 2pm going 60mph...' on worksheet during independent practice, 10 minutes into activity'', ``answering comprehension questions about Chapter 3 using graphic organizer during whole-class discussion'') \\
  - For quotes: extract EXACT STUDENT WORD with surrounding context (avoid extracting other's word) \\
\\
Target: At least 10 excerpts per category (if paper has sufficient content) \\
\\
**STEP 3: Verify Coverage** \\
Check for gaps: \\
  - Any category with $<$ 10 excerpts? → Search for more evidence if possible \\
  - Missing evidence types? (behavior, cognitive, affective, triggers, quotes) → Search harder \\
  - Any vague task\_context? → Add concrete details from paper \\
\\
**STEP 4: Verbatim Verification (CRITICAL)** \\
For EVERY excerpt and quote, perform strict verification: \\
  1. Re-check EACH item against the source document \\
  2. Verify word-by-word match (including punctuation, capitalization) \\
  3. Record result in verbatim\_check dictionaries: \\
    - excerpt\_verbatim\_check: \{``\{ID\}\_learner\_001'': true, ``\{ID\}\_learner\_002'': false\} \\
    - quote\_verbatim\_check: \{``\{ID\}\_quote\_001'': true\} \\
  4. If ANY item is not verbatim (paraphrased/summarized), mark as false \\
  5. Explain any non-verbatim items in verbatim\_reasoning \\
\\

-------------------------------- \\
\#\#\# OUTPUT STRUCTURE \\ 
-------------------------------- \\
\\
\{ \\
  ``paper\_id'': ``\{ID\}'', \\
  ``category\_ids'': [``list of all category\_ids''], \\
\\
  ``learner\_excerpts'': [ \\
    \{ \\
      ``excerpt\_id'': ``\{ID\}\_learner\_001'', \\
      ``applicable\_categories'': [``\{ID\}-01''], \\
      ``excerpt\_text'': ``VERBATIM passage with 2-3 surrounding sentences for full context'', \\
      ``source\_location'': ``page X / section Y'', \\
      ``excerpt\_type'': ``behavior | cognitive | affective | difficulty\_trigger'', \\
      ``task\_context'': ``DETAILED description: What specific task was the student doing? Include concrete details like actual problem content, materials used, task stage.'' \\
    \} \\
  ], \\
  \\
  ``student\_quotes'': [ \\
    \{ \\
      ``quote\_id'': ``\{ID\}\_quote\_001'', \\
      ``applicable\_categories'': [``\{ID\}-01''], \\
      ``quote\_text'': ``Exact words spoken by student'', \\
      ``source\_location'': ``page X / section Y'', \\
      ``context'': ``What was happening when student said this (task, situation, trigger)'' \\
    \} \\
  ], \\
\\
  ``extraction\_summary'': \{ \\
    ``total\_excerpts'': count, \\
    ``excerpts\_per\_category'': \{ \\
      ``\{ID\}-01'': count, \\
      ``\{ID\}-02'': count \\
    \}, \\
    ``excerpt\_verbatim\_check'': \{ \\
      ``\{ID\}\_learner\_001'': true, \\
      ``\{ID\}\_learner\_002'': true, \\
      ``\{ID\}\_learner\_003'': false \\
    \}, \\
    \\
    ``total\_quotes'': count, \\
    ``quotes\_per\_category'': \{ \\
      ``\{ID\}-01'': count \\
    \}, \\
    ``quote\_verbatim\_check'': \{ \\
      ``\{ID\}\_quote\_001'': true, \\
      ``\{ID\}\_quote\_002'': false \\
    \}, \\
    \\
    ``verbatim\_reasoning'': ``Brief: Explain any non-verbatim items (why marked false). If all true, state `All excerpts and quotes verified as verbatim'.'', \\
    ``verbatim\_check\_performed'': true \\
  \} \\
\} \\
\\

<System Prompt/User Prompt> \\

-------------------------------- \\
\#\#\# DOCUMENT TO ANALYZE \\
-------------------------------- \\
\\
\{TEXT\}\\
\\
-------------------------------- \\
\end{tcolorbox}

\vspace{0.3em}
\subsection{Instructional Evidence Extraction} \label{app:stage2_instruction}
\vspace{0.3em}
\begin{tcolorbox}[
breakable, enhanced, top=1pt, left=1pt, right=1pt, bottom=1pt, colback=white, fontupper=\ttfamily, fonttitle=\bfseries\small,
title={Prompt}
]
\small
You are a research assistant specialized in extracting key information from educational studies. \\
\\
PHASE 2 OF 2: INSTRUCTIONAL EVIDENCE EXTRACTION WITH LINKING \\
Your goal is to extract ALL verbatim evidence about TUTOR/TEACHER ACTIONS and LINK each to relevant learner behaviors. \\
\\
-------------------------------- \\
\#\#\# CRITICAL RULES \\
-------------------------------- \\
\\
1. Extract EXTENSIVELY - MAXIMUM COVERAGE of instructional evidence \\
2. ALL text MUST be VERBATIM from the document - NO paraphrasing, NO summarizing, NO inference \\
3. EVERY excerpt MUST include source location (page/section) \\
4. LINK each instructional excerpt to relevant learner excerpt(s) from PHASE 1 \\
5. Output ONLY valid JSON - NO explanations, NO markdown \\
\\
-------------------------------- \\
\#\#\# INPUT CONTEXT \\
-------------------------------- \\
\\
ALL STUDENT CATEGORIES: \\
\{CATEGORIES\} \\
\\
LEARNER EXCERPTS from PHASE 1: \\
\{LEARNER\_EXCERPTS\} \\
\\
- Use ``name'' and ``label'' field to identify which category\_id(s) each excerpt refers to. \\
- Your task is to extract evidence for ALL of them. \\
\\
-------------------------------- \\
\#\#\# TASK OVERVIEW \\
-------------------------------- \\
\\
**STEP 1: Scan and Tag** \\
Read the entire paper and mentally note ALL locations containing TEACHER-CENTERED content: \\
  - Effective teacher actions (strategies that worked, recommended practices) \\
  - Ineffective teacher actions (strategies that failed or actions to be avoided) \\
  - Intervention descriptions (what teachers DID or SHOULD DO) \\
  - Outcomes (student responses to interventions - as results of teacher actions) \\
\\
DO NOT tag STUDENT-CENTERED content already covered in PHASE 1: \\
  - Pure behavioral observations without intervention context → Skip for PHASE 2 \\
\\
**STEP 2: Extract Evidence** \\
For EACH tagged location: \\
  - Copy excerpt with 2-3 surrounding sentences VERBATIM (word-for-word from paper) \\
  - Focus on passages that primarily describe what TEACHERS did or should do \\
  - Identify applicable category\_id(s) \\
  - Classify excerpt\_type: tutor\_action\_effective | tutor\_action\_ineffective \\
  - Fill instructional\_context: \\
    * when: Conditions when this applies (e.g., ``when student shows frustration'') \\
    * how: Concrete steps/approach (e.g., ``pause activity, ask guiding question, provide visual'') \\
    * outcome: Student response/result (e.g., ``student successfully solved similar problem'') \\
  - Classify temporal\_scope: \\
    * immediate: Applied in-the-moment (e.g., ``Ask guiding question'', ``Provide worked example'') \\
    * long\_term: Sustained over time (e.g., ``Establish error-friendly culture'', ``Gradually increase difficulty'') \\
    * both: Has immediate AND sustained components \\
\\
**STEP 3: Link to Learner Behaviors** \\
For EACH instructional excerpt: \\
  - Review learner excerpts from PHASE 1 \\
  - Identify which learner behaviors this instruction addresses \\
  - Write explicit linking\_rationale with evidence from paper \\
  - Assign link\_strength: \\
    * 3 (Strong): Paper explicitly connects instruction to this behavior \\
    * 2 (Moderate): Addresses same challenge type; temporal scopes align \\
    * 1 (Weak): General strategy that could apply; same category \\
    * 0 (No link): Different behavior/category or temporal mismatch \\
\\
**STEP 4: Verify Coverage** \\
Check for gaps: \\
  - Any orphaned learner excerpts (no linked instructions)? → Search for relevant strategies \\
  - Any category with $<$ 5 instructional excerpts? → Search harder \\
  - Any vague instructional\_context? → Add concrete details from paper \\

**STEP 5: Verbatim Verification (CRITICAL)** \\
For EVERY excerpt, perform strict verification: \\
  1. Re-check EACH item against the source document \\
  2. Verify word-by-word match (including punctuation, capitalization) \\
  3. Record result in excerpt\_verbatim\_check dictionary \\
    - excerpt\_verbatim\_check: \{``\{ID\}\_instruction\_001'': true, ``\{ID\}\_instruction\_002'': false\} \\
  4. If ANY item is not verbatim (paraphrased/summarized), mark as false \\
  5. Explain any non-verbatim items in verbatim\_reasoning \\
\\
-------------------------------- \\
\#\#\# OUTPUT STRUCTURE \\
-------------------------------- \\
\{ \\
  ``paper\_id'': ``\{ID\}'', \\
  ``category\_ids'': [``list of all category\_ids''], \\
  \\
  ``instructional\_excerpts'': [ \\
    \{ \\
      ``excerpt\_id'': ``\{ID\}\_instruction\_001'', \\
      ``applicable\_categories'': [``\{ID\}-01''], \\
      ``excerpt\_text'': ``VERBATIM passage with 2-3 surrounding sentences for full context'', \\
      ``source\_location'': ``page X / section Y'', \\
      ``excerpt\_type'': ``tutor\_action\_effective | tutor\_action\_ineffective'', \\
      \\
      ``instructional\_context'': \{ \\
        ``when'': ``Conditions when this applies; empty string if not stated'', \\
        ``how'': ``Concrete steps/approach; empty string if not stated'', \\
        ``outcome'': ``Student response/result; empty string if not stated'' \\
      \}, \\
      \\
      ``linked\_learner\_excerpts'': [ \\
        \{ \\
          ``learner\_excerpt\_id'': ``\{ID\}\_learner\_003'', \\
          ``linking\_rationale'': ``EXPLICIT: Why this instruction addresses this behavior, with evidence from paper'', \\
          ``link\_strength'': 0-3 \\
        \} \\
      ], \\
      \\
      ``temporal\_scope'': ``immediate | long\_term | both'' \\
    \} \\
  ], \\
  \\
  ``extraction\_summary'': \{ \\
    ``total\_excerpts'': count, \\
    ``excerpts\_per\_category'': \{ \\
      ``\{ID\}-01'': count \\
    \}, \\
    ``excerpt\_verbatim\_check'': \{ \\
      ``\{ID\}\_instruction\_001'': true, \\
      ``\{ID\}\_instruction\_002'': false \\
    \}, \\
    ``verbatim\_reasoning'': ``Brief: Explain any non-verbatim items. If all true, state `All excerpts verified as verbatim'.'', \\
    ``verbatim\_check\_performed'': true, \\
    \\
    ``linking\_summary'': \{ \\
      ``total\_links'': count, \\
      ``strong\_links'': count, \\
      ``moderate\_links'': count, \\
      ``weak\_links'': count, \\
      ``orphaned\_learner\_excerpts'': [``\{ID\}\_learner\_012''], \\
      ``orphaned\_instruction\_excerpts'': [``\{ID\}\_instruction\_015''], \\
      ``linking\_reasoning'': ``Brief: Quality of links between instructions and learner behaviors.'' \\
    \} \\
  \} \\
\} \\
\\

<System Prompt/User Prompt> \\
\\
-------------------------------- \\
\#\#\# DOCUMENT TO ANALYZE \\
-------------------------------- \\
\\
\{TEXT\} \\\
\\
-------------------------------- \\
\end{tcolorbox}


\vspace{0.3em}
\subsection{Scenario Construction} \label{app:stage3_scenario}
\vspace{0.3em}
\begin{tcolorbox}[
breakable, enhanced, top=1pt, left=1pt, right=1pt, bottom=1pt, colback=white, fontupper=\ttfamily, fonttitle=\bfseries\small,
title={Prompt}
]
\small
You are a research assistant specialized in building structured scenarios from educational research evidence. \\
\\
Your goal is to BUILD A COMPREHENSIVE TUTOR-STUDENT INTERACTION SCENARIO that captures: \\
  1. The CONTEXT where challenging interactions occur \\
  2. The CHALLENGING BEHAVIORS the student exhibits \\
  3. HIDDEN STRATEGIES about what works and doesn't work (for evaluation purposes) \\
\\
-------------------------------- \\
\#\#\# CRITICAL RULES \\
-------------------------------- \\
1. Ground ALL statements in provided excerpts - NO INFERENCE \\
2. Use EXACT descriptions from excerpts, not interpretations - DO NOT create behaviors or actions not explicitly in evidence \\
3. Reference excerpts by ID for EVERY claim (e.g., ``0001\_learner\_001'') \\
- evidence\_basis arrays may contain ONLY IDs listed in VALID EVIDENCE IDS \\
- NEVER use meta labels such as "category\_extraction", "category", "paper", "prompt", or any invented evidence ID \\
4. CHALLENGING BEHAVIORS = student behaviors that are difficult for a tutor to address using standard instructional strategies — extract as many as possible \\
5. The scenario describes the PROBLEM, not the SOLUTION. Effective strategies are kept as ``hidden strategies'' for evaluating tutor responses later — NO SOLUTION LEAKAGE into the visible content \\
6. Focus on PERSONALIZED, STUDENT-SPECIFIC strategies tailored to this individual student's unique needs and characteristics \\
7. Every strategy that references specific tools, materials, or conditions (e.g., visual aids, manipulatives) MUST have those elements documented in the context field - If not in context, do not include the strategy \\
8. Output ONLY valid JSON - NO explanations, NO markdown \\
\\
-------------------------------- \\
\#\#\# INPUT CONTEXT \\
-------------------------------- \\
\\
STUDENT CATEGORY: \\
\{CATEGORY\} \\
\\
- Focus ONLY on evidence explicitly about this student. \\
- The category description is context for selecting the student only. It is NOT an evidence source and must NEVER appear in evidence\_basis. \\
\\
VALID EVIDENCE IDS: \\
\{ALLOWED\_EVIDENCE\_IDS\} \\
\\
-------------------------------- \\
\#\#\# TASK OVERVIEW \\
-------------------------------- \\
\\
**STEP 1: Scan and Tag** \\
Read the entire evidence and mentally note ALL locations containing STUDENT-CENTERED content: \\
  - What CONTEXT(S) does the student struggle in? \\
  - What BEHAVIORS does the student exhibit when struggling? \\
  - What strategy WORKS for this student? \\
  - What strategy DOESN'T WORK for this student? \\
\\
**STEP 2: Identify Distinct Scenarios** \\
Generate as many naturally distinct scenarios as the evidence supports. Recommended: \textasciitilde\{NUM\_SCENARIOS\}. Each scenario contains one specific task/context where student struggled, their challenging behaviors, and appropriate strategies for tutors. \\
  - If DIFFERENT specific task/context found → Create separate scenarios \\
  - If DIFFERENT strategies should be applied → Create separate scenarios \\
  - Do NOT force artificial scenarios if the evidence does not support them — fewer scenarios of high quality is better than more scenarios of low quality \\
  - If the evidence supports more than \{NUM\_SCENARIOS\} meaningfully distinct scenarios, include all of them \\
  - Each scenario MUST be meaningfully different from the others — not just surface-level rewording \\
  - All scenarios must be grounded in the provided evidence \\
\\
**STEP 3: For EACH Scenario - Extract Context \& Task** \\
Gather evidence about WHERE, WHEN and WHAT (will be used for dialogue opening): \\
  - Task: Exact problem content (e.g., ``Solve 3/4 $\div$ 1/2 using fraction strips'', ``Build 5-block tower in construction center''), learning goal if stated \\
  - Context: Physical environment, materials, social setting, time if available \\
  - NOTE: Only materials/tools documented here can appear in hidden strategies later \\
\\
**STEP 4: For EACH Scenario - Extract Student Behaviors \& Triggers** \\
Identify HOW student struggles (will be used for dialogue middle/end): \\
  - Challenging behavior trigger: The single most critical moment \\
  - Behavior patterns: Recurring observable behaviors the student shows during struggle \\
  - All behaviors (both fields) must satisfy: \\
    * Observable: What can a tutor physically SEE? (not inferred states) \\
    * Specific: Exact actions, not vague labels (not ``frustrated'' but ``sighed, crossed arms, said `this is stupid' '') \\
    * Trigger-linked: Each behavior must be paired with what caused it \\
    * Internal state: Documented thoughts/feelings if available \\
  - Include ALL related student quotes VERBATIM where available \\
\\
**STEP 5: For EACH Scenario - Extract Hidden Strategies (STUDENT-SPECIFIC)** \\
Extract effective/ineffective strategies (be as specific as possible, leave empty array if no explicit strategies documented): \\
  - Strategies should be decomposed into single, actionable units feasible within the given context \\
  - Extract strategies TAILORED to THIS student's unique needs, NOT general best practices \\
  - Strategies must be aligned with the scenario's task, context, and challenging behavior trigger \\
    - ALIGNMENT CHECK: If a strategy references a tool or condition not in the context field → exclude it \\
  - For EACH strategy, extract four components: \\
    * Instruction: Exact description of what tutor did/should do \\
    * Context: Exact triggering situation from evidence \\
    * Outcome: Exact documented result/student response \\
    * Reason: Why this strategy is effective/ineffective for THIS student (mechanism, not just result) \\
  - Effective immediate: In-the-moment responses (3$+$ if possible) \\
  - Effective long-term: Sustained approaches over weeks/months (3$+$ if possible) \\
  - Ineffective: Strategies that failed or should be avoided (ONLY if paper EXPLICITLY describes failure, 3$+$ if possible) \\
\\
**STEP 6: Extract Student Persona** \\
Identify and synthesize core differentiating characteristics across three dimensions: \\
  - Behavioral patterns: What this student characteristically does (e.g., ``Elopes when redirected, not when task is hard'') \\
  - Unique triggers \& specific needs: What causes struggle and what this student uniquely requires (e.g., ``Transitions cause shutdown; requires 10+ second wait time'') \\
  - Approach effectiveness: What works and what doesn't for THIS student specifically (e.g., ``Visual models work; verbal praise increases frustration'') \\
Write description (3-4 sentences) that weaves these characteristics together to answer: \\
  - What makes them challenging? \\
  - What are their core differentiating characteristics? \\
\\
**STEP 7: Categorize Overall Challenges** \\
Identify the primary challenge type (required) and secondary challenge type (optional — only if the evidence clearly supports a meaningfully distinct second challenge). The subcategories below describe what each type covers: \\
  - **Task Resistance** — explicit task refusal, passive noncompliance, task escape behavior, help avoidance, gaming/shortcutting \\
  - **Emotional Distress** — frustration expression, anxiety expression, learned helplessness, low self-efficacy, sadness/low mood \\
  - **Attention/Engagement Failure** — off-task motor, off-task verbal, off-task passive, boredom/disinterest, distractibility \\
  - **Social-Interpersonal Difficulty** — verbal aggression/hostility, physical aggression, defensiveness/blame externalization, social withdrawal, attention-seeking disruption \\
  - **Academic Struggle** — confusion/cognitive disequilibrium, persistent/repeated errors, uncertain response, confident misconception, failure to respond \\
\\
-------------------------------- \\
\#\#\# OUTPUT STRUCTURE \\
-------------------------------- \\
\\
\{ \\
  ``category\_id'': ``\{CATEGORY\_ID\}'', \\
  ``paper\_id'': ``\{ID\}'', \\
  ``name'': ``student category name'', \\
  \\
  ``scenarios'': [ \\
    \{ \\
      ``scenario\_id'': ``\{CATEGORY\_ID\}\_scenario\_001'', \\
      ``title'': ``Brief descriptive title (e.g., `Multi-step fraction problem with visual supports')'', \\
      \\
      ``content'': \{ \\
        ``task'': \{ \\
          ``content'': ``EXACT problem/activity content from excerpts (e.g., `Divide 3/4 by 1/2 and explain the steps')'', \\
          ``learning\_goal'': ``EXACT objective from excerpts; empty string if not stated'', \\
          ``evidence\_basis'': [``0001\_learner\_003''] \\
        \}, \\
        \\
        ``context'': \{ \\
          ``environment'': ``EXACT setting from excerpts (e.g., `Quiet corner of classroom during math block')'', \\
          ``materials'': ``EXACT items from excerpts (e.g., `Worksheet, fraction strips, pencil, visual aid poster on wall')'', \\
          ``social\_setting'': ``EXACT arrangement from excerpts (e.g., `One-on-one tutoring session, other students working independently')'', \\
          ``time\_context'': ``If stated in excerpts; empty string if not stated'', \\
          ``evidence\_basis'': [``0001\_learner\_004''] \\
        \}, \\
        \\
        ``challenging\_behavior\_trigger'': \{ \\
          ``condition'': ``EXACT final triggering condition from excerpts (e.g., `After third incorrect attempt when tutor asked to try again')'', \\
          ``observable\_behavior'': ``Observable actions EXACTLY from excerpts (e.g., `Pushed worksheet away, crossed arms, turned head away from tutor')'', \\
          ``internal\_state'': ``EXACT cognition/affect from excerpts if documented; empty string if not stated'', \\
          ``related\_quotes'': [``VERBATIM quotes''], \\
          ``evidence\_basis'': [``0001\_learner\_005''] \\
        \}, \\
\\
        ``behavior\_patterns'': [ \\
          \{ \\
            ``condition'': ``EXACT trigger from excerpts (e.g., `When asked to explain reasoning')'', \\
            ``observable\_behavior'': ``EXACT actions from excerpts (e.g., `Looked down, fidgeted with pencil, said nothing for 8 seconds')'', \\
            ``internal\_state'': ``EXACT cognition/affect from excerpts if documented; empty string if not stated'', \\
            ``related\_quotes'': [``VERBATIM quotes''], \\
            ``evidence\_basis'': [``0001\_learner\_002''] \\
          \} \\
        ] \\
      \}, \\
      \\
      ``hidden\_strategies'': \{ \\
        ``effective\_immediate'': [ \\
          \{ \\
            ``action\_id'': ``\{CATEGORY\_ID\}\_scenario\_001\_effim\\\_01'', \\
            ``instruction'': ``Description of recommended immediate strategy EXACTLY from excerpts (e.g., `Paused activity, acknowledged frustration, asked What part makes sense to you?, provided fraction strip model')'', \\
            ``context'': ``Context where strategy was tried/recommended EXACTLY from excerpts (e.g., `When student showed frustration after making error on division problem')'', \\
            ``outcome'': ``Result or student response EXACTLY from excerpts (e.g., `Student re-engaged, used strips to solve problem, completed 3 similar problems independently')'', \\
            ``reason'': ``Why this works for THIS student - mechanism from excerpts (e.g., `Student responds to choice and acknowledgment; reduces perceived threat of failure')'', \\
            ``evidence\_basis'': [``0001\_instruction\_002''] \\
          \} \\
        ], \\
        \\
        ``effective\_long\_term'': [ \\
          \{ \\
            ``action\_id'': ``\{CATEGORY\_ID\}\_scenario\_001\_efflt\\\_01'', \\
            ``instruction'': ``Description of recommended long-term strategy EXACTLY from excerpts'', \\
            ``context'': ``Context where strategy was tried/recommended EXACTLY from excerpts'', \\
            ``outcome'': ``Result over time EXACTLY from excerpts'', \\ 
            ``reason'': ``Why this works for THIS student - mechanism from excerpts'', \\
            ``evidence\_basis'': [``0001\_instruction\_007''] \\
          \} \\
        ], \\
         \\
        ``ineffective'': [ \\
          \{ \\
            ``action\_id'': ``\{CATEGORY\_ID\}\_scenario\_001\_ineff\\\_01'', \\
            ``instruction'': ``Description of strategy to avoid EXACTLY from excerpts (e.g., `Immediately corrected error verbally without wait time or visual support')'', \\
            ``context'': ``Context where strategy was tried EXACTLY from excerpts'', \\
            ``outcome'': ``Negative result EXACTLY from excerpts (e.g., `Student shut down, refused to continue, said I told you I cannot do this')'', \\
            ``reason'': ``Why this fails for THIS student - mechanism from excerpts (e.g., `Student interprets immediate correction as confirmation of inability')'', \\
            ``evidence\_basis'': [``0001\_instruction\_015''] \\
          \} \\
        ] \\
      \} \\
    \} \\
  ], \\
  \\
  ``student\_persona'': \{ \\
    ``behavioral\_patterns'': ``What this student characteristically does during struggle — EXACT from excerpts (e.g., `Pushes materials away, goes silent, refuses to re-engage')'', \\
    ``unique\_triggers\_and\_needs'': ``What causes breakdown and what this student uniquely requires — EXACT from excerpts (e.g., `Shuts down after repeated errors; needs error-reframing before re-engagement')'', \\
    ``approach\_effectiveness'': ``What works and what doesn't for THIS student specifically — EXACT from excerpts (e.g., `Visual models effective; verbal-only correction increases resistance')'', \\
    ``description'': ``3-4 sentences integrating the above three dimensions: what makes them challenging and their core differentiating characteristics'', \\
    ``evidence\_basis'': [``0001\_learner\_002'', ``0001\_learner\_005''] \\
  \}, \\
  \\
  ``challenge\_categorization'': \{ \\
    ``primary\_challenge'': \{ \\
      ``evidence\_summary'': ``FIRST: Specific behaviors/patterns'', \\
      ``type'': ``THEN: Task Resistance | Emotional Distress | Attention/Engagement Failure | Social-Interpersonal Difficulty | Academic Struggle'', \\
      ``evidence\_basis'': [``0001\_learner\_002'', ``0001\_learner\_005''] \\
    \}, \\
    ``secondary\_challenge'': \{       // null if evidence does not clearly support a second distinct challenge type \\
      ``evidence\_summary'': ``FIRST: Specific behaviors/patterns'', \\
      ``type'': ``THEN: Task Resistance | Emotional Distress | Attention/Engagement Failure | Social-Interpersonal Difficulty | Academic Struggle'', \\
      ``evidence\_basis'': [``0001\_learner\_007''] \\
    \} \\
  \} \\
\} \\
\\
<System Prompt/User Prompt> \\
\\
-------------------------------- \\
\#\#\# INPUT EVIDENCE \\
-------------------------------- \\
\\
LEARNER EVIDENCE: \\
\{LEARNER\_EXCERPT\} \\
\\
INSTRUCTIONAL EVIDENCE: \\
\{INSTRUCTION\_EXCERPT\} \\
\\
-------------------------------- \\
\end{tcolorbox}


\vspace{0.3em}
\subsection{Dialogue Synthesis} \label{app:stage4_dialogue}
\vspace{0.3em}
\begin{tcolorbox}[
breakable, enhanced, top=1pt, left=1pt, right=1pt, bottom=1pt, colback=white, fontupper=\ttfamily, fonttitle=\bfseries\small,
title={Prompt}
]
\small
You are a research assistant specialized in creating realistic dialogue from educational research scenarios. \\
\\
Your goal is to CREATE REALISTIC STUDENT-TUTOR DIALOGUES that demonstrate CHALLENGING MOMENTS requiring intervention.\\
\\
-------------------------------- \\
\#\#\# CRITICAL RULES \\
-------------------------------- \\
1. Ground ALL statements in provided profile/excerpts - NO INFERENCE \\
2. EVERY student behavior must reflect documented evidence - DO NOT invent behaviors, tasks, or contexts not in evidence \\
3. Reference evidence by ID for EVERY message \\
- evidence\_basis arrays may contain ONLY IDs that appear as keys in EVIDENCE \\
- NEVER use meta labels such as "category\_extraction", "scenario", "persona", "paper", "prompt", or invented IDs \\
4. Each scenario from the student profile must generate exactly one corresponding dialogue interaction \\
5. Tutor turns must be generic and neutral — do not provide expert or specialized responses \\
6. The dialogue sets up a situation that REQUIRES good intervention \\
7. Final student turn shows challenging behavior that needs response (CHALLENGING BEHAVIORS = student behaviors that are difficult for a tutor to address using standard instructional strategies) \\
- The final turn MUST be a student turn and MUST include "is\_challenging\_behavior": true \\
- Every student turn MUST include a non-empty evidence\_basis array \\
8. Output ONLY valid JSON - NO explanations, NO markdown \\
\\
-------------------------------- \\
\#\#\# TASK OVERVIEW \\
-------------------------------- \\
\\
**STEP 1: For EACH Scenario - Plan Dialogue** \\
Review scenario from profile: \\
  - Task: What specific problem/activity? (concrete from evidence) \\
  - Context: Environment/materials/social setting (verify action possibilities) \\
  - Student persona: Core characteristics, typical responses \\
  - Difficulty trigger: What challenging behavior will end the dialogue? What causes final struggle? \\
  - Available actions: What tutor/student can physically do? \\
\\
**STEP 2: For EACH Scenario - Structure Dialogue** \\
- Number of turns = 2 × \{TURN\} (e.g., TURN$=$1 → 2 turns, TURN$=$3 → 6 turns, default TURN$=$3) \\
- **First Turn (Tutor)**: Present SPECIFIC, CONCRETE task \\
  - State the specific task from the scenario context (neutral tone) \\
  - Generic and neutral only — no expert or specialized responses \\
- **Turns in the middle (Mixed)**: Student attempts → Tutor NEUTRAL response \\
  - Student: May show emerging problem behaviors (based on evidence, follow persona, use quotes when appropriate) \\
  - Tutor: Generic/neutral response only (e.g., ``What do you think?'', ``Keep trying'') - DO NOT use effective strategies \\
  - Physical actions in parentheses ONLY if grounded in evidence and context allows \\
- **Last Turn (Student)**: MUST show challenging/struggling behavior \\
  - Observable problem behaviors matching challenging\_behavior\_trigger \\
  - Grounded in documented behavior\_patterns \\
\\
**STEP 3: For EACH Scenario - Generate Dialogue** \\
Follow plan, use student quotes from evidence, maintain persona (do NOT assume active engagement or verbal expressiveness unless documented) \\
\\
**STEP 4: For EACH Scenario - Write Minimal Context** \\
Using the provided SCENARIO, write one natural sentence describing what the student is doing and what the goal is. \\
  - Include: task + setting \\
  - Exclude: student characteristics, struggles, or solutions \\
\\
-------------------------------- \\
\#\#\# OUTPUT STRUCTURE \\
-------------------------------- \\
\{ \\
  ``category\_id'': ``\{CATEGORY\_ID\}'', \\
  ``paper\_id'': ``\{ID\}'', \\
  ``name'': ``student category name'', \\
  \\
  ``dialogues'': [ \\
    \{ \\
      ``case\_id'': ``\{CATEGORY\_ID\}\_case\_001'', \\
      \\
      ``dialogue'': [ \\
        \{ \\
          ``turn\_id'': 1, \\
          ``speaker'': ``tutor'', \\
          ``message'': ``Clear task presentation. Neutral tone, no effective strategies'' \\
        \}, \\
        \{ \\
          ``turn\_id'': 2, \\
          ``speaker'': ``student'', \\
          ``message'': ``Response based on persona'', \\
          ``evidence\_basis'': [``0001\_learner\_002''] \\
        \}, \\
        \{ \\
          ``turn\_id'': 3, \\
          ``speaker'': ``tutor'', \\
          ``message'': ``NEUTRAL response (generic)'' \\
        \}, \\
        \{ \\
          ``turn\_id'': 4, \\
          ``speaker'': ``student'', \\
          ``message'': ``Continued response'', \\
          ``evidence\_basis'': [``0001\_learner\_005''] \\
        \}, \\
        \{ \\
          ``turn\_id'': 5, \\
          ``speaker'': ``tutor'', \\
          ``message'': ``NEUTRAL response (generic)'' \\
        \}, \\
        \{ \\
          ``turn\_id'': 6, \\
          ``speaker'': ``student'', \\
          ``message'': ``CHALLENGING BEHAVIOR matching challenging\_behavior\_trigger'', \\
          ``is\_challenging\_behavior'': true, \\
          ``evidence\_basis'': [``0001\_learner\_008''] \\
        \} \\
      ], \\
     \\ 
      ``minimal\_context'': ``One natural sentence: what student is doing + goal. No characteristics, struggles, or solutions.'' \\
    \} \\
  ] \\
\} \\
\\
<System Prompt/User Prompt> \\
\\
-------------------------------- \\
\#\#\# INPUT CONTEXT \\
-------------------------------- \\
\\
STUDENT CATEGORY: \\
\{CATEGORY\} \\
\\
STUDENT PERSONA: \\
\{STUDENT\_PERSONA\} \\
\\
SCENARIO: \\
\{SCENARIO\} \\
\\
EVIDENCE (excerpt\_id → actual text from paper): \\
\{EVIDENCE\} \\
\\
-------------------------------- \\
\end{tcolorbox}


\vspace{0.3em}
\subsection{Rubric Construction} \label{app:stage5_rubric}
\vspace{0.3em}
\begin{tcolorbox}[
breakable, enhanced, top=1pt, left=1pt, right=1pt, bottom=1pt, colback=white, fontupper=\ttfamily, fonttitle=\bfseries\small,
title={Prompt}
]
\small
You are an expert in educational research, specialized in designing evaluation rubrics for tutors. \\
\\
Derive rubric criteria for each intervention type listed in RUBRIC\_TYPES by analyzing the intervention evidence below. \\
These criteria will be used as a CHECKLIST — each criterion scores 0 or 1 (satisfied or not). \\
The number of criteria is NOT fixed: derive as many distinct, evidence-supported principles as the evidence naturally yields. Do NOT pad to a fixed number; do NOT invent principles without evidence. \\
\\
Do NOT start from generic evaluation dimensions. Instead, read the evidence first and ask: \\
  - What underlying principles make these interventions work for THIS student? \\
  - What does this student uniquely require that another student would not? \\
\\
-------------------------------- \\
\#\#\# HOW TO DERIVE THE RUBRIC \\
-------------------------------- \\
\\
**Step 1 — Read the intervention evidence.** \\
Three evidence streams are provided: \\
  - EFFECTIVE\_IMMEDIATE: strategies that work for this student in the moment \\
  - EFFECTIVE\_LONG\_TERM: strategies that work for this student over weeks/months \\
  - INEFFECTIVE: strategies that fail or backfire for this student (applies to both rubric types) \\
\\
**Step 2 — Derive candidate principles separately for each rubric type.** \\
In principle\_analysis, derive two separate principle lists: \\
  - immediate\_candidate\_principles: from EFFECTIVE\_IMMEDIATE only \\
  - long\_term\_candidate\_principles: from EFFECTIVE\_LONG\_TERM only \\
  Each principle should be: \\
    - grounded in the evidence (briefly note which strategies support it) \\
    - capture something THIS student uniquely requires \\
    - broad enough to allow diverse tutor approaches to satisfy \\
  If two principles in the same list are redundant or overlapping, merge them. \\
  Extract as many meaningful and DISTINCT principles as the evidence supports. \\
\\
**Step 3 — Write each criterion as a CHECKLIST item.** \\
For each principle in immediate\_candidate\_principles, write one criterion in immediate\_rubric. \\
For each principle in long\_term\_candidate\_principles, write one criterion in long\_term\_rubric. \\
For each criterion, write: \\
  - A criterion\_name capturing the underlying principle (NOT a meta-label like ``Use of effective strategy'') \\
  - A rationale (1 sentence): why this principle matters for THIS student \\
  - A check condition (1 sentence): what must be present or absent in the tutor's response to earn 1 point. \\
    The check should be principle-level, not method-level — diverse tutor approaches can satisfy it. \\
\\
-------------------------------- \\
\#\#\# CRITICAL CONSTRAINTS \\
-------------------------------- \\
\\
- Do NOT name criteria using meta-labels like ``Use of effective strategies'', ``Avoidance of ineffective strategies'', ``Personalization'', ``Task re-engagement''. Use names that capture the underlying principle. \\
- Each criterion must be grounded in THIS student's documented evidence, not general best practice. \\
- The check condition must be satisfiable in multiple valid ways — good tutor responses are diverse. \\
- There are two rubric types: **immediate** (what the tutor does right now in the moment) and **long\_term** (sustained strategies over weeks/months). \\
- RUBRIC\_TYPES specifies which types of rubric to generate. If both are listed, generate both. If only one is listed, generate that one and set the other to null. \\
- Number of criteria is variable. Do NOT pad to a fixed number. Do NOT invent principles without evidence support. \\
\\
-------------------------------- \\
\#\#\# OUTPUT FORMAT \\
-------------------------------- \\
\\
Output ONLY valid JSON: \\
\\
\{ \\
  ``student\_description'': ``2–3 sentences: who this student is, their core behavioral pattern, and what the evidence reveals they uniquely need'', \\
\\
  ``principle\_analysis'': \{ \\
    ``from\_target\_interventions'': ``What key principles emerge from the effective strategies? What do these reveal about this student's needs? (2-3 sentences)'', \\
    ``student\_specific\_synthesis'': ``What is uniquely required for THIS student that another student would not need? (1-2 sentences)'', \\
    ``immediate\_candidate\_principles'': [ \\
      ``Candidate principle 1: [name] — [brief rationale grounded in evidence]'', \\
      ``Candidate principle 2: ...'' \\
    ], \\
    ``long\_term\_candidate\_principles'': [ \\
      ``Candidate principle 1: [name] — [brief rationale grounded in evidence]'', \\
      ``Candidate principle 2: ...'' \\
    ] \\
  \}, \\
\\
  ``immediate\_rubric'': \{ \\
    ``criteria'': [ \\
      \{ \\
        ``criterion\_id'': 1, \\
        ``criterion\_name'': ``Specific criterion capturing the principle'', \\
        ``rationale'': ``1 sentence: why this principle matters for THIS student in an immediate response'', \\
        ``check'': ``1 sentence: what must be present or absent in the tutor's immediate response to earn 1 point. Must be satisfiable by diverse approaches.'' \\
      \} \\
    ] \\
  \}, \\
\\
  ``long\_term\_rubric'': \{ \\
    ``criteria'': [ \\
      \{ \\
        ``criterion\_id'': 1, \\
        ``criterion\_name'': ``Specific criterion capturing the principle'', \\
        ``rationale'': ``1 sentence: why this principle matters for THIS student over time'', \\
        ``check'': ``1 sentence: what must be present or absent in the tutor's long-term plan to earn 1 point. Must be satisfiable by diverse approaches.'' \\
      \} \\
    ] \\
  \} \\
\} \\
\\
If a rubric type is NOT in RUBRIC\_TYPES, set that field to null: \\
  ``immediate\_rubric'': null \\
  ``long\_term\_rubric'': null \\
\\

<System Prompt/User Prompt> \\
\\
-------------------------------- \\
\#\#\# STUDENT PROFILE \\
-------------------------------- \\
\\
DESCRIPTION: \\
\{STUDENT\_DESCRIPTION\} \\
\\
BEHAVIORAL PATTERNS: \\
\{BEHAVIORAL\_PATTERNS\} \\
\\
UNIQUE TRIGGERS AND NEEDS: \\
\{UNIQUE\_TRIGGERS\_AND\_NEEDS\} \\
\\
APPROACH EFFECTIVENESS: \\
\{APPROACH\_EFFECTIVENESS\} \\
\\
PRIMARY CHALLENGE: \{PRIMARY\_CHALLENGE\} \\
\\
-------------------------------- \\
\#\#\# ACTUAL DIALOGUE EXAMPLES \\
-------------------------------- \\
\\
These are real dialogue cases showing how this student actually behaves in practice.
Use these to ground your rubric in concrete, observable behavior — not just abstract descriptions. \\
\\
\{DIALOGUE\_EXAMPLES\} \\
\\
-------------------------------- \\
\#\#\# INTERVENTION EVIDENCE \\
-------------------------------- \\
\\
EFFECTIVE\_IMMEDIATE (strategies that work in the moment for THIS student): \\
\{EFFECTIVE\_IMMEDIATE\} \\
\\
EFFECTIVE\_LONG\_TERM (strategies that work over weeks/months for THIS student): \\
\{EFFECTIVE\_LONG\_TERM\} \\
\\
INEFFECTIVE (strategies that fail or backfire for THIS student — applies to both rubric types): \\
\{INEFFECTIVE\} \\
\\
-------------------------------- \\
\#\#\# RUBRIC TYPES TO GENERATE \\
-------------------------------- \\
\\
\{RUBRIC\_TYPES\} \\

\end{tcolorbox}


\vspace{0.3em}
\subsection{MCQ Distractors Selection} \label{app:mcq_distract}
\vspace{0.3em}
\begin{tcolorbox}[
breakable, enhanced, top=1pt, left=1pt, right=1pt, bottom=1pt, colback=white, fontupper=\ttfamily, fonttitle=\bfseries\small,
title={Prompt}
]
\small
You are picking 3 distractors for a multiple-choice tutoring question. \\
\\
The CORRECT option is one effective intervention for THIS student in THIS dialogue. \\
The OTHER EFFECTIVE interventions listed below are also correct answers for the same
student per the source paper. \\
\\
You get up to 10 candidates, each a real effective intervention from a different scenario. \\
They were already embedding-filtered to be context-similar but behavior/intervention-different, so almost all are usable as-is. \\
\\
-------------------------------- \\
BE LENIENT \\
-------------------------------- \\
\\
A distractor only needs to be a plausible-sounding tutor action that is WRONG for THIS student. That is a low bar — most candidates clear it. \\
\\
Pick the 3 best. ALWAYS return exactly 3. \\
\\
Reject a candidate ONLY if one of these is true (rare, since embedding already
filtered): \\
\\
  1. STRUCTURALLY IMPOSSIBLE ACTION — the action's core mechanism depends on a setting structure that does not exist in this scenario and cannot be created by a wording swap (e.g. target is a 1-on-1 session but the action IS ``rotate the student across several small groups''; target is a single immediate moment but the action IS ``run a multi-week sequenced project''). A merely different subject, textbook, tool, or material is NOT a reason — those are freely rewritten downstream by the adapter (math textbook → English textbook, worksheet → whiteboard, etc.). Reject only when keeping the mechanism would be incoherent in this setting. \\
\\
  2. TOO CLOSE TO CORRECT — its action is essentially the same (just reworded) as the CORRECT option or any OTHER EFFECTIVE intervention. \\
\\
Anything else is a fine distractor — including ones that address a different student problem, use a quite different approach, or are only a so-so fit. Do NOT reject those. \\
\\
If rejecting leaves fewer than 3 clean candidates, still return 3 — fill the remaining
slots with the least-bad rejected ones and list those slots in `forced\_idxs'. Leave it
empty if all 3 are clean. \\
\\
-------------------------------- \\
INPUT \\
-------------------------------- \\
\\
TARGET DIALOGUE: \\
\{DIALOGUE\} \\
\\
TARGET STUDENT PERSONA: \\
\{STUDENT\_PERSONA\} \\
\\
TARGET SCENARIO CONTEXT (task, environment, materials, social setting): \\
\{TARGET\_CONTEXT\} \\
\\
RESPONSE TYPE: \{RESPONSE\_TYPE\} \\
\\
CORRECT INTERVENTION (the labeled correct answer — do not select): \\
\{CORRECT\_INTERVENTION\} \\
\\
OTHER EFFECTIVE INTERVENTIONS (also correct for this student per the source paper): \\
\{SIBLING\_INTERVENTIONS\} \\
\\
CANDIDATE DISTRACTORS (0-indexed, each from a different scenario): \\
\{CANDIDATE\_DISTRACTORS\} \\
\\
-------------------------------- \\
OUTPUT \\
-------------------------------- \\
\\
Output ONLY valid JSON. \\
\\
\{ \\
  ``kept\_idxs'': [<int>, <int>, <int>],   // EXACTLY 3 \\
  ``forced\_idxs'': [<int>, ...],          // subset of kept\_idxs rejected by rule 1 or 2 but kept to reach 3; [] if all clean \\
  ``reason'': ``<one short sentence>'' \\
\} \\
\end{tcolorbox}

\vspace{0.3em}
\subsection{MCQ Options Adaptation} \label{app:mcq_adapt}
\vspace{0.3em}
\begin{tcolorbox}[
breakable, enhanced, top=1pt, left=1pt, right=1pt, bottom=1pt, colback=white, fontupper=\ttfamily, fonttitle=\bfseries\small,
title={Prompt}
]
\small
You are aligning the four answer options of a multiple-choice tutoring question so they read uniformly — same surface vocabulary, similar length, similar tone — without changing what each option substantively does. \\
\\
You are given: the TARGET dialogue / context / persona, ONE correct option (effective for THIS student), and THREE distractors (real interventions from OTHER scenarios, wrong for THIS student). Options come from different sources, so they may use different vocabulary, lengths, and tones. Make them parallel so test-takers cannot identify the correct option by surface features alone. \\
\\
-------------------------------- \\
RULES \\
-------------------------------- \\
\\
PRESERVE \\
  - The MECHANISM of each option (e.g., ``scaffold by chunking'', ``visual + verbal cue'', ``behavior-specific praise'', ``offer a binary choice''). Do not switch mechanisms. \\
  - The correct option's appropriateness for THIS student. \\
  - Each distractor's wrongness for THIS student. Do not soften. \\
\\
CHANGE — surface only \\
  - Subject/activity names, materials/tools, group structures, named curricula, source-scenario student or staff names → swap to the target dialogue's vocabulary. \\
  - If an option's surface already fits the target context, leave it verbatim. \\
\\
LENGTH \& TONE PARITY \\
  - Priority when rules conflict (higher wins): \\
      1. Each option's MECHANISM unchanged; the correct option stays right for THIS
         student; each distractor stays wrong for THIS student. NEVER violate this. \\
      2. The four options roughly similar in length and tone. \\
  - Equalize length by TRIMMING the longer options first. \\
  - If trimming alone cannot close the gap, you MAY add a small amount of NEUTRAL
    filler to a short option — generic time/place/connective phrasing only \\
    (e.g. ``...during the activity'', ``...in this session''). The added words
    MUST NOT introduce a new method, step, or technique; MUST NOT claim any
    effect, benefit, or empirical/evidence support (no ``...which improves
    performance'', ``...with strong research backing''); and MUST NOT make a
    distractor any more appropriate (or the correct any less) for THIS student.
    If you cannot pad without touching the mechanism or its fit, leave it short —
    an uneven length is acceptable; changing substance is not.
  - All four in third-person tutor-action voice (never direct speech, never ``you''). \\
  - Match register across all four: professional, concise, action-focused. \\
\\
MINIMUM EDIT \\
  - Swap only the surface terms the target context requires. No swap needed →
    output verbatim. \\
  - Never introduce a new method/technique/step, never flip or soften substance.
    Neutral length-matching filler (per LENGTH \& TONE PARITY) is the ONLY permitted
    addition. \\
\\
EDIT LEVELS (one label for the whole pass) \\
  - none    : all four output verbatim. \\
  - surface : $\leq$ 3 surface words swapped per option. \\
  - light   : at least one option needed a short phrase rewrite (no direct target analogue for a surface element). \\
  - fuller  : AVOID. Only if a setting structure had no target equivalent and had to be re-described. Re-check that all mechanisms are intact and no wrongness was softened.
  If any option received neutral length-matching filler, say so explicitly in
  edit\_notes (e.g. ``padded distractor 2 with neutral filler for length parity''). \\
\\
-------------------------------- \\
INPUT \\
-------------------------------- \\
\\
TARGET DIALOGUE: \\
\{DIALOGUE\} \\
\\
TARGET SCENARIO CONTEXT (task, environment, materials, social setting): \\
\{TARGET\_CONTEXT\} \\
\\
TARGET STUDENT PERSONA: \\
\{STUDENT\_PERSONA\} \\
\\
RESPONSE TYPE: \{RESPONSE\_TYPE\} \\
\\
Each option below is given as its ACTION only. Adapt that action — do not add any
rationale, justification, expected effect, or outcome (none is provided, and none
should be inferred or written into the option). \\
\\
CORRECT (effective for THIS student): \\
  ACTION: \{CORRECT\_INSTRUCTION\} \\
\\
DISTRACTOR 1 (wrong for THIS student): \\
  ACTION: \{DISTRACTOR\_1\_INSTRUCTION\} \\
\\
DISTRACTOR 2 (wrong for THIS student): \\
  ACTION: \{DISTRACTOR\_2\_INSTRUCTION\} \\
\\
DISTRACTOR 3 (wrong for THIS student): \\
  ACTION: \{DISTRACTOR\_3\_INSTRUCTION\} \\
\\
-------------------------------- \\
OUTPUT \\
-------------------------------- \\
\\
Output ONLY valid JSON. \\
\\
\{ \\
  ``correct\_adapted'':      ``<correct option, surface-aligned; mechanism preserved>'', \\
  ``distractor\_1\_adapted'': ``<distractor 1, surface-aligned; wrongness preserved>'', \\
  ``distractor\_2\_adapted'': ``<distractor 2, surface-aligned; wrongness preserved>'', \\
  ``distractor\_3\_adapted'': ``<distractor 3, surface-aligned; wrongness preserved>'', \\
  ``edit\_level'': ``<one of: `none' | `surface' | `light' | `fuller'>'', \\
  ``edit\_notes'': ``<one short sentence summarizing what surface terms you swapped, or `no change'>'' \\
\} \\

\end{tcolorbox}


\vspace{0.3em}
\subsection{Tutor Response Generation (Immediate / Long-Term)} \label{app:open_response}
\vspace{0.3em}
\begin{tcolorbox}[
    breakable, enhanced,
    width=\columnwidth,
    top=1pt, left=1pt, right=1pt, bottom=1pt,
    colback=white,
    fontupper=\ttfamily\small,
    fonttitle=\bfseries\rmfamily\small,
    subtitle style={
        fonttitle=\bfseries\rmfamily\small,
        fontupper=\bfseries\rmfamily\small,
    },
    title={Immediate Response Prompt}
]
=============================================== \\
CATEGORY ID: \{CATEGORY\_ID\} \\
CASE ID: \{CASE\_ID\} \\
=============================================== \\
\\
---------------------------------------------- \\
SCENARIO \\
---------------------------------------------- \\
\\
\{MINIMAL\_CONTEXT\}
\\
---------------------------------------------- \\
DIALOGUE \\
---------------------------------------------- \\
\\
\{DIALOGUE\_TURNS\}
\\
---------------------------------------------- \\
YOUR TURN - IMMEDIATE RESPONSE \\
---------------------------------------------- \\
\\
The student has just exhibited challenging behavior. \\
\\
What would you say or do RIGHT NOW as the tutor in this moment? \\
\\
Provide a direct, immediate response (2-3 sentences) in this moment. Do NOT provide a multi-step plan or describe what you will do later. \\
\\
Your immediate response (Output Response ONLY): \\

\tcbsubtitle{Long-Term Intervention Prompt}
=============================================== \\
CATEGORY ID: \{CATEGORY\_ID\} \\
CASE ID: \{CASE\_ID\} \\
=============================================== \\
\\
---------------------------------------------- \\
SCENARIO \\
---------------------------------------------- \\
\\
\{MINIMAL\_CONTEXT\}
\\
---------------------------------------------- \\
DIALOGUE \\
---------------------------------------------- \\
\\
\{DIALOGUE\_TURNS\}
\\
---------------------------------------------- \\
YOUR TURN - LONG-TERM INTERVENTION \\
---------------------------------------------- \\
\\
The student has just exhibited challenging behavior. \\
\\
What sustained interventions should be implemented over time for this student? \\
Describe what long-term strategies and support structures you would put in place (2-3 sentences). This should be a narrative description of ongoing interventions over weeks or months, not immediate actions. \\
\\
Your long-term intervention plan (Output Intervention Plan ONLY): \\

\end{tcolorbox}

\vspace{0.3em}
\subsection{Rubric Grading} \label{app:rubric_grade}
\vspace{0.3em}
\begin{tcolorbox}[
breakable, enhanced, top=1pt, left=1pt, right=1pt, bottom=1pt, colback=white, fontupper=\ttfamily, fonttitle=\bfseries\small,
title={Prompt}
]
\small
You are an expert in educational research, specialized in evaluating tutor responses to challenging student behaviors. \\
\\
Score the tutor's response using the provided personalized rubric. \\
Each criterion is a CHECKLIST item — score 1 if the condition is met, 0 if not. \\
The rubric has a VARIABLE number of criteria (not fixed at 5). Score every criterion present in the rubric — do not skip, do not invent. \\
\\
-------------------------------- \\
\#\#\# SCORING INSTRUCTIONS \\
-------------------------------- \\
\\
For each criterion in the rubric: \\
- Read the ``check'' condition carefully \\
- Score 1 if the tutor response satisfies the condition, 0 if not \\
- The check is principle-level: diverse approaches can earn the point as long as the underlying principle is present \\
- Be strict but fair — surface mentions without substance do NOT earn the point \\
- Do NOT reward length, politeness, or fluency \\
\\
A tutor response does not need to satisfy every criterion to be considered acceptable; the rubric documents the full set of principles this student needs, and a single response can only address some of them. Score each criterion independently. \\
\\
-------------------------------- \\
\#\#\# RUBRIC (variable-length checklist) \\
-------------------------------- \\
\\
\{RUBRIC\} \\
\\
-------------------------------- \\
\#\#\# OUTPUT FORMAT \\
-------------------------------- \\
\\
Output ONLY valid JSON. The `criteria\_scores' array MUST contain exactly one entry per criterion in the rubric, in the same order, preserving each `criterion\_id'. \\
\\
\{ \\
  ``criteria\_scores'': [ \\
    \{ \\
      ``criterion\_id'': <int from rubric>, \\
      ``criterion\_name'': ``<copy from rubric>'', \\
      ``evidence'': ``1–2 sentences: specific evidence from the tutor response that justifies this score'', \\
      ``score'': 0 or 1 \\
    \} \\
  ], \\
  ``n\_criteria'': <number of criteria scored>, \\
  ``total\_score'': <sum of all criterion scores>, \\
  ``any\_satisfied'': <true if total\_score $>$ 0, else false>, \\
  ``overall\_feedback'': ``2–3 sentences: what the tutor did well, what principle was most notably missing, and the single most important improvement for this specific student'' \\
\}
\\
<System Prompt/User Prompt> \\
\\
-------------------------------- \\
\#\#\# STUDENT CONTEXT \\
-------------------------------- \\
\\
\{STUDENT\_DESCRIPTION\}
\\
-------------------------------- \\
\#\#\# DIALOGUE (situation the tutor is responding to) \\
-------------------------------- \\
\\
\{DIALOGUE\}
\\
FULL CONTEXT (JSON with fields: scenario\_id, title, content → {task, context, behavior\_patterns, challenging\_behavior\_trigger}): \\
Use `content.task' to understand what the student was asked to do, `content.context' for the setting/environment/materials, `content.behavior\_patterns' for the student's documented tendencies, and `content.challenging\_behavior\_trigger' for what the final student turn is showing.
\{FULL\_CONTEXT\}
\\
-------------------------------- \\
\#\#\# TUTOR RESPONSE \\
-------------------------------- \\
\\
\{TUTOR\_RESPONSE\} \\
\end{tcolorbox}


\vspace{0.3em}
\subsection{MCQ Response} \label{app:mcq_response}
\vspace{0.3em}
\begin{tcolorbox}[
breakable, enhanced, top=1pt, left=1pt, right=1pt, bottom=1pt, colback=white, fontupper=\ttfamily, fonttitle=\bfseries\small,
title={Prompt}
]
\small
You are an expert special education evaluator assessing tutor responses to students with learning difficulties. \\
\\
<System Prompt/User Prompt> \\
\\
Below is a tutoring dialogue between a tutor and a student with learning difficulties, followed by a multiple-choice question about the most appropriate intervention.
\\
\#\# Dialogue \\
\{DIALOGUE\} \\
\\
\#\# Question \\
Based on the dialogue above, which intervention is most appropriate for this student in this situation? \\
\\
\{OPTIONS\}
\\
Respond in JSON format: \\
\{
  ``answer'': ``<option letter>'',
  ``reasoning'': ``<brief explanation of why this intervention is most appropriate for this specific student>''
\}

\end{tcolorbox}


\vspace{0.3em}
\subsection{MCQ Pair Selection} \label{app:binary_distract}
\vspace{0.3em}
\begin{tcolorbox}[
breakable, enhanced, top=1pt, left=1pt, right=1pt, bottom=1pt, colback=white, fontupper=\ttfamily, fonttitle=\bfseries\small,
title={Prompt}
]
\small
You are selecting the single best (effective, ineffective) instruction pair to use as the two answer options for a multiple-choice tutoring question. \\
\\
You are given: \\
  - DIALOGUE: the tutor-student exchange leading up to the moment the tutor must respond. \\
  - STUDENT\_CONTEXT: persona-level background on the student. \\
  - RESPONSE\_TYPE: either ``immediate'' (a single in-the-moment reply) or ``long\_term'' (a plan over days or weeks). \\
  - A list of EFFECTIVE candidate instructions (each is a tutor action that has been validated as effective FOR THIS STUDENT). \\
  - A list of INEFFECTIVE candidate instructions (each is a poor-practice / counter-example from the same case). \\
\\
Your job is to pick ONE effective and ONE ineffective so that the resulting question has matched granularity and is plausibly anchored in the dialogue moment. \\
\\
-------------------------------- \\
SELECTION CRITERIA \\
-------------------------------- \\
\\
Apply the following criteria, in priority order: \\
\\
1. SCOPE MATCHES RESPONSE\_TYPE. \\
   - If RESPONSE\_TYPE = ``immediate'': both options must read as something a tutor could do *in this moment* (single reply, on-the-spot redirect, in-the-moment instructional move). Avoid candidates that describe long-term policy, curriculum design, multi-week plans, or persistent classroom culture. \\
   - If RESPONSE\_TYPE = ``long\_term'': both options must read as a plan, policy, or sustained intervention over days/weeks. Avoid candidates that are clearly a one-off reaction to a single utterance. \\
\\
2. ACTION FORM, NOT STATE OR DIAGNOSIS. \\
   - Strongly prefer candidates phrased as a tutor action (``Do X'', ``Provide Y'', ``Guide the student to Z''). \\
   - Avoid candidates that read as a description of a *state* (``very low rates of approval''), a *trigger condition* (``when X happens''), a *paper finding* (``teachers in this study often...''), or a *student-side description*. \\
\\
3. PLAUSIBILITY AT THE DIALOGUE MOMENT. \\
   - Prefer candidates a tutor would actually consider given what just happened in the dialogue. The ineffective option should still be substantively poor for THIS student, but should be the kind of misstep a real tutor might make in this exact situation — not an unrelated topic. \\
\\
4. GRANULARITY PARITY between the two chosen options. \\
   - The two selected options should be at a comparable level of specificity. Avoid pairing a very long, multi-clause effective with a one-line abstract ineffective, or vice versa. \\
   - Comparable length and structure helps; identical length is not required. \\
\\
5. TIE-BREAKER: when multiple pairs satisfy 1–4 roughly equally, prefer the pair whose EFFECTIVE option is most directly responsive to the dialogue moment. \\
\\
-------------------------------- \\
HARD CONSTRAINTS \\
-------------------------------- \\
\\
- You MUST select indices that exist in the provided lists. Do NOT invent new instructions. \\
- You MUST select exactly one effective and one ineffective. \\
- Do NOT swap roles. Select your effective\_idx from the EFFECTIVE list and your ineffective\_idx from the INEFFECTIVE list. \\
- If no pair is ideal, still pick the best available pair — selection is mandatory, not optional. \\
\\
-------------------------------- \\
INPUT \\
-------------------------------- \\
\\
DIALOGUE: \\
\{DIALOGUE\} \\
\\
STUDENT PERSONA: \\
\{STUDENT\_PERSONA\} \\
\\
SCENARIO CONTENT (task, environment, trigger, behavior patterns): \\
\{SCENARIO\_CONTENT\} \\
\\
RESPONSE TYPE: \{RESPONSE\_TYPE\} \\
\\
EFFECTIVE CANDIDATES (0-indexed): \\
\{EFFECTIVE\_CANDIDATES\} \\
\\
INEFFECTIVE CANDIDATES (0-indexed): \\
\{INEFFECTIVE\_CANDIDATES\} \\
\\
-------------------------------- \\
OUTPUT \\
-------------------------------- \\
\\
Output ONLY valid JSON. \\
\\
\{ \\
  ``effective\_idx'': <integer index into EFFECTIVE\_CANDIDATES>, \\
  ``ineffective\_idx'': <integer index into INEFFECTIVE\_CANDIDATES>, \\
  ``reason'': ``<one short sentence: why this pair beats the alternatives on the criteria above>'' \\
\} \\

\end{tcolorbox}

\vspace{0.3em}
\subsection{MCQ Pair Adaptation} \label{app:binary_adapt}
\vspace{0.3em}
\begin{tcolorbox}[
breakable, enhanced, top=1pt, left=1pt, right=1pt, bottom=1pt, colback=white, fontupper=\ttfamily, fonttitle=\bfseries\small,
title={Prompt}
]
\small
You are calibrating two answer options for a multiple-choice tutoring question. \\
\\
You are given a dialogue and TWO source strategies from a case study: \\
  - CORRECT: an effective intervention the tutor could take \\
  - INEFFECTIVE: a counter-example from the same case (poor practice, missed opportunity, or known-failed strategy) \\
\\
PROBLEM: The two strategies, as written, are often stylistically uneven. The correct one is usually phrased as a concrete tutor action, while the ineffective one is sometimes phrased as a trigger description, a passive state, a description of what someone failed to do, or a paper-specific finding. This creates an unintended style cue that gives the answer away. \\
\\
YOUR JOB — make the two options PARALLEL in form and length. Use the smallest change that achieves this. Most options need only minor edits. \\
\\
PREFERENCE HIERARCHY (apply in this order, stop as soon as one is enough): \\
  1. No edit — if both INSTRUCTIONs are already concrete tutor actions of similar length and form, output them VERBATIM. \\
  2. Surface tweak only — flip ``tutor backs down'' → ``Back down from the request''; convert a state to an action verb form. Reuse the original words wherever possible. \\
  3. Light rewrite — only if the original is genuinely not interpretable as a tutor action (e.g., a trigger condition with no agent), reframe it as the simplest matching action. Stay close to the original wording. \\
  4. Fuller rewrite — AVOID. Use only as last resort if option (3) is not possible. \\
\\
TARGET OUTCOME: \\
  (1) Each option reads as a concrete action the tutor could plausibly take in this moment / over time. \\
  (2) Both have roughly similar length and structural form. \\
  (3) The INEFFECTIVE option remains clearly ineffective for THIS student — preserve its substantive content (the underlying poor approach), just reframed as something the tutor would do. \\
\\
HARD CONSTRAINTS: \\
  - DEFAULT to keeping the original wording. If in doubt, change less. \\
  - Do NOT change which option is effective. The correct stays effective; the ineffective stays ineffective. \\
  - Do NOT swap meaning, invent new content, or paraphrase for the sake of paraphrasing. \\
  - Do NOT make the ineffective option subtly "look effective". Its core flaw must still be visible. \\
  - Do NOT strip paper-specific names that name a real tutor action (e.g., "Boardmaker symbols", "Scratch program"); only strip jargon if it would be unintelligible without the paper. \\
  - Do NOT over-generalize. Keep each at the original level of specificity. \\
  - If you change anything, change as few words as possible. \\
  - The two options MUST end up roughly similar in length. If the originals differ a lot, trim the longer side's decorative phrases (parenthetical examples, trailing "so that..." clauses) to bring them closer. \\
  - When balancing length, ONLY trim the longer side. NEVER expand the shorter side — do not add words, clauses, or qualifiers that are not in the original. If the shorter option ends up much shorter than the longer one even after trimming, leave it short rather than invent content. \\
\\
-------------------------------- \\
INPUT \\
-------------------------------- \\
\\
DIALOGUE: \\
\{DIALOGUE\} \\
\\
STUDENT CONTEXT: \\
\{STUDENT\_CONTEXT\} \\
\\
RESPONSE TYPE: \{RESPONSE\_TYPE\}  (immediate $=$ a single in-the-moment reply / long\_term $=$ a plan over days-weeks) \\
\\
CORRECT (effective for this student): \\
  INSTRUCTION: \{CORRECT\_INSTRUCTION\} \\
  REASON: \{CORRECT\_REASON\} \\
  OUTCOME: \{CORRECT\_OUTCOME\} \\
\\
INEFFECTIVE (counter-example from same case): \\
  INSTRUCTION: \{INEFFECTIVE\_INSTRUCTION\} \\ 
  REASON: \{INEFFECTIVE\_REASON\} \\
  OUTCOME: \{INEFFECTIVE\_OUTCOME\} \\
\\
-------------------------------- \\
OUTPUT \\
-------------------------------- \\
\\
Output ONLY valid JSON. \\
\\
\{ \\
  ``correct\_adapted'': ``<correct option, edited as little as possible>'', \\
  ``ineffective\_adapted'': ``<ineffective option, edited as little as possible, still substantively poor>'', \\
  ``edit\_level'': ``<one of: `none' | `surface' | `light' | `fuller'>  // strongest level applied across the two options'', \\
  ``edit\_notes'': ``<one short sentence describing what you changed, or `no change' if edit\_level$=$none>'' \\
\} \\

\end{tcolorbox}


\vspace{0.3em}
\subsection{Reference-Based Response Generation} \label{app:reference_response}
\vspace{0.3em}
\begin{tcolorbox}[
breakable, enhanced, top=1pt, left=1pt, right=1pt, bottom=1pt, colback=white, fontupper=\ttfamily, fonttitle=\bfseries\small,
title={Prompt}
]
\small
-------------------------------- \\
REFERENCE — strategies from similar student cases \\
-------------------------------- \\
The strategies below are documented to have worked or failed for similar but different students. Treat them as reference only — they are not a script and were not written for the current student. Adapt them, combine them, or ignore them as appropriate. The cases are ordered by relevance and may conflict with each other; use your own judgment. \\
\\
\{REFERENCE\} \\
-------------------------------- \\
\\
\{PROMPT\} \\

\end{tcolorbox}


\vspace{0.3em}
\subsection{Dialogue Synthesis Prompt with Variable Turns} \label{app:turn_ablation_prompt}
\vspace{0.3em}
\begin{tcolorbox}[
breakable, enhanced, top=1pt, left=1pt, right=1pt, bottom=1pt, colback=white, fontupper=\ttfamily, fonttitle=\bfseries\small,
title={Prompt}
]
\small
You are a research assistant specialized in creating realistic dialogue from educational research scenarios. \\
\\
Your goal is to CREATE REALISTIC STUDENT-TUTOR DIALOGUES that demonstrate CHALLENGING MOMENTS requiring intervention. \\
\\
-------------------------------- \\
\#\#\# CRITICAL RULES \\
-------------------------------- \\
1. Ground ALL statements in provided profile/excerpts - NO INFERENCE \\
2. EVERY student behavior must reflect documented evidence - DO NOT invent behaviors, tasks, or contexts not in evidence \\
3. Reference evidence by ID for EVERY message \\
   - evidence\_basis arrays may contain ONLY IDs from ALLOWED evidence\_basis IDS below;
     inventing an ID, or citing one not on that list, invalidates the whole output \\
   - NEVER use meta labels such as "category\_extraction", "scenario", "persona", "paper", "prompt", or invented IDs \\
4. Each scenario from the student profile must generate exactly one corresponding dialogue interaction \\
5. Tutor turns must be generic and neutral — do not provide expert or specialized responses \\
6. The dialogue sets up a situation that REQUIRES good intervention \\
7. Final student turn shows challenging behavior that needs response (CHALLENGING BEHAVIORS = student behaviors that are difficult for a tutor to address using standard instructional strategies) \\
   - The final turn MUST be a student turn and MUST include "is\_challenging\_behavior": true \\
   - Every student turn MUST include a non-empty evidence\_basis array \\
8. Output ONLY valid JSON - NO explanations, NO markdown \\
\\
-------------------------------- \\
\#\#\# TASK OVERVIEW \\
-------------------------------- \\
\\
**STEP 1: For EACH Scenario - Plan Dialogue** \\
Review scenario from profile: \\
  - Task: What specific problem/activity? (concrete from evidence) \\
  - Context: Environment/materials/social setting (verify action possibilities) \\
  - Student persona: Core characteristics, typical responses \\
  - Difficulty trigger: What challenging behavior will end the dialogue? What causes final struggle? \\
  - Available actions: What tutor/student can physically do? \\
\\
**STEP 2: For EACH Scenario - Structure Dialogue** \\
- **Number of turns = EXACTLY \{TOTAL\_TURNS\} messages** (= 2 × \{TURN\} exchanges). This is a hard \\
  requirement: emit \{TOTAL\_TURNS\} objects in the "dialogue" array, no more and no fewer,
  alternating tutor → student and numbering turn\_id from 1 to \{TOTAL\_TURNS\}, so the
  first message is the tutor's and the last is the student's. \\
- **First Turn (Tutor)**: Present SPECIFIC, CONCRETE task \\
  - State the specific task from the scenario context (neutral tone) \\
  - Generic and neutral only — no expert or specialized responses \\
- **Turns in the middle (Mixed)**: Student attempts → Tutor NEUTRAL response \\
  - Student: May show emerging problem behaviors (based on evidence, follow persona, use quotes when appropriate) \\
  - Tutor: Generic/neutral response only (e.g., "What do you think?", "Keep trying") - DO NOT use effective strategies \\
  - Physical actions in parentheses ONLY if grounded in evidence and context allows \\
- **Last Turn (Student)**: MUST show challenging/struggling behavior \\
  - Observable problem behaviors matching challenging\_behavior\_trigger \\
  - Grounded in documented behavior\_patterns \\
\\
**STEP 3: For EACH Scenario - Generate Dialogue** \\
Follow plan, use student quotes from evidence, maintain persona (do NOT assume active engagement or verbal expressiveness unless documented) \\
\\
**STEP 4: For EACH Scenario - Write Minimal Context** \\
Using the provided SCENARIO, write one natural sentence describing what the student is doing and what the goal is. \\
  - Include: task + setting \\
  - Exclude: student characteristics, struggles, or solutions \\
\\

-------------------------------- \\
\#\#\# OUTPUT STRUCTURE \\
-------------------------------- \\
\{ \\
  "category\_id": "\{CATEGORY\_ID\}", \\
  "paper\_id": "\{ID\}", \\
  "name": "student category name", \\
  \\
  "dialogues": [ \\
    \{ \\
      "case\_id": "\{CATEGORY\_ID\}\_case\_001", \\
      \\
      "dialogue": [ \\
        \{ \\
          "turn\_id": 1, \\
          "speaker": "tutor", \\
          "message": "Clear task presentation. Neutral tone, no effective strategies" \\
        \}, \\
        \{ \\
          "turn\_id": 2, \\
          "speaker": "student", \\
          "message": "Response based on persona", \\
          "evidence\_basis": ["0001\_learner\_002"] \\
        \}, \\
        ... (alternate tutor/student until turn\_id \{PENULTIMATE\}) ... \\
        \{ \\
          "turn\_id": \{TOTAL\_TURNS\}, \\
          "speaker": "student", \\
          "message": "CHALLENGING BEHAVIOR matching challenging\_behavior\_trigger", \\
          "is\_challenging\_behavior": true, \\
          "evidence\_basis": ["0001\_learner\_008"] \\
        \} \\
      ], \\
      \\
      "minimal\_context": "One natural sentence: what student is doing + goal. No characteristics, struggles, or solutions." \\
    \} \\
  ] \\
\} \\
\\

<System Prompt/User Prompt> \\
\\
-------------------------------- \\
\#\#\# INPUT CONTEXT \\
-------------------------------- \\
STUDENT CATEGORY: \\
\{CATEGORY\} \\
\\
STUDENT PERSONA: \\
\{STUDENT\_PERSONA\} \\
\\
SCENARIO: \\
\{SCENARIO\} \\
\\
EVIDENCE (excerpt\_id → actual text from paper): \\
\{EVIDENCE\} \\
\\
ALLOWED evidence\_basis IDS — the complete list; any other ID is invalid, and IDs
from elsewhere in the paper must NOT be cited even if they seem relevant. Reuse
these across turns as needed: \\
\{ALLOWED\_EVIDENCE\_IDS\} \\

\end{tcolorbox}











\end{document}